\documentclass[acmtog,authorversion]{acmart}

\usepackage{booktabs} 

\setcitestyle{square}

\usepackage[ruled]{algorithm2e} 

\SetAlFnt{\small}
\SetAlCapFnt{\small}
\SetAlCapNameFnt{\small}
\SetAlCapHSkip{0pt}
\IncMargin{-\parindent}
\usepackage{tikz}
\acmJournal{TOG}

\usepackage{tabularx}
\newcolumntype{C}[1]{>{\centering\arraybackslash}m{#1}}

\acmSubmissionID{463}

\setcopyright{acmlicensed}

\acmYear{2022}\acmVolume{41}\acmNumber{4}\acmArticle{102}\acmMonth{7}
\acmDOI{10.1145/3528223.3530127}

\graphicspath{{./Figures/}{./figs/result/}}

\usepackage{booktabs} 
\usepackage{tabularx}
\newcolumntype{Y}{>{\centering\arraybackslash}X}

\usepackage{xargs}
\usepackage{microtype}
\usepackage[prologue,table]{xcolor}
\usepackage[binary-units]{siunitx}
\DeclareSIUnit{\nothing}{\relax}
\DeclareSIUnit{\fps}{FPS}
\DeclareSIUnit{\px}{px}

\usepackage{overpic}
\usepackage{stmaryrd}
\usepackage{makecell}
\usepackage{wrapfig}

\usepackage{enumitem}

\usepackage{multirow}
\usepackage{pgfplots}
\DeclareUnicodeCharacter{2212}{−}
\usepgfplotslibrary{groupplots,dateplot}
\usetikzlibrary{patterns,shapes.arrows}
\pgfplotsset{compat=newest}

\usepackage{pifont}

\def\commentType{1}

\ifnum\commentType=0
    \newcommandx{\customComment}[3]{}
    \newcommandx{\customTODO}[3]{}
\fi
\ifnum\commentType=1
    \newcommandx{\customComment}[3]{\textcolor{#2}{\textsl{#1: #3}}}
    \newcommandx{\customTODO}[3]{\textcolor{#2}{\textsl{#1: #3}}}
\fi
\ifnum\commentType=2
    \usepackage{pdfcomment}
    \newcommandx{\customComment}[3]{\pdfcomment[icon=Comment,opacity=0.5,color=#2,author=#1]{#3}}
    \newcommandx{\customTODO}[3]{\pdfcomment[icon=Note,opacity=0.5,color=#2,author=#1]{#3}}
\fi
\ifnum\commentType=3
    \usepackage{pdfcomment} 
    \usepackage{todonotes} 
    \usepackage{silence}
    \newcommandx{\customComment}[3]{\todo[color=#2!40,size=\small]{\textbf{#1:} #3}}
    \newcommandx{\customTODO}[3]{\todo[color=#2!40,size=\small]{\textbf{#1:} #3}}
\fi

\let\originalleft\left 
\let\originalright\right 
\renewcommand{\left}{\mathopen{}\mathclose\bgroup\originalleft} 
\renewcommand{\right}{\aftergroup\egroup\originalright} 

\definecolor{amber}{rgb}{1.0, 0.49, 0.0}
\definecolor{darkgreen}{rgb}{0.0, 0.5, 0.0}
\newcommandx{\All}[1]{\customComment{All}{red}{#1}}
\newcommandx{\AlexKeller}[1]{\customComment{AKeller}{blue}{#1}}
\newcommandx{\AlexEvans}[1]{\customComment{AEvans}{amber}{#1}}
\newcommandx{\Thomas}[1]{\customComment{Thomas}{darkgreen}{#1}}
\newcommandx{\Christoph}[1]{\customComment{Christoph}{orange}{#1}}

\newcommandx{\TODO}[1]{\customTODO{TODO}{red}{#1}}
\newcommandx{\AlexKellerTODO}[1]{\customTODO{AKeller}{blue}{#1}}
\newcommandx{\AlexEvansTODO}[1]{\customTODO{AEvans}{amber}{#1}}
\newcommandx{\ThomasTODO}[1]{\customTODO{Thomas}{darkgreen}{#1}}
\newcommandx{\ChristophTODO}[1]{\customTODO{Christoph}{orange}{#1}}

\newcommand{\IGNORE}[1]{}

\newcommand{\REMOVE}[1]{} 
\newcommand{\ADD}[1]{#1} 

\def\equationautorefname~#1\null{%
  Equation~(#1)\null
}

\newcommand{\Uniform}{{\mathcal{U}}}

\newcommand{\Params}{\theta}

\newcommand{\numAuxDims}{E}
\newcommand{\auxDims}{\mathbf{\xi}}
\newcommand{\enc}{\mathrm{enc}}
\newcommand{\encOut}{\mathbf{y}}
\newcommand{\interpWeight}{\mathbf{w}}
\newcommand{\nn}{m}
\newcommand{\hash}{h}
\newcommand{\primeNumber}{\pi}
\newcommand{\smoothstep}{S_1}

\newcommand{\entriesPerLevel}{T}

\newcommand{\featuresPerEntry}{F}

\newcommand{\levels}{L}
\newcommand{\level}{l}
\newcommand{\resolution}{N}
\newcommand{\minResolution}{N_\mathrm{min}}
\newcommand{\maxResolution}{N_\mathrm{max}}
\newcommand{\perLevelScale}{b}

\newcommand{\BigO}{\mathcal{O}}

\newcommand{\pos}{\mathbf{x}}

\newcommand{\Loss}{\mathcal{L}}

\newcommand{\R}{\mathbb{R}}
\newcommand{\Z}{\mathbb{Z}}

\newcommand{\tikzcircle}[2][red,fill=red]{\tikz[baseline=-0.7ex]\draw[#1,radius=#2] (0,0) circle ;}%

\newcommand{\sceneLego}{\textsc{Lego}\xspace}
\newcommand{\sceneDrums}{\textsc{Drums}\xspace}
\newcommand{\sceneShip}{\textsc{Ship}\xspace}
\newcommand{\sceneMic}{\textsc{Mic}\xspace}
\newcommand{\sceneFicus}{\textsc{Ficus}\xspace}
\newcommand{\sceneChair}{\textsc{Chair}\xspace}
\newcommand{\sceneHotdog}{\textsc{Hotdog}\xspace}
\newcommand{\sceneMaterials}{\textsc{Materials}\xspace}

\newcommand{\sceneTokyo}{\textsc{Tokyo}\xspace}

\newcommand{\scenePluto}{\textsc{Pluto}\xspace}
\newcommand{\sceneMars}{\textsc{Mars}\xspace}

\newcommand{\sceneClockwork}{\textsc{Clockwork}\xspace}
\newcommand{\sceneBeardedMan}{\textsc{Bearded Man}\xspace}

\newcommand{\sceneLizard}{\textsc{Lizard}\xspace}
\newcommand{\sceneCow}{\textsc{Cow}\xspace}

\definecolor{hgblue}{RGB}{138,200,224}
\definecolor{hgred}{RGB}{245,138,143}

\definecolor{gold}{RGB}{221, 196, 65}
\definecolor{silver}{RGB}{215, 215, 215}
\definecolor{bronze}{RGB}{126, 66, 5}

\begin{document}

\title{Instant Neural Graphics Primitives with a Multiresolution Hash Encoding}

\author{Thomas M\"uller}
\affiliation{%
  \institution{NVIDIA}
   \city{Z\"urich}
   \country{Switzerland}
  }
\email{tmueller@nvidia.com}

\author{Alex Evans}
\affiliation{%
  \institution{NVIDIA}
   \city{London}
   \country{United Kingdom}
  }
\email{alexe@nvidia.com}

\author{Christoph Schied}
\affiliation{%
  \institution{NVIDIA}
   \city{Seattle}
   \country{USA}
  }
\email{cschied@nvidia.com}

\author{Alexander Keller}
\affiliation{%
  \institution{NVIDIA}
   \city{Berlin}
   \country{Germany}
  }
\email{akeller@nvidia.com}

\renewcommand\shortauthors{M\"uller et al.}

\ccsdesc[300]{Computing methodologies~Massively parallel algorithms}
\ccsdesc[300]{Computing methodologies~Vector / streaming algorithms}
\ccsdesc[500]{Computing methodologies~Neural networks}

\keywords{Image Synthesis, Neural Networks, Encodings, Hashing, GPUs, Parallel Computation, Function Approximation. }

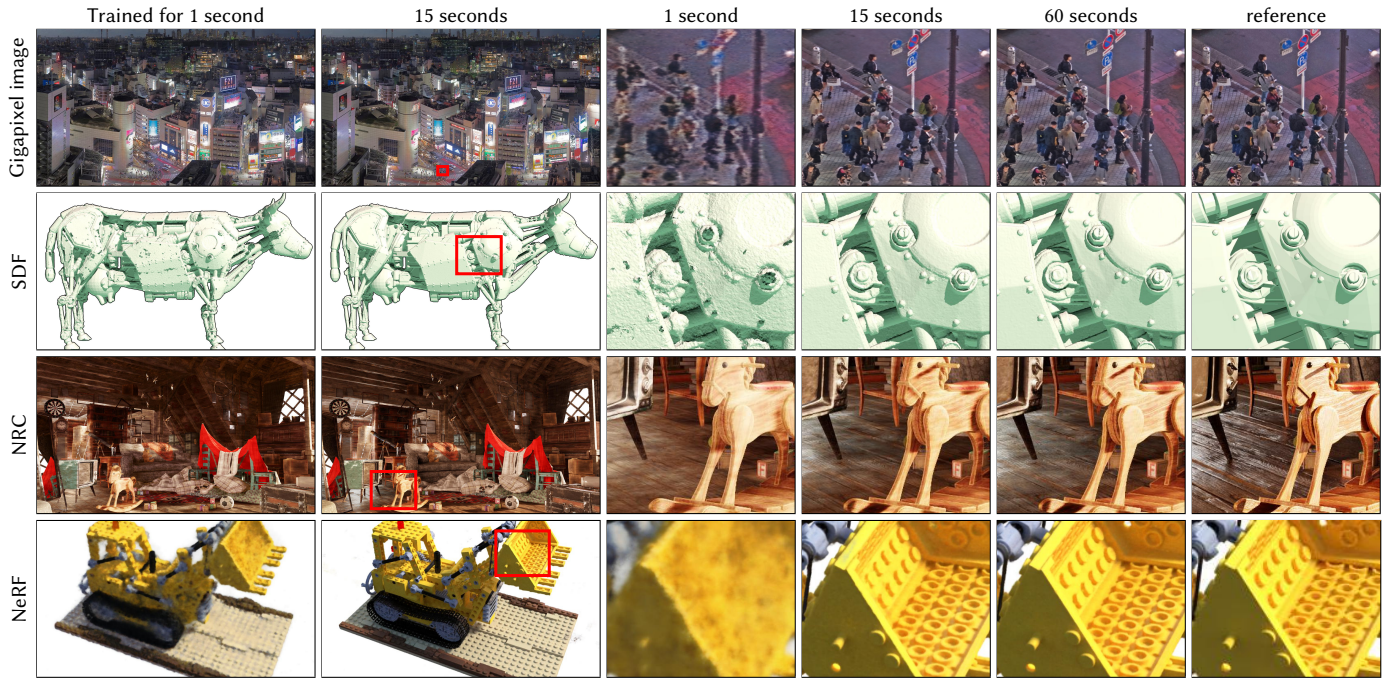
\begin{teaserfigure}
  {\Large\urlstyle{tt}\url{https://nvlabs.github.io/instant-ngp}}\\[1mm]
  \vspace{0mm}
  
{\centering
\setlength{\tabcolsep}{1.5pt}
\renewcommand{\arraystretch}{1}
\small
\sffamily
\hspace{-4.0mm}\begin{tabular}{C{0.28cm}C{3.685cm}C{3.685cm}C{2.4873749999999997cm}C{2.4873749999999997cm}C{2.4873749999999997cm}C{2.4873749999999997cm}}
	& Trained for $1$ second& $15$ seconds & $1$ second & $15$ seconds & $60$ seconds & reference \\[0pt]

	\rotatebox{90}{Gigapixel image} &
	\frame{\includegraphics[height=2.0728125cm]{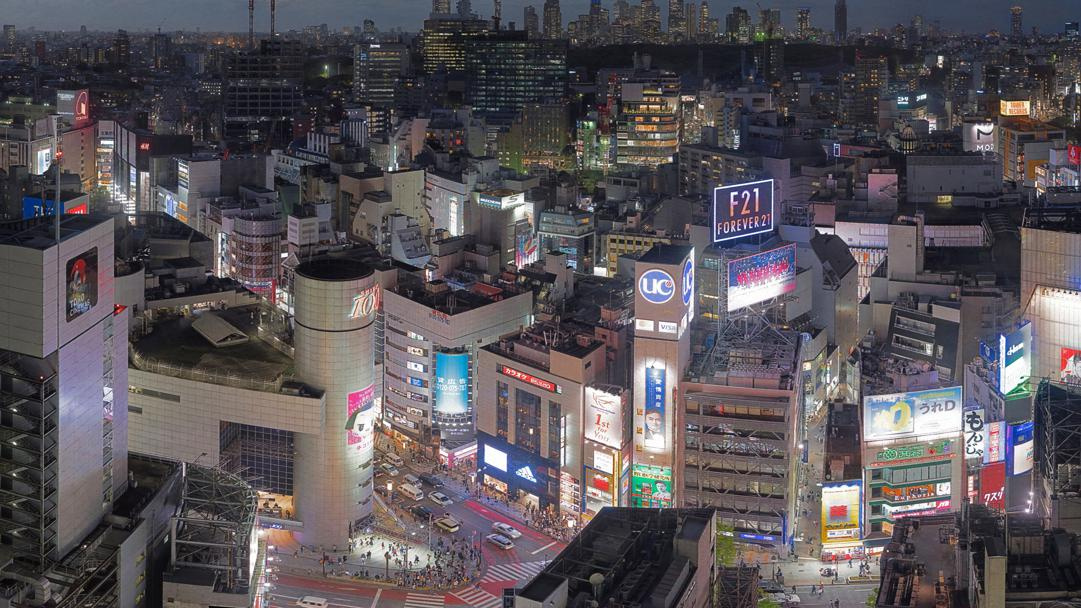}} &
	\frame{\begin{tikzpicture}
		\node[anchor=south west, inner sep=0] (image) at (0,0)
		{\includegraphics[height=2.0728125cm]{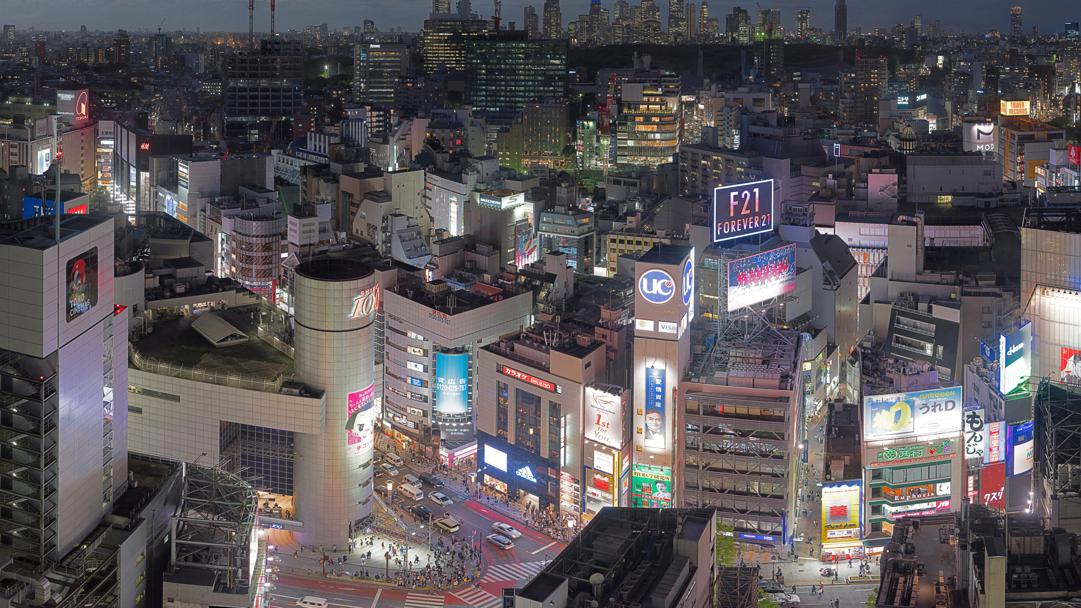}};
		\begin{scope}[x={(image.south east)},y={(image.north west)}]
		\draw[red, line width=0.4mm] (0.41632936278478083, 0.125) rectangle (0.4518561350757488, 0.07236842105263153); \end{scope}
	\end{tikzpicture}} &
	\frame{\includegraphics[height=2.0728125cm]{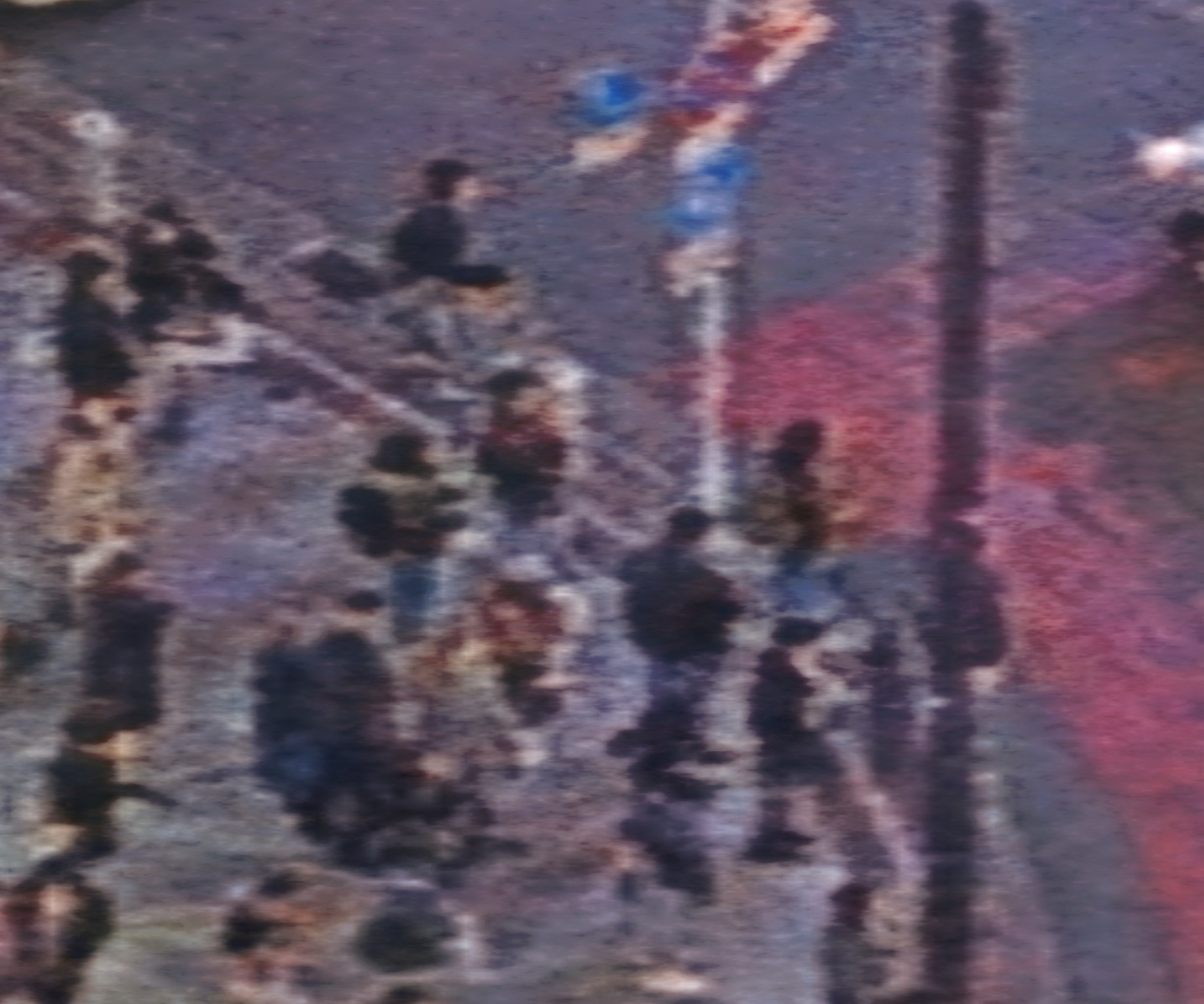}} &
	\frame{\includegraphics[height=2.0728125cm]{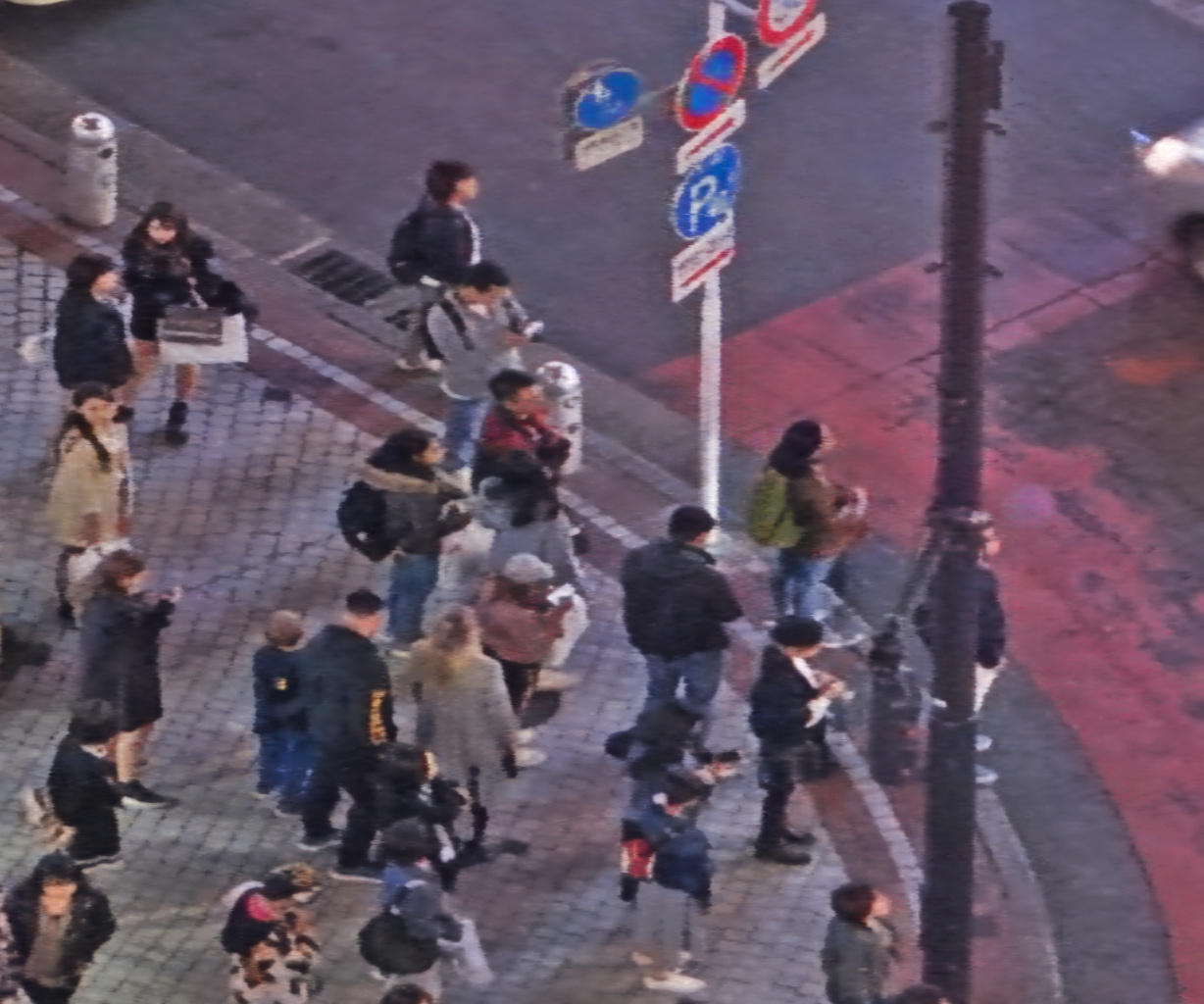}} &
	\frame{\includegraphics[height=2.0728125cm]{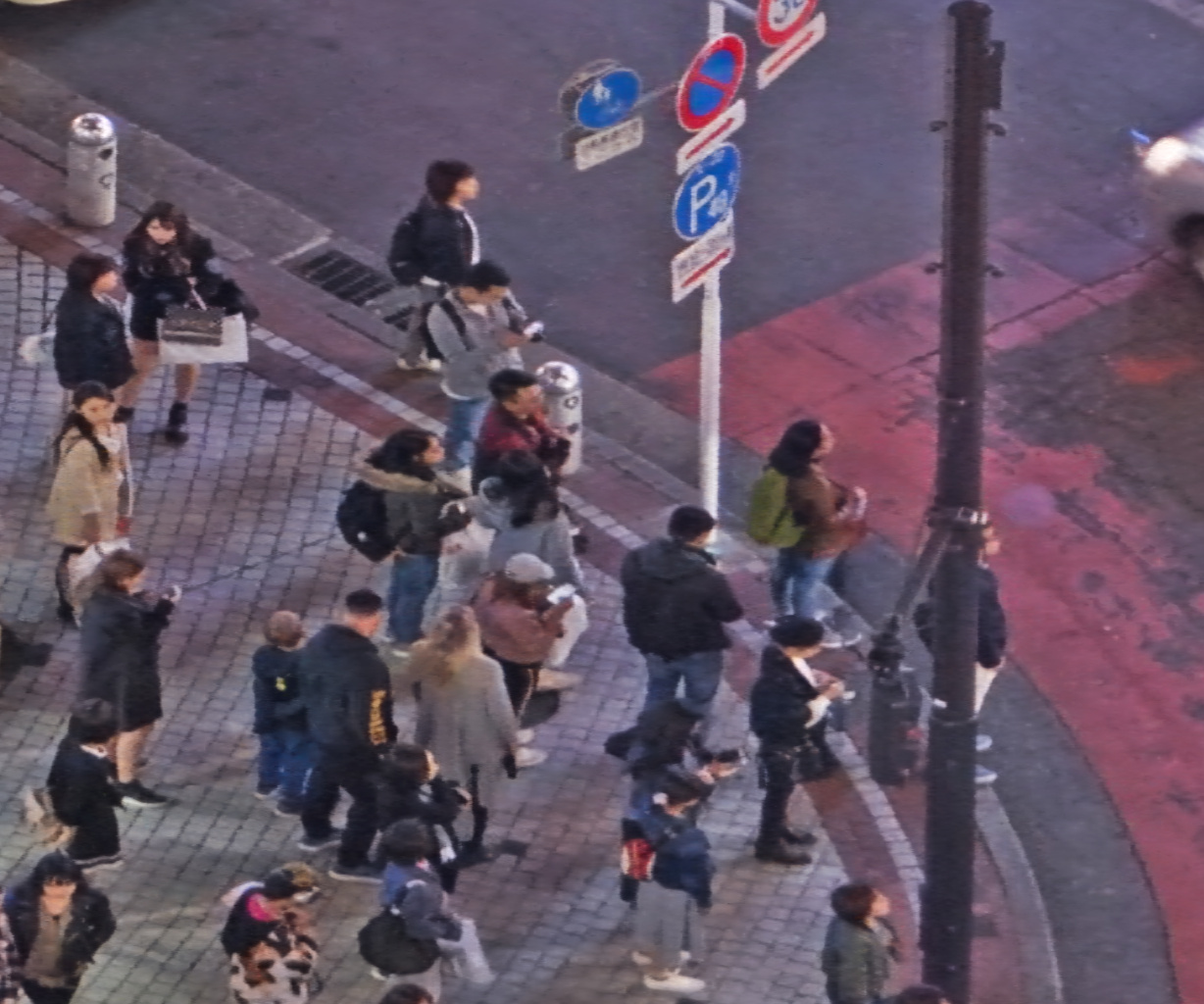}} &
	\frame{\includegraphics[height=2.0728125cm]{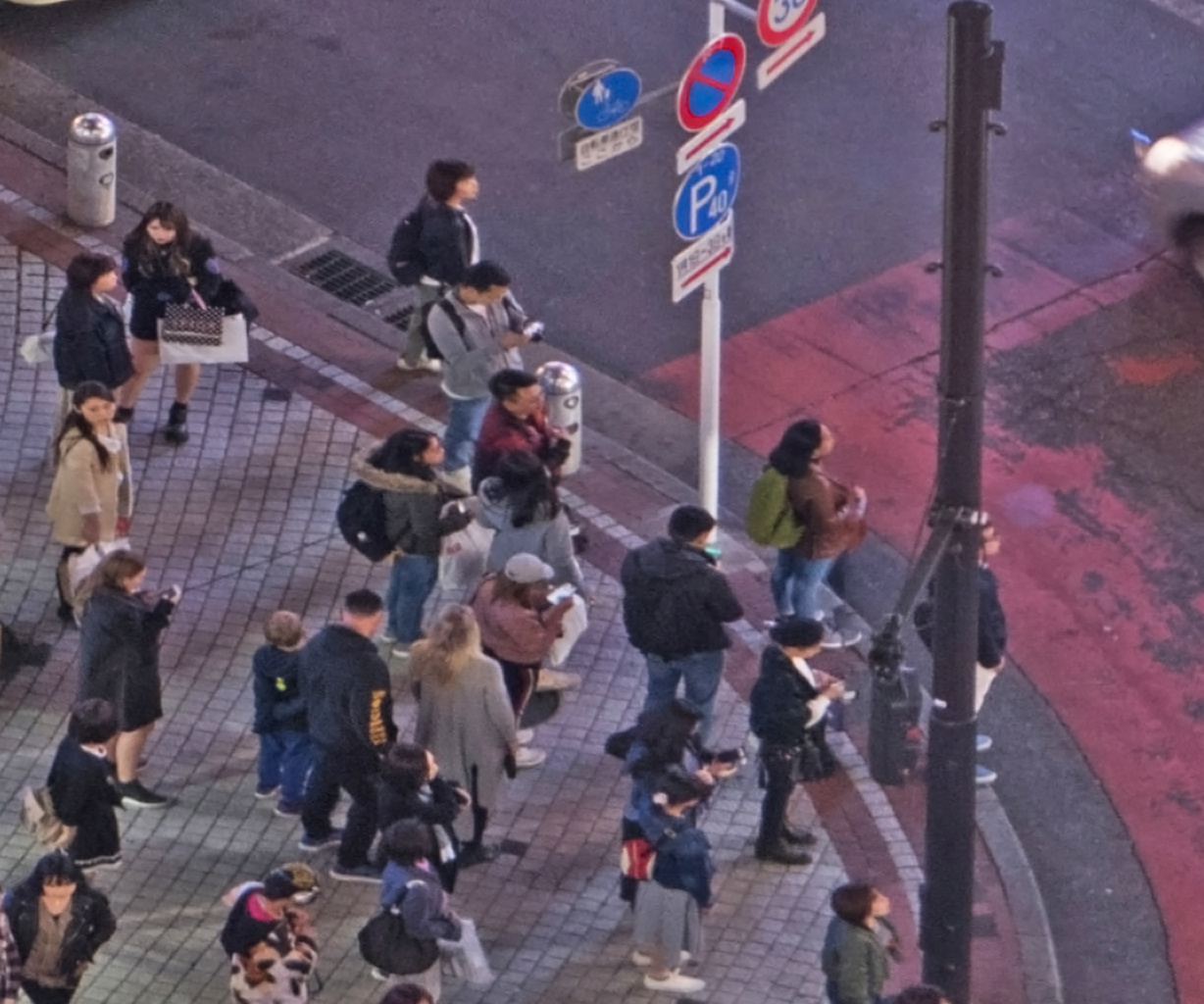}}
	\\[-0pt]

	\rotatebox{90}{SDF} &
	\frame{\includegraphics[height=2.0728125cm]{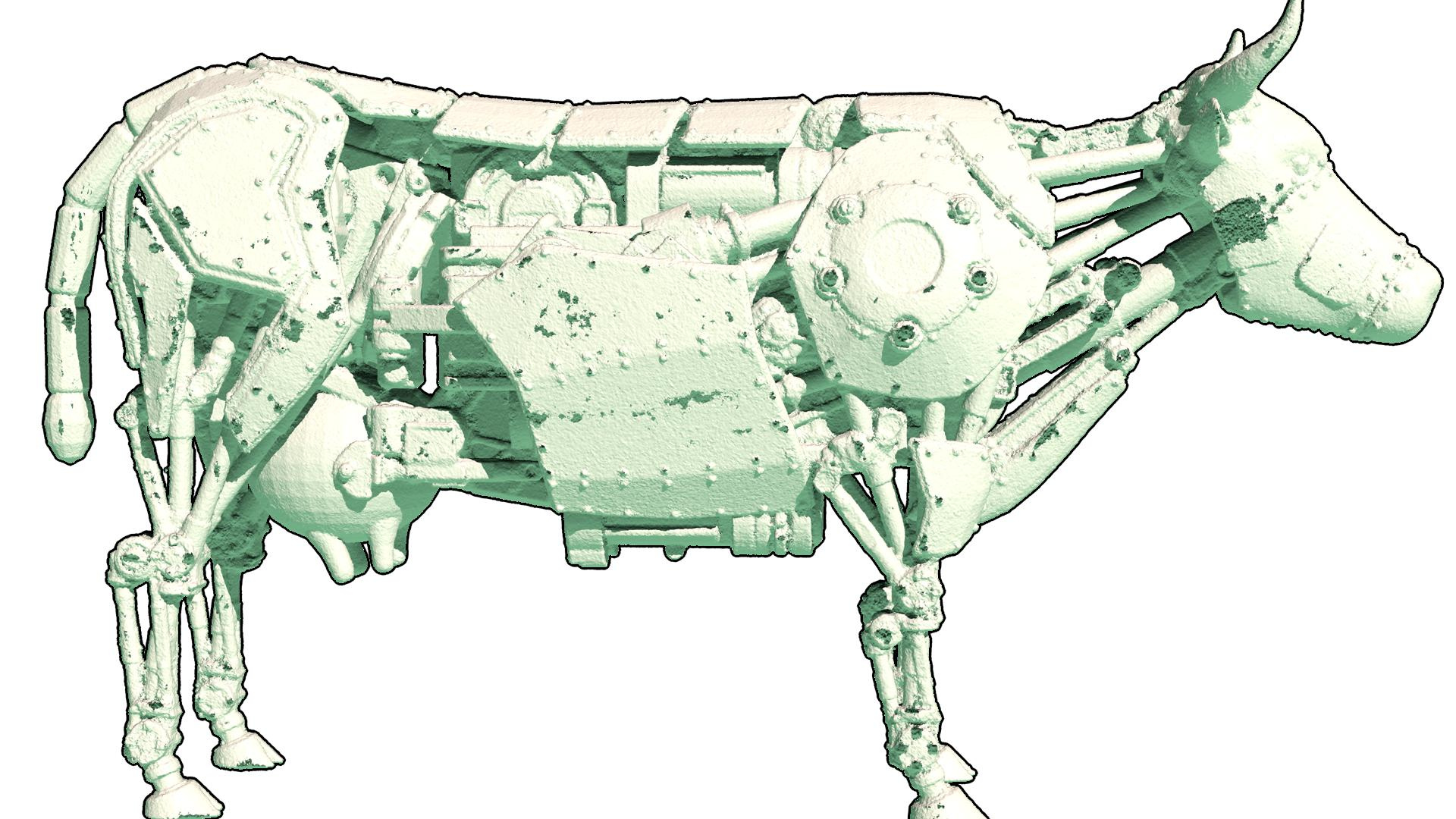}} &
	\frame{\begin{tikzpicture}
		\node[anchor=south west, inner sep=0] (image) at (0,0)
		{\includegraphics[height=2.0728125cm]{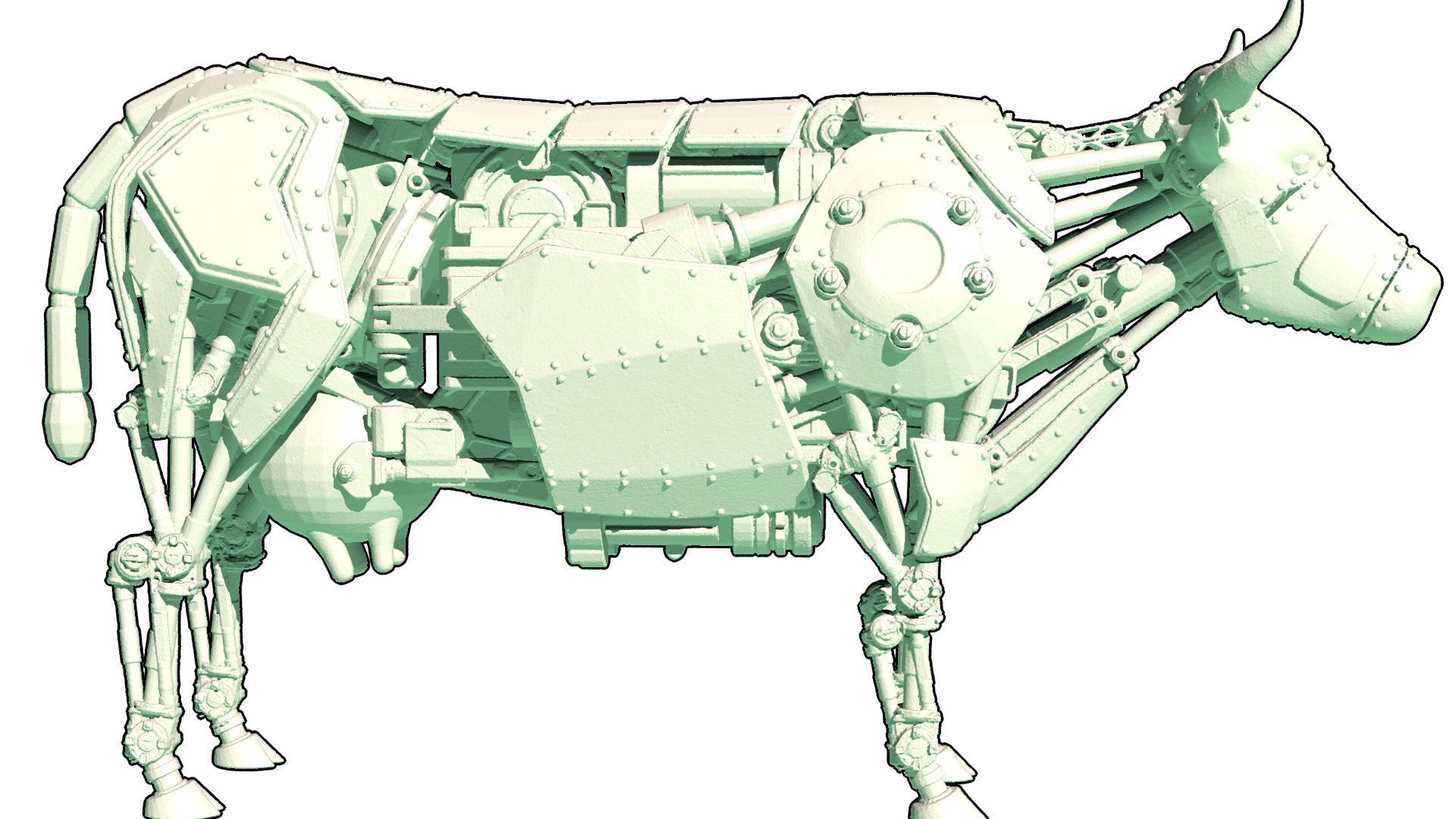}};
		\begin{scope}[x={(image.south east)},y={(image.north west)}]
		\draw[red, line width=0.4mm] (0.484375, 0.7222222222222222) rectangle (0.644375, 0.48518518518518516); \end{scope}
	\end{tikzpicture}} &
	\frame{\includegraphics[height=2.0728125cm]{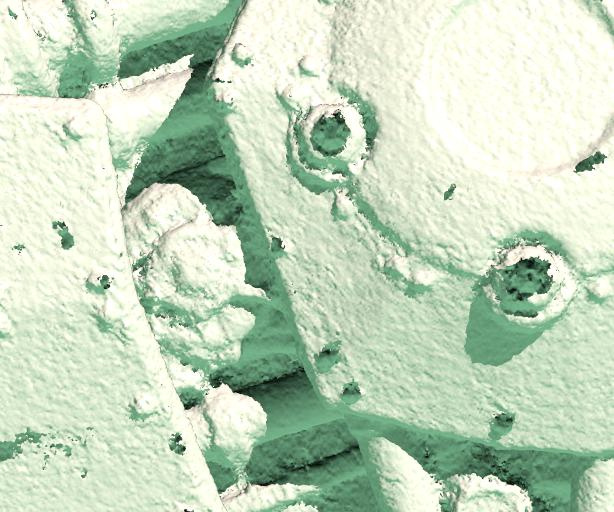}} &
	\frame{\includegraphics[height=2.0728125cm]{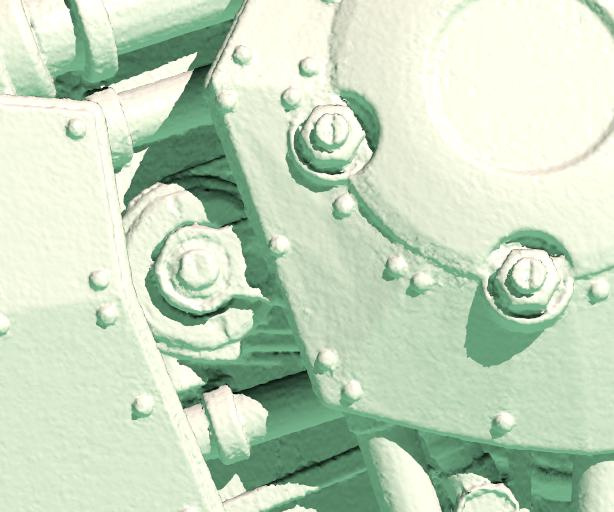}} &
	\frame{\includegraphics[height=2.0728125cm]{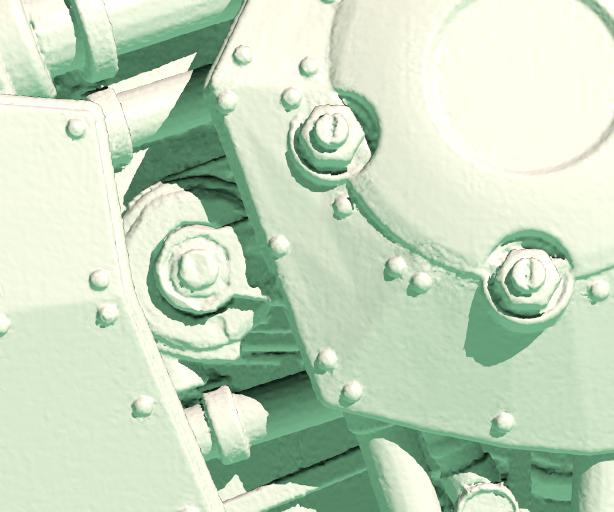}} &
	\frame{\includegraphics[height=2.0728125cm]{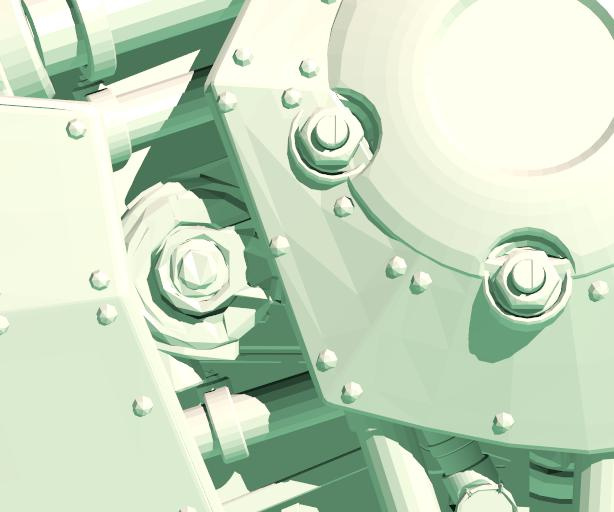}}
	\\[-0pt]

	\rotatebox{90}{NRC} &
	\frame{\includegraphics[height=2.0728125cm]{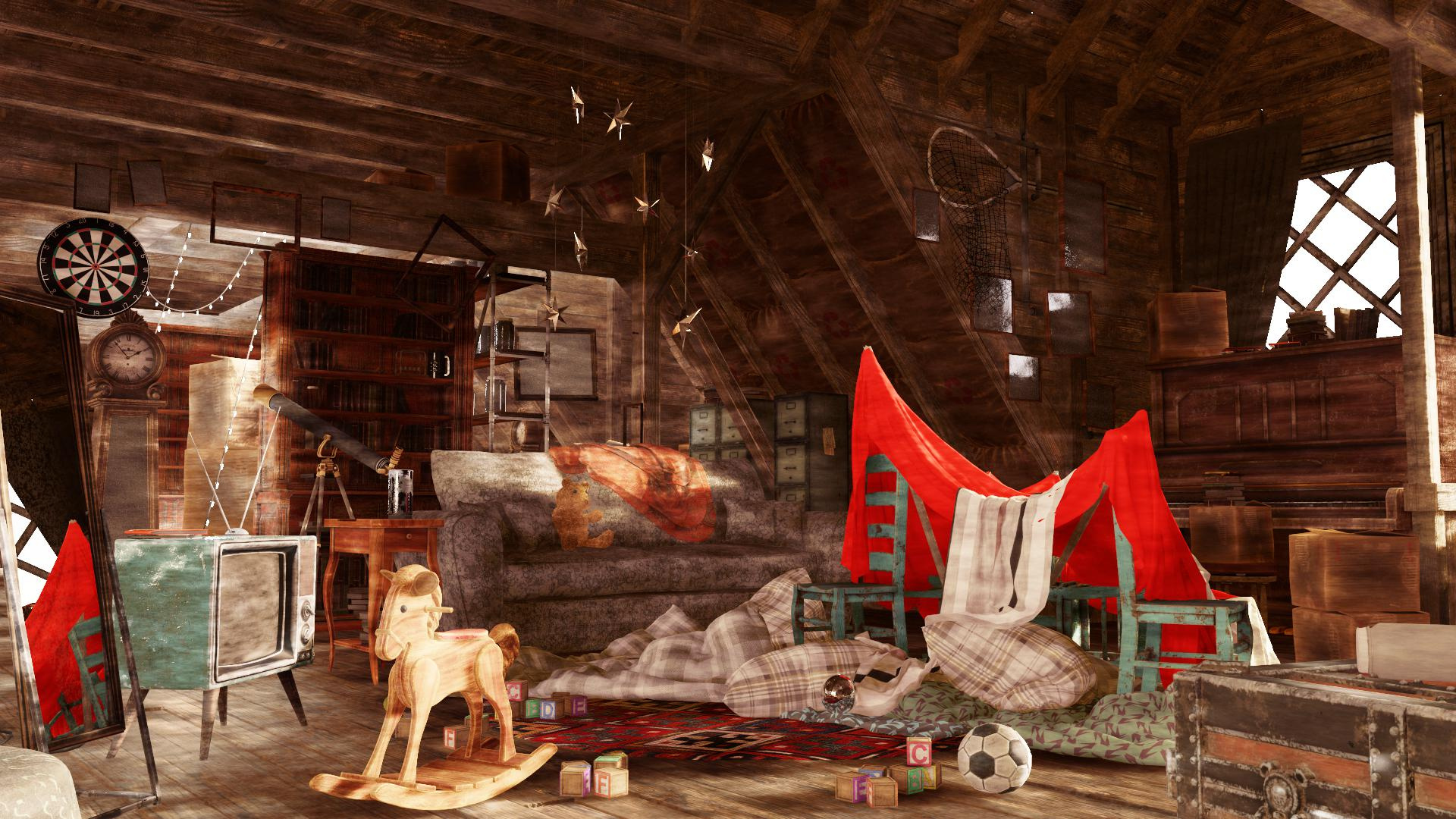}} &
	\frame{\begin{tikzpicture}
		\node[anchor=south west, inner sep=0] (image) at (0,0)
		{\includegraphics[height=2.0728125cm]{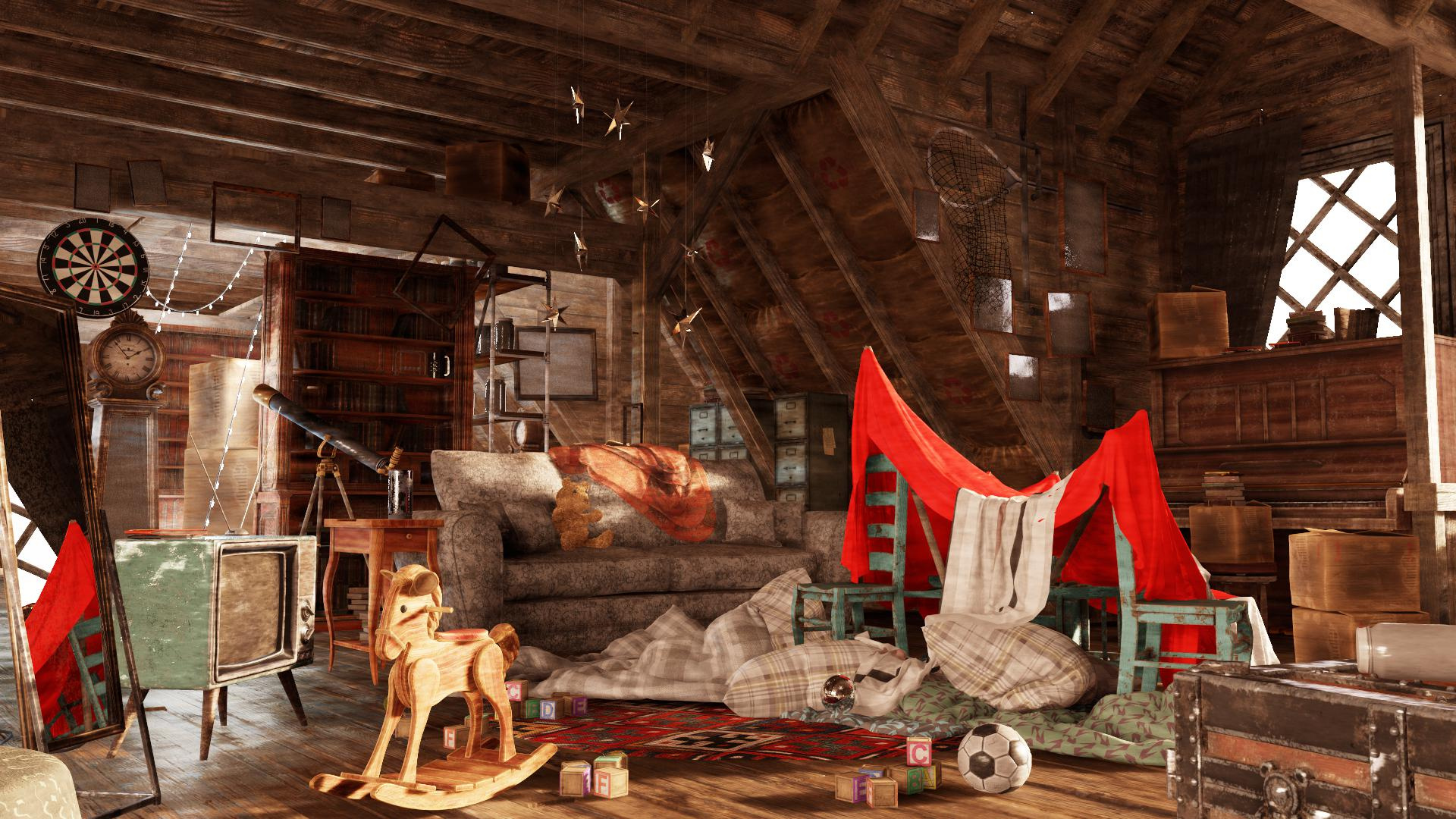}};
		\begin{scope}[x={(image.south east)},y={(image.north west)}]
		\draw[red, line width=0.4mm] (0.17708333333333334, 0.2685185185185185) rectangle (0.33708333333333335, 0.031481481481481444); \end{scope}
	\end{tikzpicture}} &
	\frame{\includegraphics[height=2.0728125cm]{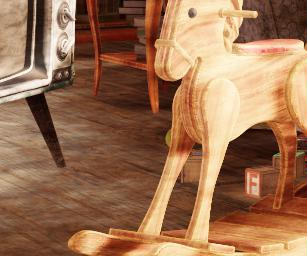}} &
	\frame{\includegraphics[height=2.0728125cm]{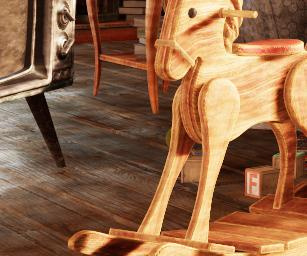}} &
	\frame{\includegraphics[height=2.0728125cm]{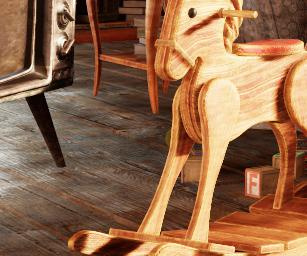}} &
	\frame{\includegraphics[height=2.0728125cm]{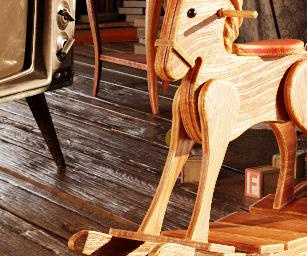}}
	\\[-0pt]

	\rotatebox{90}{NeRF} &
	\frame{\includegraphics[height=2.0728125cm]{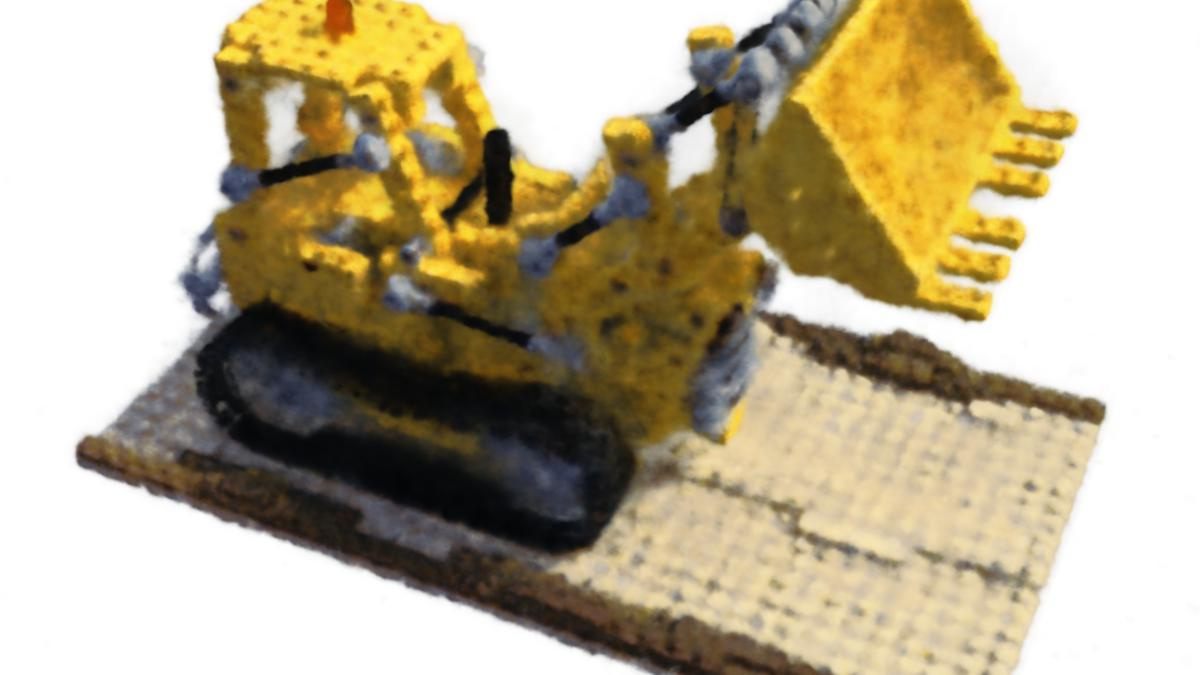}} &
	\frame{\begin{tikzpicture}
		\node[anchor=south west, inner sep=0] (image) at (0,0)
		{\includegraphics[height=2.0728125cm]{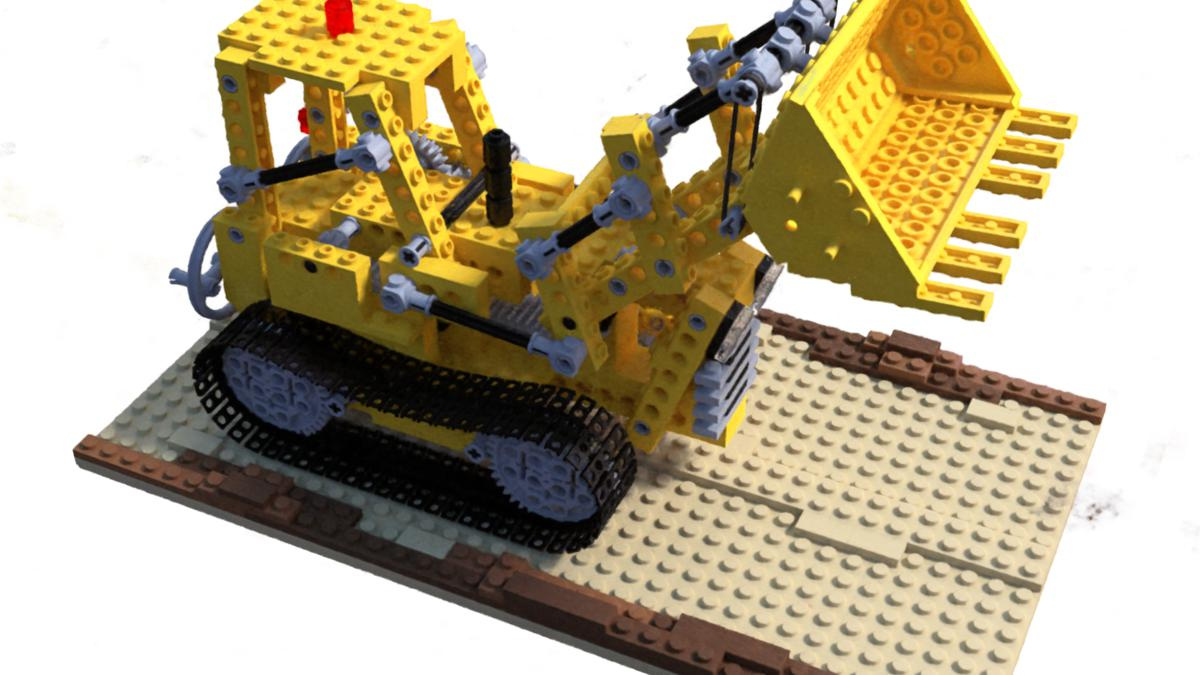}};
		\begin{scope}[x={(image.south east)},y={(image.north west)}]
		\draw[red, line width=0.4mm] (0.625, 0.9333333333333333) rectangle (0.817, 0.6488888888888888); \end{scope}
	\end{tikzpicture}} &
	\frame{\includegraphics[height=2.0728125cm]{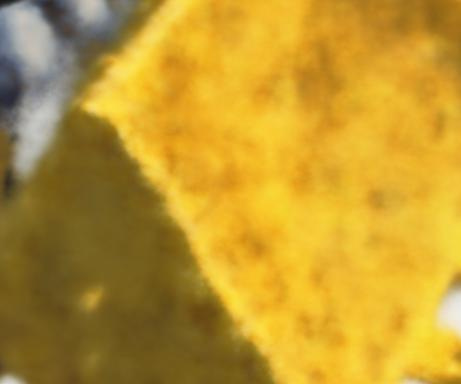}} &
	\frame{\includegraphics[height=2.0728125cm]{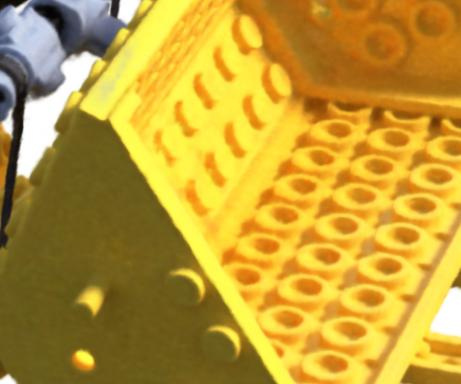}} &
	\frame{\includegraphics[height=2.0728125cm]{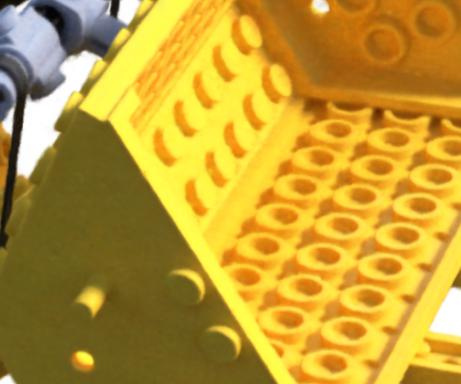}} &
	\frame{\includegraphics[height=2.0728125cm]{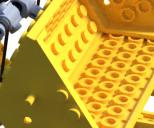}}
	\\[-0pt]

\end{tabular}}

  \vspace{-2mm}
  \caption{\label{fig:teaser}%
    We demonstrate instant training of neural graphics primitives on a single GPU for multiple tasks.
    In \emph{Gigapixel image} we represent a gigapixel image by a neural network.
    \emph{SDF} learns a signed distance function in 3D space whose zero level-set represents a 2D surface.  
    Neural radiance caching (\emph{NRC})~\cite{mueller2021realtime} employs a neural network that is trained in real-time to cache costly lighting calculations.
    Lastly, \emph{NeRF}~\cite{mildenhall2020nerf} uses 2D images and their camera poses to reconstruct a volumetric radiance-and-density field that is visualized using ray marching.
    In all tasks, our encoding and its efficient implementation provide clear benefits: rapid training, high quality, and simplicity.
    Our encoding is task-agnostic: we use the same implementation and hyperparameters across all tasks and only vary the hash table size which trades off quality and performance.
    Photograph \copyright Trevor Dobson \href{https://creativecommons.org/licenses/by-nc-nd/2.0/}{(CC BY-NC-ND 2.0)}
  }
\end{teaserfigure}

\begin{abstract}
Neural graphics primitives, parameterized by fully connected neural networks, can be costly to train and evaluate.
We reduce this cost with a versatile new input encoding that permits the use of a smaller network without sacrificing quality, thus significantly reducing the number of floating point and memory access operations:
a small neural network is augmented by a multiresolution hash table of trainable feature vectors whose values are optimized through stochastic gradient descent.
The multiresolution structure allows the network to disambiguate hash collisions, making for a simple architecture that is trivial to parallelize on modern GPUs.
We leverage this parallelism by implementing the whole system using fully-fused CUDA kernels with a focus on minimizing wasted bandwidth and compute operations.
We achieve a combined speedup of several orders of magnitude, enabling training of high-quality neural graphics primitives in a matter of seconds, and rendering in tens of milliseconds at a resolution of ${1920\!\times\!1080}$.
\end{abstract}

\maketitle

\section{Introduction}

Computer graphics primitives are fundamentally represented by mathematical functions that parameterize appearance.
The quality and performance characteristics of the mathematical representation are crucial for visual fidelity: we desire representations that remain fast and compact while capturing high-frequency, local detail.
Functions represented by multi-layer perceptrons (MLPs), used as \emph{neural graphics primitives}, have been shown to match these criteria (to varying degree), for example as representations of shape \cite{park2019deepsdf,martel2021acorn} and radiance fields~\cite{mildenhall2020nerf,liu2020neural,mueller2020neural,mueller2021realtime}.

The important commonality of the these approaches is an encoding that maps neural network inputs to a higher-dimensional space, which is key for extracting high approximation quality from compact models.
Most successful among these encodings are trainable, task-specific data structures~\cite{takikawa2021nglod,liu2020neural} that take on a large portion of the learning task.
This enables the use of smaller, more efficient MLPs.
However, such data structures rely on heuristics and structural modifications (such as pruning, splitting, or merging) that may complicate the training process, limit the method to a specific task, or limit performance on GPUs where control flow and pointer chasing is expensive.

We address these concerns with our multiresolution hash encoding, which is adaptive and efficient, independent of the task.
It is configured by just two values---the number of parameters $\entriesPerLevel$ and the desired finest resolution $\maxResolution$---yielding state-of-the-art quality on a variety of tasks (\autoref{fig:teaser}) after a few seconds of training.

Key to both the task-independent adaptivity and efficiency is a multiresolution hierarchy of hash tables:
\begin{itemize}[leftmargin=*]
  \item {\bf Adaptivity:} we map a cascade of grids to corresponding fixed-size arrays of feature vectors. At coarse resolutions, there is a 1:1 mapping from grid points to array entries. At fine resolutions, the array is treated as a hash table and indexed using a spatial hash function, where multiple grid points alias each array entry. Such hash collisions cause the colliding training gradients to average, meaning that the largest gradients---those most relevant to the loss function---will dominate. The hash tables thus \emph{automatically} prioritize the sparse areas with the most important fine scale detail.
  Unlike prior work, no structural updates to the data structure are needed at any point during training.
  \item {\bf Efficiency:} our hash table lookups are $\BigO(1)$ and do not require control flow. This maps well to modern GPUs, avoiding execution divergence and serial pointer-chasing inherent in tree traversals. The hash tables for all resolutions may be queried in parallel.
\end{itemize}
We validate our multiresolution hash encoding in four representative tasks (see \autoref{fig:teaser}):
\begin{enumerate}[leftmargin=*]
  \item \textbf{Gigapixel image:} the MLP learns the mapping from 2D coordinates to RGB colors of a high-resolution image.
  \item \textbf{Neural signed distance functions (SDF):} the MLP learns the mapping from 3D coordinates to the distance to a surface.
  \item \textbf{Neural radiance caching (NRC):} the MLP learns the 5D light field of a given scene from a Monte Carlo path tracer.
  \item \textbf{Neural radiance and density fields (NeRF):} the MLP learns the 3D density and 5D light field of a given scene from image observations and corresponding perspective transforms.
\end{enumerate}

In the following, we first review prior neural network encodings (\autoref{Sec:RelatedWork}), then we describe our encoding (\autoref{Sec:Algorithm}) and its implementation (\autoref{Sec:Implementation}), followed lastly by our experiments (\autoref{Sec:Experiments}) and discussion thereof (\autoref{Sec:Discussion}).

\begin{figure*}
  \vspace{-2mm}
  
\sffamily\small
\setlength{\tabcolsep}{1.8pt}%
\renewcommand{\arraystretch}{1}%
\hspace*{-0.9mm}\begin{tabular}{cccccc}
	\textbf{(a)} No encoding &
	\makecell{\textbf{(b)} Frequency \\ \cite{mildenhall2020nerf}} &
	\makecell{\textbf{(c)} Dense grid \\ Single resolution} &
	\makecell{\textbf{(d)} Dense grid \\ Multi resolution} &
	\makecell{\textbf{(e)} Hash table\ (ours) \\ $T=2^{14}$} &
	\makecell{\textbf{(f)} Hash table\ (ours) \\ $T=2^{19}$} \\

	\includegraphics[width=0.16\linewidth]{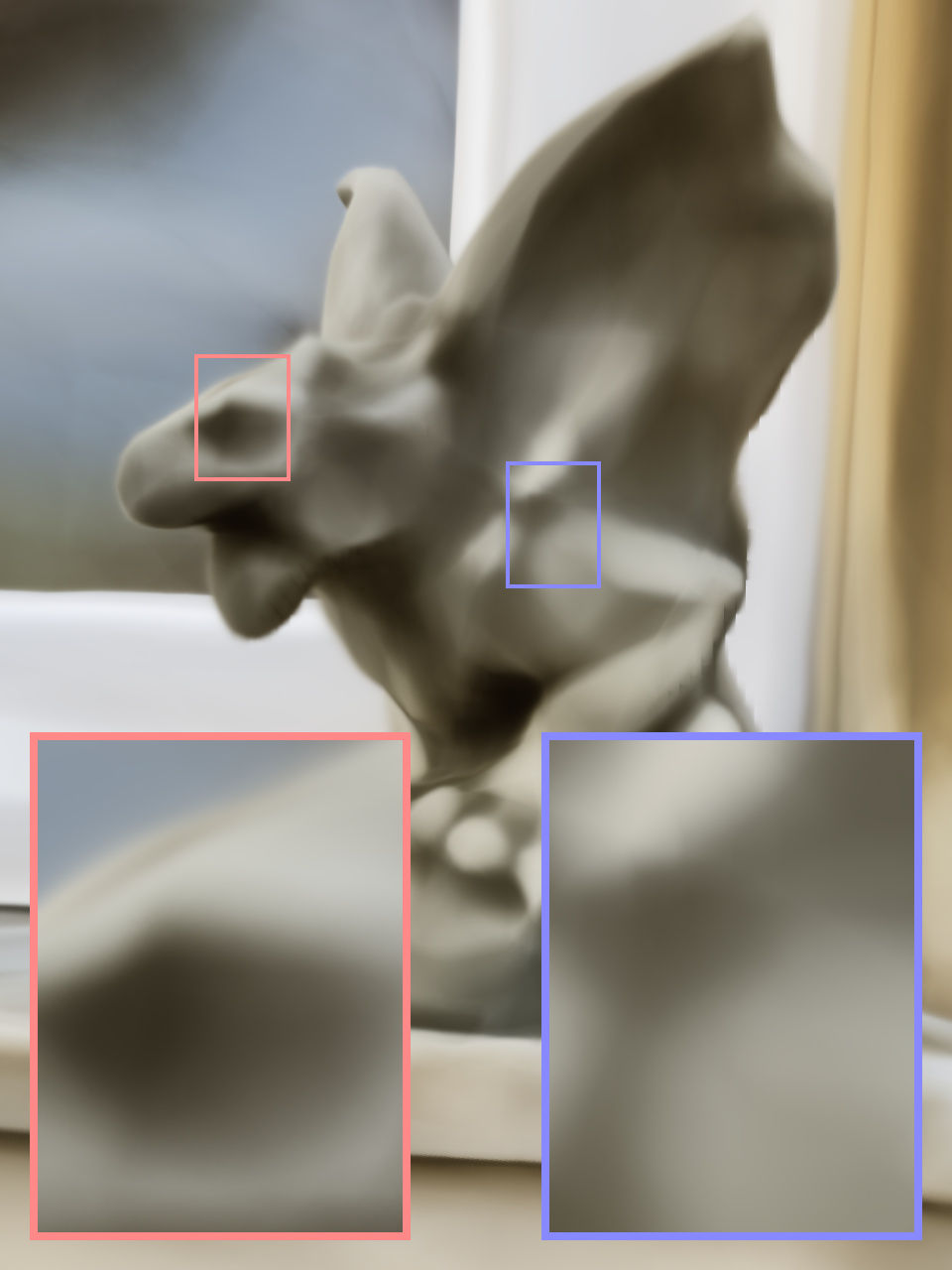} &
	\includegraphics[width=0.16\linewidth]{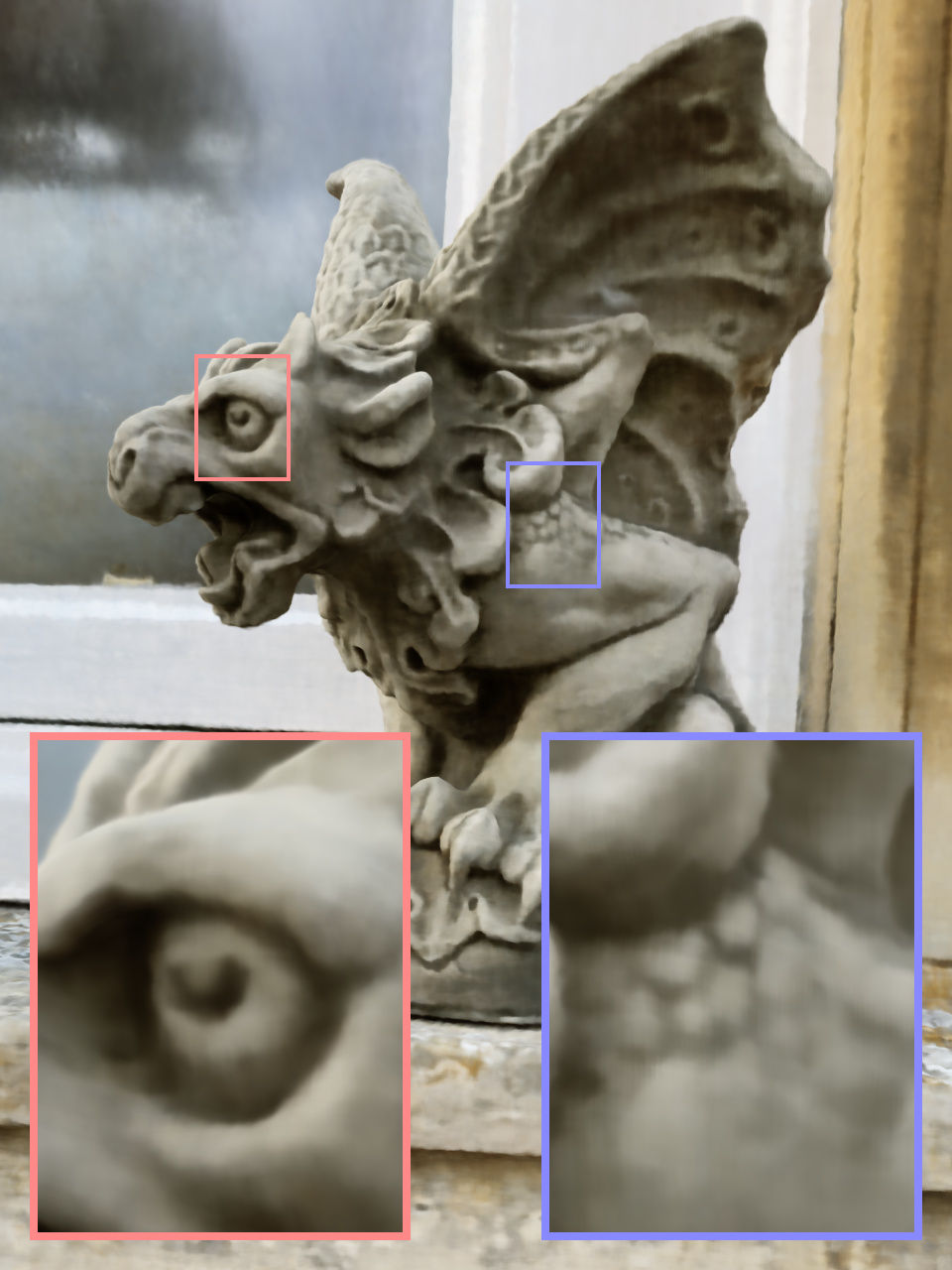} &
	\includegraphics[width=0.16\linewidth]{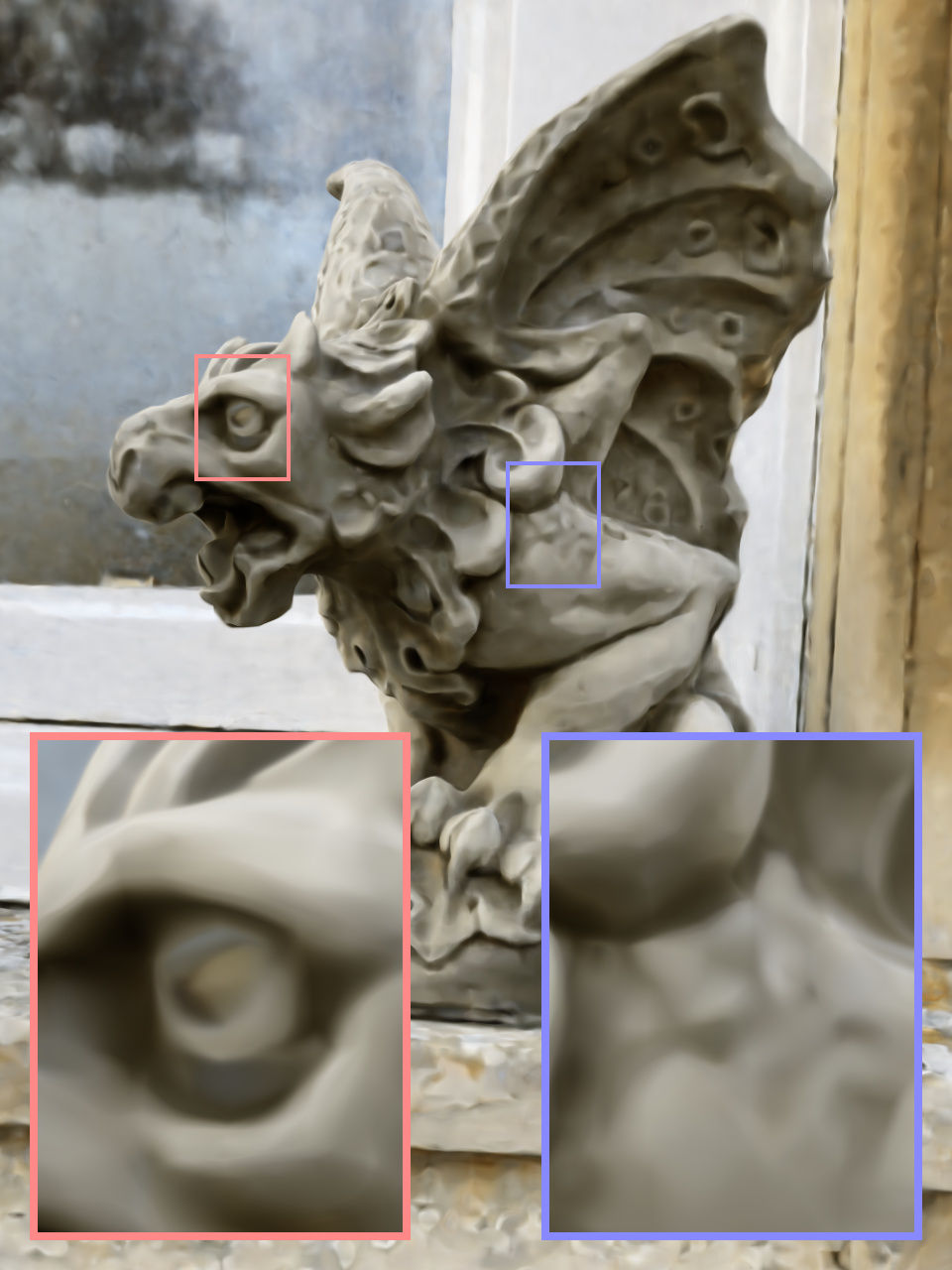} &
	\includegraphics[width=0.16\linewidth]{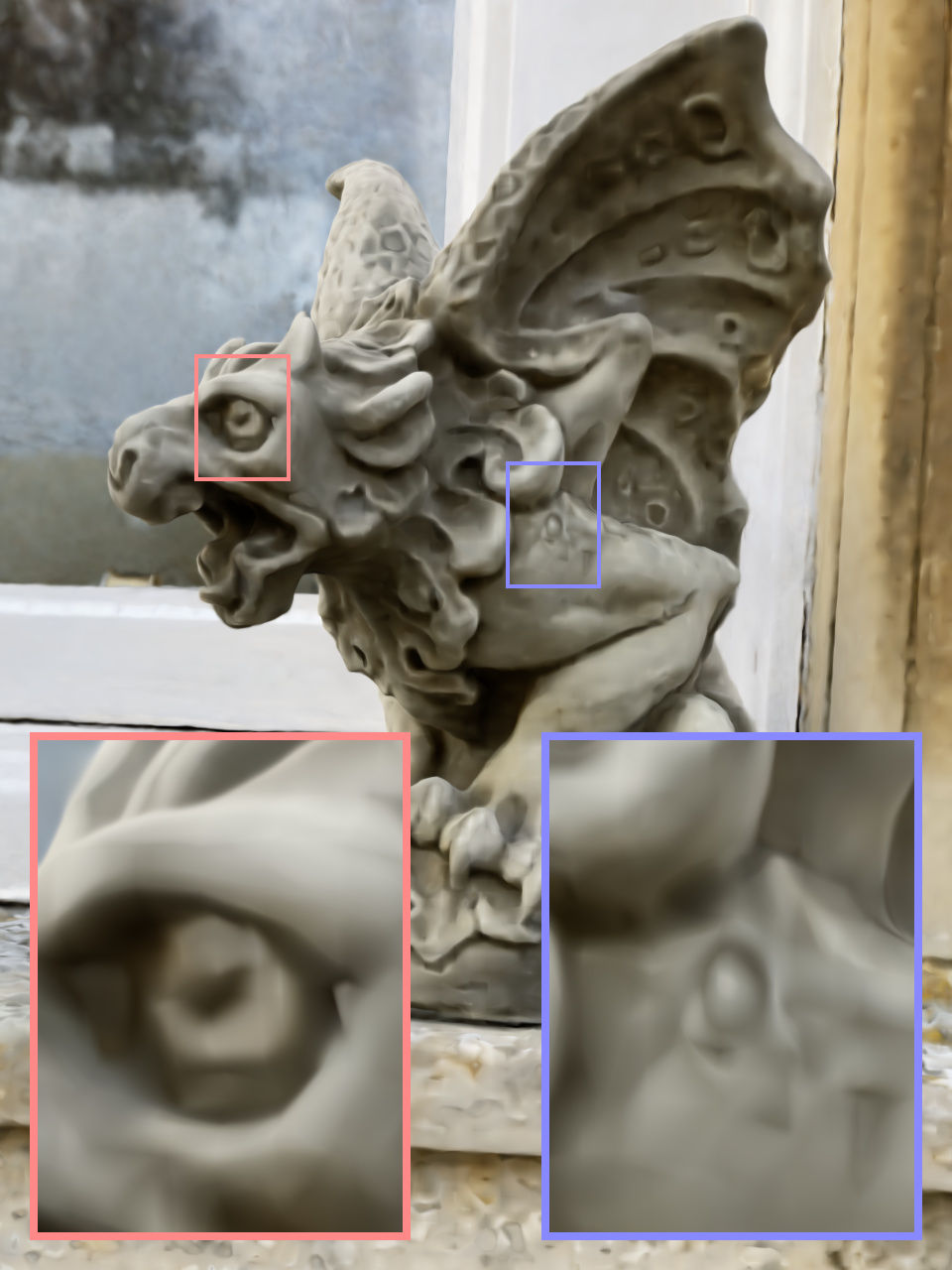} &
	\includegraphics[width=0.16\linewidth]{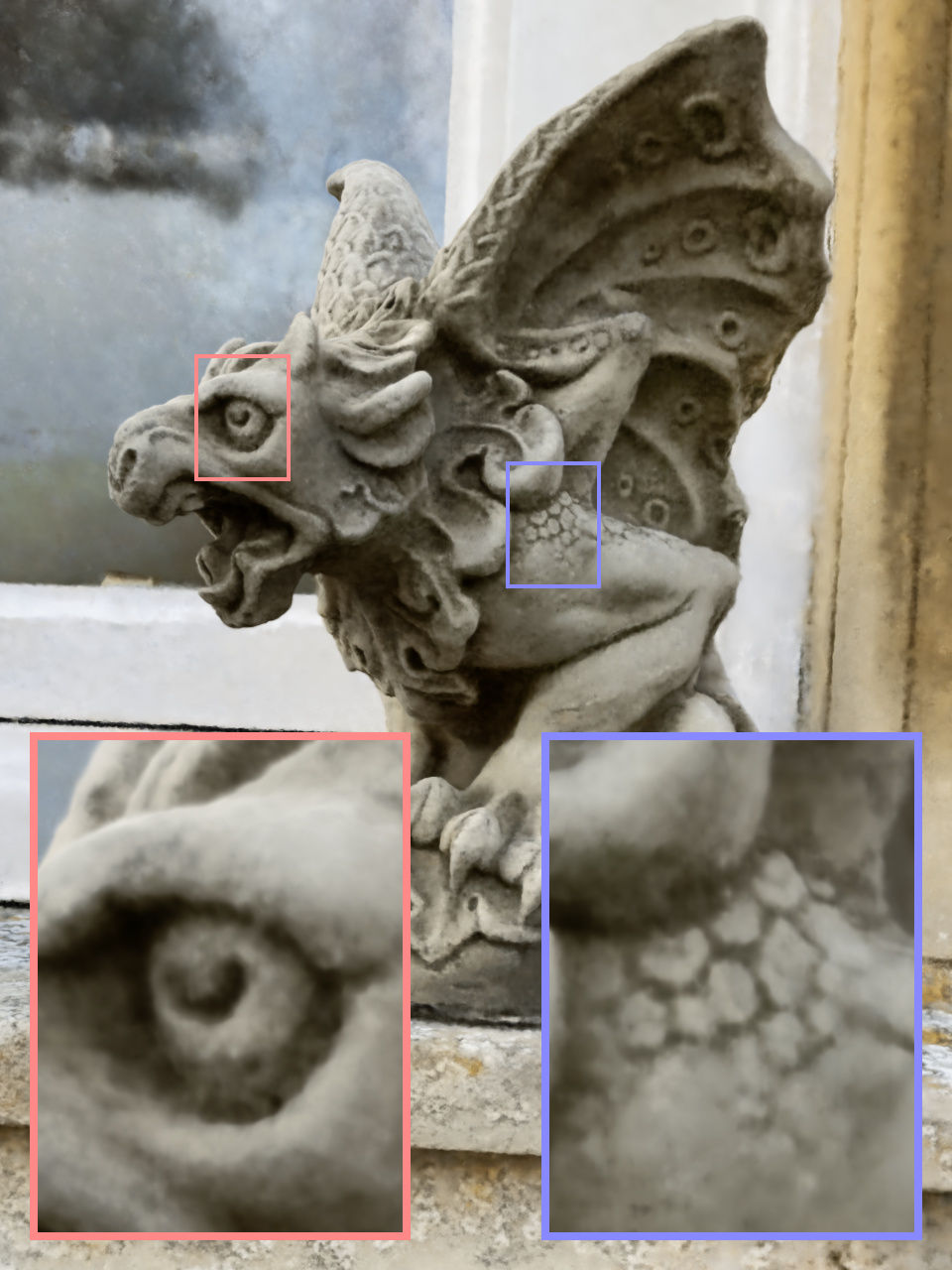} &
	\includegraphics[width=0.16\linewidth]{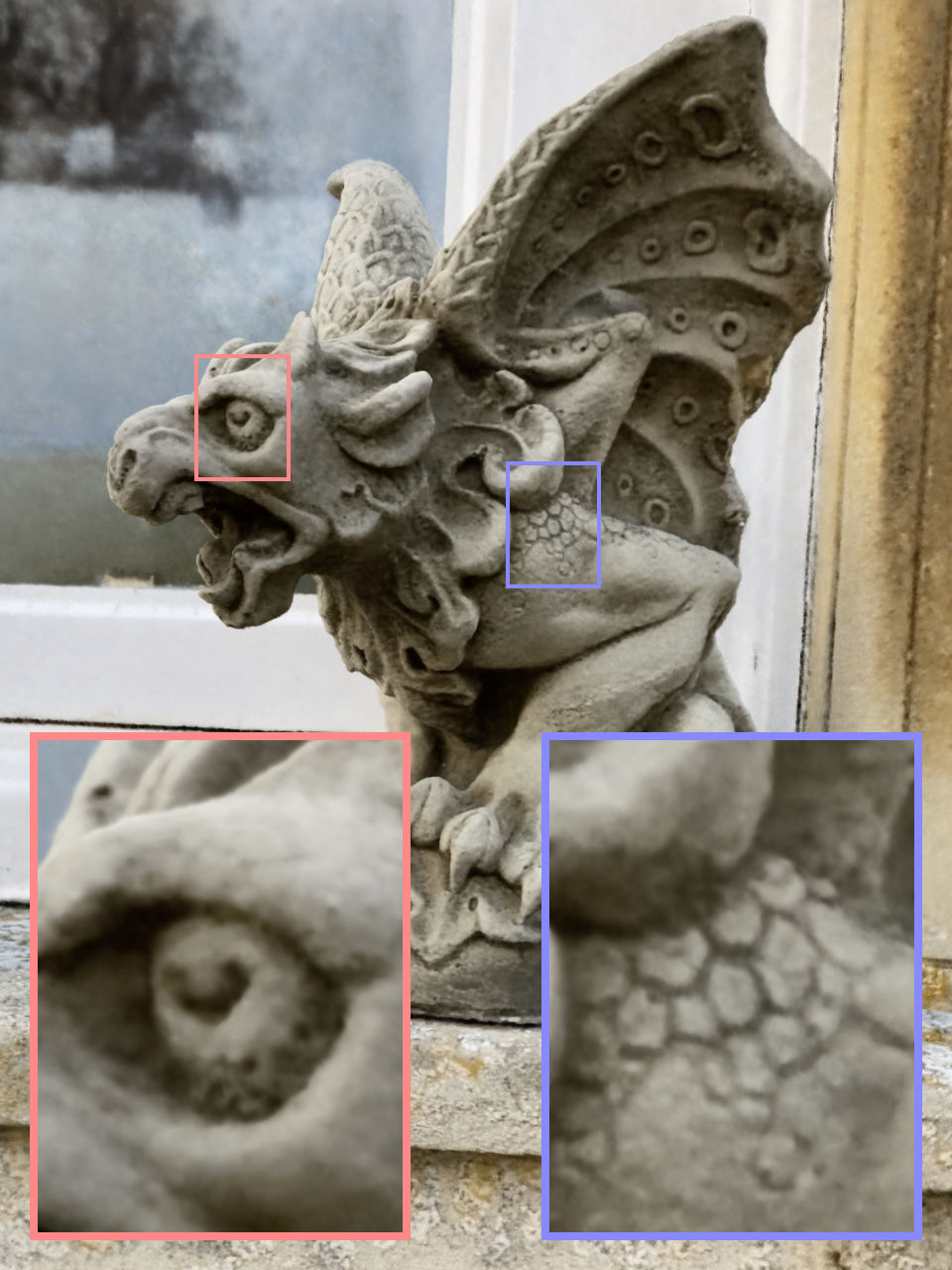} \\

	\SI{411}{\kilo\nothing} + 0 parameters &
	\SI{438}{\kilo\nothing} + 0 &
	\SI{10}{\kilo\nothing} + \SI{33.6}{\mega\nothing} &
	\SI{10}{\kilo\nothing} + \SI{16.3}{\mega\nothing} &
	\SI{10}{\kilo\nothing} + \SI{494}{\kilo\nothing} &
	\SI{10}{\kilo\nothing} + \SI{12.6}{\mega\nothing} \\
	11:28 (mm:ss) / PSNR 18.56 &
	12:45 / PSNR 22.90  &
	1:09 / PSNR 22.35 &
	1:26 / PSNR 23.62 &
	1:48 / PSNR 22.61 &
	1:47 / PSNR 24.58 \\
\end{tabular}

  \vspace{-2mm}
  \caption{\label{fig:grids}A demonstration of the reconstruction quality of different encodings and parametric data structures for storing trainable feature embeddings. Each configuration was trained for \num{11000} steps using our fast NeRF implementation (\autoref{Sec:Experiments:nerf}), varying only the input encoding \ADD{and MLP size}. The number of trainable parameters (MLP weights + encoding parameters), training time and reconstruction accuracy (PSNR) are shown below each image. Our encoding \textbf{(e)} with a similar total number of trainable parameters as the frequency encoding configuration \textbf{(b)} trains over $8\times$ faster, due to the sparsity of updates to the parameters and smaller MLP.\@ Increasing the number of parameters \textbf{(f)} further improves reconstruction accuracy without significantly increasing training time.}\vspace{-3mm}
\end{figure*}

\section{Background and Related Work} \label{Sec:RelatedWork}

Early examples of encoding the inputs of a machine learning model into a higher-dimensional space include the one-hot encoding~\cite{HARRIS201354} and the kernel trick~\citep{patternrecognition} by which complex arrangements of data can be made linearly separable.

For neural networks, input encodings have proven useful in the attention components of recurrent architectures~\cite{gehring2017convolutional} and, subsequently, transformers~\cite{vaswani2017attention}, where they help the neural network to identify the location it is currently processing.
Vaswani et al.~\shortcite{vaswani2017attention} encode scalar positions $x \in \R$ as a multiresolution sequence of ${\levels \in \mathbb{N}}$ sine and cosine functions
\begin{align}
  \enc(x) = \big( &\sin(2^0 x),  \sin(2^1 x),  \ldots, \sin(2^{\levels-1} x), \nonumber \\
  &\cos(2^0 x), \cos(2^1 x), \ldots, \cos(2^{\levels-1} x) \, \big) \,.
\end{align}
This has been adopted in computer graphics to encode the spatio-directionally varying light field and volume density in the NeRF algorithm~\cite{mildenhall2020nerf}.
The five dimensions of this light field are \emph{independently} encoded using the above formula; this was later extended to randomly oriented parallel wavefronts~\cite{tancik2020fourfeat} and level-of-detail filtering~\cite{barron2021mipnerf}.
We will refer to this family of encodings as \emph{frequency encodings}.
Notably, frequency encodings followed by a linear transformation have been used in other computer graphics tasks, such as approximating the visibility function~\citep{jansen2010fourier,annen2007convolutional}.

M\"uller et al.~\shortcite{mueller2019nis,mueller2020neural} suggested a continuous variant of the one-hot encoding based on rasterizing a kernel, the \emph{one-blob} encoding, which can achieve more accurate results than frequency encodings in bounded domains at the cost of being single-scale.

\paragraph{Parametric encodings.}
Recently, state-of-the-art results have been achieved by parametric encodings which blur the line between classical data structures and neural approaches.
\ADD{The idea is to arrange additional trainable parameters (beyond weights and biases) in an auxiliary data structure, such as a grid~\cite{liu2020neural,chabra2020,chiyu2020,peng2020,mehta2021modulated,sun2021direct,yu2021plenoxels,compstream4d}} or a tree~\cite{takikawa2021nglod}, and to look-up and (optionally) interpolate these parameters depending on the input vector ${\pos \in \R^d}$. This arrangement trades a larger memory footprint for a smaller computational cost: whereas for each gradient propagated backwards through the network, every weight in the fully connected MLP network must be updated, for the trainable input encoding parameters (``feature vectors''), only a very small number are affected. For example, with a trilinearly interpolated 3D grid of feature vectors, only 8 such grid points need to be updated for each sample back-propagated to the encoding. In this way, although the total number of parameters is much higher for a parametric encoding than a fixed input encoding, the number of FLOPs and memory accesses required for the update during training is not increased significantly. By reducing the size of the MLP, such parametric models can typically be trained to convergence much faster without sacrificing approximation quality.

Another parametric approach uses a tree subdivision of the domain $\R^d$, wherein a large auxiliary \emph{coordinate encoder} neural network (ACORN)~\citep{martel2021acorn} is trained to output dense feature grids in the leaf node around $\pos$.
These dense feature grids, which have on the order of \num{10000} entries, are then linearly interpolated, as in \citet{liu2020neural}.
This approach tends to yield a larger degree of adaptivity compared with the previous parametric encodings, albeit at greater computational cost which can only be amortized when sufficiently many inputs $\pos$ fall into each leaf node.

\paragraph{Sparse parametric encodings.}
While existing parametric encodings tend to yield much greater accuracy than their non-parametric predecessors, they also come with downsides in efficiency and versatility. Dense grids of trainable features consume much more memory than the neural network weights. To illustrate the trade-offs and to motivate our method, \autoref{fig:grids} shows the effect on reconstruction quality of a neural radiance field for several different encodings. Without any input encoding at all \textbf{(a)}, the network is only able to learn a fairly smooth function of position, resulting in a poor approximation of the light field. The frequency encoding \textbf{(b)} allows the same moderately sized network (8 hidden layers, each 256 wide) to represent the scene much more accurately. The middle image \textbf{(c)} pairs a smaller network with a \ADD{dense grid of $128^3$ trilinearly interpolated, 16-dimensional feature vectors}, for a total of 33.6 million trainable parameters. The large number of trainable parameters can be efficiently updated, as each sample only affects 8 grid points. 

However, the dense grid is wasteful in two ways. First, it allocates as many features to areas of empty space as it does to those areas near the surface.
The number of parameters grows as $\BigO(N^3)$, while the visible surface of interest has surface area that grows only as $\BigO(N^2)$.
In this example, the grid has resolution $128^3$, but only \num{53807} $(2.57\%)$ of its cells touch the visible surface. 

Second, natural scenes exhibit smoothness, motivating the use of a multi-resolution decomposition \cite{hadadan2021, chibane2020}.
\autoref{fig:grids}~\textbf{(d)} shows the result of using an encoding in which interpolated features 
are stored in eight co-located grids with resolutions from $16^3$ to $173^3$, \ADD{each containing 2-dimensional feature vectors.
These are concatenated to form a 16-dimensional (same as \textbf{(c)}) input to the network.}
Despite having less than half the number of parameters as \textbf{(c)}, the reconstruction quality is similar.

If the surface of interest is known a~priori, a data structure such as an octree \cite{takikawa2021nglod} or sparse grid \cite{liu2020neural,chabra2020,chiyu2020,peng2020,hadadan2021,chibane2020} can be used to cull away the unused features in the dense grid.
However, in the NeRF setting, surfaces only emerge during training.
NSVF~\cite{liu2020neural} and several concurrent works~\citep{yu2021plenoxels,sun2021direct} adopt a multi-stage, coarse to fine strategy in which regions of the feature grid are progressively refined and culled away as necessary.
While effective, this leads to a more complex training process in which the sparse data structure must be periodically updated.

Our method---\autoref{fig:grids}~\textbf{(e,f)}---combines both ideas to reduce waste.
\ADD{We store the trainable feature vectors in a compact spatial hash table, whose size is a hyper-parameter $\entriesPerLevel$ which can be tuned to trade the number of parameters for reconstruction quality.
It neither relies on progressive pruning during training nor on a~priori knowledge of the geometry of the scene.}
Analogous to the multi-resolution grid in \textbf{(d)}, we use multiple separate hash tables indexed at different resolutions, whose interpolated outputs are concatenated before being passed through the MLP.\@
The reconstruction quality is comparable to the dense grid encoding, despite having ${20\times}$ fewer parameters.

Unlike prior work that used spatial hashing~\citep{SpatialHash:03} for 3D reconstruction~\citep{niessner2013reconstruction}, we do not explicitly handle collisions of the hash functions by typical means like probing, bucketing, or chaining.
Instead, we rely on the neural network to learn to disambiguate hash collisions itself, avoiding control flow divergence, reducing implementation complexity and improving performance.
Another performance benefit is the predictable memory layout of the hash tables that is independent of the data that is represented. While good caching behavior is often hard to achieve with tree-like data structures, our hash tables can be fine-tuned for low-level architectural details such as cache size.

\begin{figure*}
  \centering
  \small\sffamily
  \vspace{1mm}
    \begin{overpic}[width=0.98\linewidth]{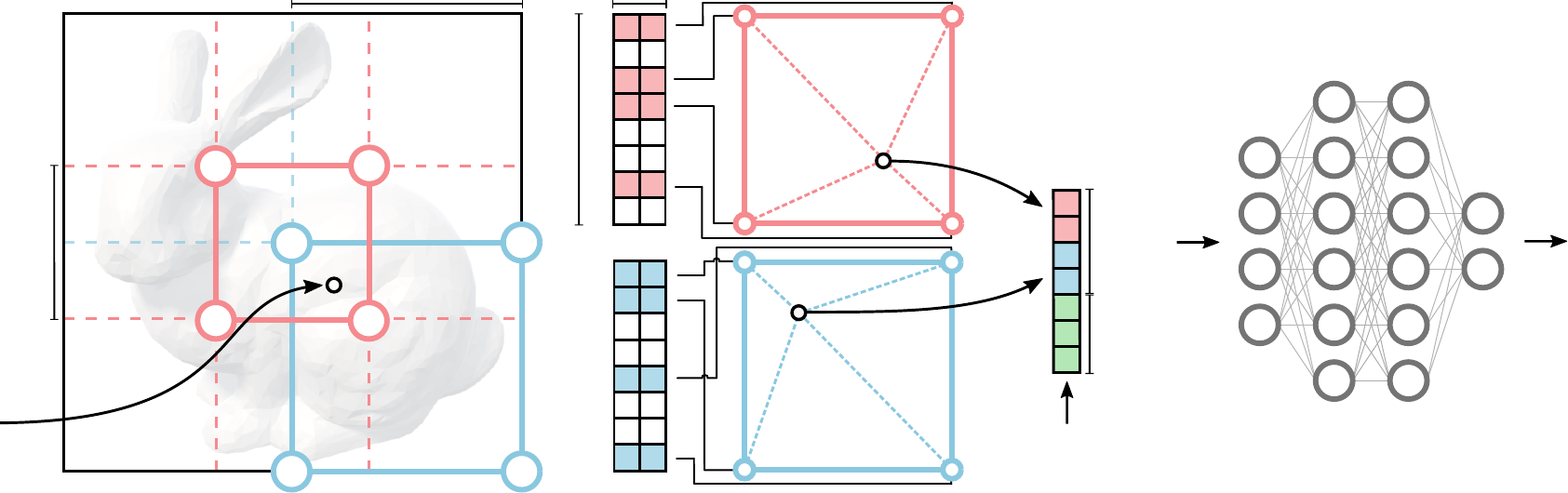}
    \put(-2,4.1) { $\pos$ }
    \put(67.2,20.2) { $\encOut$ }
    \put(84.3,27.6) { $\nn(\encOut; \Phi)$ }
    \put(34.8,23.2) { $\entriesPerLevel$ }
    \put(39.8,31.7) { $\featuresPerEntry$ }
    \put(69.7,15.4) { ${\levels \cdot \featuresPerEntry}$ }
    \put(69.7,9.8) { ${\numAuxDims}$ }
    \put(67.2,2.5) { ${\auxDims}$ }
    \put(4,31) { ${\levels = 2,\,\,\,\perLevelScale = 1.5}$ }
    \put(-0.9,15.5) { ${1 / \resolution_1}$ }
    \put(23.5,31.8) { ${1 / \resolution_0}$ }
    \put(27.7,16.35) { \textcolor{hgblue}{${l=0}$} }
    \put(18.5,21.25) { \textcolor{hgred}{${l=1}$} }
    \put(37.4,29.3) { $0$ }
    \put(37.4,27.6) { $1$ }
    \put(37.4,25.9) { $2$ }
    \put(37.4,24.2) { $3$ }
    \put(37.4,22.5) { $4$ }
    \put(37.4,20.8) { $5$ }
    \put(37.4,19.1) { $6$ }
    \put(37.4,17.4) { $7$ }
    \put(37.4,13.6) { $0$ }
    \put(37.4,11.9) { $1$ }
    \put(37.4,10.2) { $2$ }
    \put(37.4,8.5) { $3$ }
    \put(37.4,6.8) { $4$ }
    \put(37.4,5.1) { $5$ }
    \put(37.4,3.4) { $6$ }
    \put(37.4,1.7) { $7$ }
    \put(22.75,20.3) { $0$ }
    \put(12.95,20.3) { $2$ }
    \put(12.95,10.45) { $3$ }
    \put(22.75,10.45) { $6$ }
    \put(17.8,15.4) { $0$ }
    \put(32.5,15.4) { $4$ }
    \put(17.8,0.85) { $1$ }
    \put(32.5,0.85) { $7$ }
    \put(8, -3) { \small \textbf{(1)} Hashing of voxel vertices }
    \put(36, -3) { \small \textbf{(2)} Lookup }
    \put(45.3, -3) { \small \textbf{(3)} Linear interpolation }
    \put(63, -3) { \small \textbf{(4)} Concatenation }
    \put(80, -3) { \small \textbf{(5)} Neural network }
  \end{overpic}
  \vspace{5mm}
  \caption{\label{fig:encoding}%
    Illustration of the multiresolution hash encoding in $2$D.
    \textbf{(1)} for a given input coordinate $\pos$, we find the surrounding voxels at $\levels$ resolution levels and assign indices to their corners by hashing their integer coordinates.
    \textbf{(2)} for all resulting corner indices, we look up the corresponding $\featuresPerEntry$-dimensional feature vectors from the hash tables $\Params_\level$ and \textbf{(3)} linearly interpolate them according to the relative position of $\pos$ within the respective $l$-th voxel.
    \textbf{(4)} we concatenate the result of each level, as well as auxiliary inputs ${\auxDims \in \R^\numAuxDims}$, producing the encoded MLP input ${y \in \R^{\levels\featuresPerEntry + \numAuxDims}}$, which \textbf{(5)} is evaluated last.
    To train the encoding, loss gradients are backpropagated through the MLP \textbf{(5)}, the concatenation \textbf{(4)}, the linear interpolation \textbf{(3)}, and then accumulated in the looked-up feature vectors.
  }\vspace{-2mm}
\end{figure*}
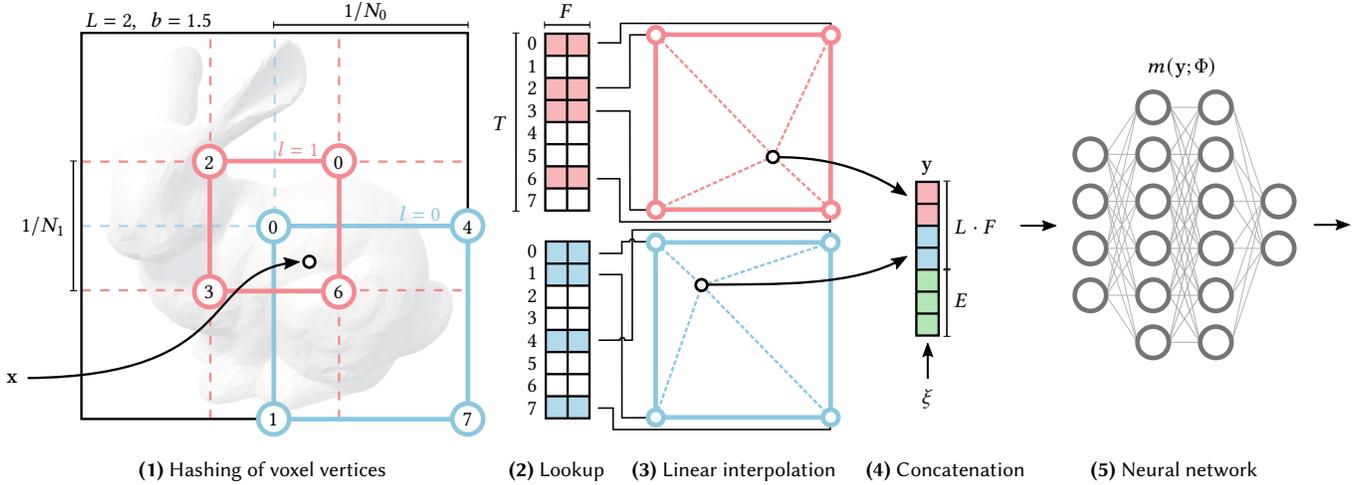

\section{Multiresolution Hash Encoding}
\label{Sec:Algorithm}

Given a fully connected neural network $\nn(\encOut; \Phi)$, we are interested in an \emph{encoding} of its inputs ${\encOut = \enc(\pos; \Params)}$ that improves the approximation quality and training speed across a wide range of applications without incurring a notable performance overhead.
\begin{table}
  \small
  \vspace{1mm}
  \caption{\label{tab:params}Hash encoding parameters and their ranges in our results. Only the hash table size $\entriesPerLevel$ and max.\ resolution $\maxResolution$ need to be tuned to the task.}%
  \vspace{-4mm}
  \newcolumntype{R}{>{\raggedleft\arraybackslash}X}
  \begin{tabularx}{\linewidth}{llR}%
    \toprule%
    Parameter & Symbol & Value \\
    \midrule%
    Number of levels & $\levels$ & 16 \\
    Max.\ entries per level (hash table size) & $\entriesPerLevel$ & $2^{14}$ to $2^{24}$ \\            
    Number of feature dimensions per entry & $\featuresPerEntry$ & 2 \\
    Coarsest resolution & $\minResolution$ & $16$ \\
    Finest resolution & $\maxResolution$ & $512$ to $524288$ \\
    \bottomrule%
  \end{tabularx}
  \vspace{-3mm}
\end{table}
Our neural network not only has trainable weight parameters $\Phi$, but also trainable encoding parameters $\Params$.
These are arranged into $\levels$ levels, each containing up to $\entriesPerLevel$ feature vectors with dimensionality $\featuresPerEntry$.
Typical values for these hyperparameters are shown in \autoref{tab:params}. 
\autoref{fig:encoding} illustrates the steps performed in our multiresolution hash encoding. Each level (two of which are shown as red and blue in the figure) is independent and conceptually stores feature vectors at the vertices of a grid, the resolution of which is chosen to be a geometric progression between the coarsest and finest resolutions $[\minResolution, \maxResolution]$:
\begin{align}
  \resolution_\level &:= \left\lfloor \minResolution \cdot \perLevelScale^\level \right\rfloor \,,\\
  \perLevelScale &:= \exp\left( \frac{\ln{\maxResolution} - \ln{\minResolution}}{\levels-1} \right) \,.\label{Eqn:PerLevelScale}
\end{align}
${\maxResolution}$ is chosen to match the finest detail in the training data. Due to the large number of levels $\levels$, the growth factor is usually small. Our use cases have \ADD{${\perLevelScale \in [1.26, 2]}$}.

Consider a single level $\level$.
The input coordinate $\pos \in \R^d$ is scaled by that level's grid resolution before rounding down and up $\lfloor \pos_\level\rfloor := \lfloor \pos \cdot \resolution_\level \rfloor$, $\lceil \pos_\level\rceil := \lceil \pos \cdot \resolution_\level \rceil$.

${\lfloor \pos_\level\rfloor}$ and ${\lceil \pos_\level\rceil}$ span a voxel with $2^d$ integer vertices in $\Z^d$. 
We map each corner to an entry in the level's respective feature vector array, which has fixed size of at most $\entriesPerLevel$.
For coarse levels where a dense grid requires fewer than $\entriesPerLevel$ parameters, i.e.\ \ADD{${(\resolution_\level + 1)^d \leq \entriesPerLevel}$}, this mapping is 1:1.
At finer levels, we use a hash function ${\hash : \Z^d \rightarrow \Z_\entriesPerLevel}$ to index into the array, effectively treating it as a hash table, although there is no explicit collision handling.
We rely instead on the gradient-based optimization to store appropriate sparse detail in the array, and the subsequent neural network $\nn(\encOut; \Phi)$ for collision resolution.
The number of trainable encoding parameters $\Params$ is therefore $\BigO(\entriesPerLevel)$ and bounded by ${\entriesPerLevel \cdot \levels \cdot \featuresPerEntry}$ which in our case is always ${\entriesPerLevel \cdot 16 \cdot 2}$ (\autoref{tab:params}).

We use a spatial hash function~\cite{SpatialHash:03} of the form
\begin{equation} \label{Eqn:HashFunc}
  h(\pos) = \left(\bigoplus_{i=1}^{d} x_i \primeNumber_i \right) \mod \entriesPerLevel \,,
\end{equation}
where $\oplus$ denotes the bit-wise XOR operation and $\primeNumber_i$ are unique, large prime numbers.
Effectively, this formula XORs the results of a per-dimension linear congruential (pseudo-random) permutation~\citep{Lehm51}, \emph{decorrelating} the effect of the dimensions on the hashed value.
Notably, to achieve (pseudo-)independence, only ${d-1}$ of the $d$ dimensions must be permuted, so we choose ${\primeNumber_1 := 1}$ for better cache coherence, ${\primeNumber_2 = \num{2654435761}}$, and ${\primeNumber_3 = \num{805459861}}$.

Lastly, the feature vectors at each corner are $d$-linearly interpolated according to the relative position of $\pos$ within its hypercube, i.e.\ the interpolation weight is $\interpWeight_\level := \pos_\level - \lfloor \pos_\level\rfloor$.

Recall that this process takes place independently for each of the $\levels$ levels.
The interpolated feature vectors of each level, as well as auxiliary inputs ${\auxDims \in \R^\numAuxDims}$ (such as the encoded view direction and textures in neural radiance caching), are concatenated to produce ${\encOut \in \R^{\levels\featuresPerEntry + \numAuxDims}}$, which is the encoded input $\enc(\pos; \Params)$ to the MLP $\nn(\encOut; \Phi)$.

\begin{figure*}
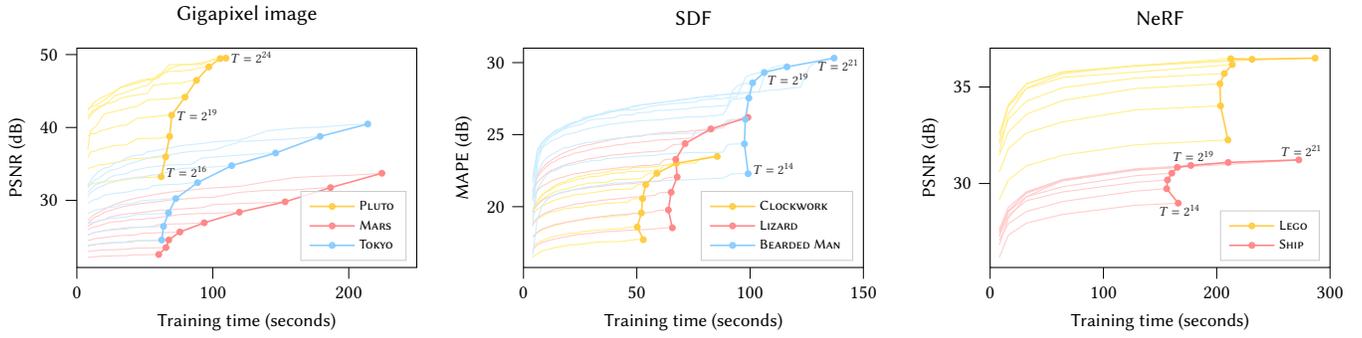

  \sffamily
  \small
  \vspace{-5mm}
  \begin{center}
  \pgfplotsset{width=6.1cm, height=4.5cm}
  \input{Figures/T_sweep_image.tex}
  \hfill
  \input{Figures/T_sweep_sdf.tex}
  \hfill
  \begin{tikzpicture}
\tikzstyle{every node}=[font=\footnotesize]

\definecolor{color0}{rgb}{1,0.933333333333333,0.533333333333333}
\definecolor{color1}{rgb}{1,0.8,0.266666666666667}
\definecolor{color2}{rgb}{1,0.8,0.8}
\definecolor{color3}{rgb}{1,0.533333333333333,0.533333333333333}

\begin{axis}[
legend cell align={left},
legend style={
  fill opacity=0.8,
  draw opacity=1,
  text opacity=1,
  at={(0.97,0.03)},
  anchor=south east,
  draw=white!80!black,
  nodes={scale=0.75, transform shape}
},
tick align=outside,
tick pos=left,
title={{\small \sffamily NeRF}},
x grid style={white!69.0196078431373!black},
xlabel={Training time (seconds)},
xmin=0, xmax=300,
xtick style={color=black},
y grid style={white!69.0196078431373!black},
ylabel={PSNR (dB)},
ymin=25.6499456781147, ymax=37.0122839688719,
ytick style={color=black}
]
\addplot [color0, forget plot]
table {%
8.09373307228088 29.1630324265388
16.0323576927185 30.1869619636652
32.0793335437775 30.9194909199443
64.0897214412689 31.4660465046261
128.050508975983 32.0023977129052
209.925636529922 32.2576160987412
209.925636529922 32.2576160987412
};
\addplot [color0, forget plot]
table {%
8.05036950111389 30.5829734874412
16.0552365779877 31.8097062279288
32.0132322311401 32.6190006214372
64.0039558410645 33.1917639688894
128.053673028946 33.8139810723941
203.399200439453 34.023771217675
203.399200439453 34.023771217675
};
\addplot [color0, forget plot]
table {%
8.07111978530884 31.4472086639694
16.0904173851013 32.6627072805105
32.047515630722 33.5928906164644
64.0934362411499 34.3038211925948
128.016080856323 34.9177572782646
202.856100082397 35.1648293771629
202.856100082397 35.1648293771629
};
\addplot [color0, forget plot]
table {%
8.0594277381897 32.0591491495519
16.0798299312592 33.3277181645991
32.0380666255951 34.2672174413608
64.0092711448669 34.963299715008
128.013519048691 35.4620966724652
206.719816923141 35.6970455781236
206.719816923141 35.6970455781236
};
\addplot [color0, forget plot]
table {%
8.00361490249634 32.5786416616835
16.0490343570709 34.0058775976134
32.0848748683929 34.9153684597577
64.0436646938324 35.4946136381959
128.013769865036 35.7728619666011
213.731005907059 36.1631343913293
213.731005907059 36.1631343913293
};
\addplot [color0, forget plot]
table {%
8.10317873954773 32.5047332009673
16.0188596248627 34.0868741931285
32.0533757209778 35.176583194627
64.1085493564606 35.7942572240713
128.022009849548 36.1025223452177
212.402442455292 36.4651391639235
212.402442455292 36.4651391639235
};
\addplot [color0, forget plot]
table {%
8.05712747573853 32.1870244909271
16.0118291378021 34.0216532082277
32.0686717033386 35.1557546472977
64.0784502029419 35.7493364610637
128.105428695679 36.0740891126509
231.250281333923 36.4361386595409
231.250281333923 36.4361386595409
};
\addplot [color0, forget plot]
table {%
8.11955189704895 31.6324674883997
16.0933399200439 33.6102921093532
32.0556244850159 34.9455946974619
64.1094555854797 35.6874013973242
128.052837371826 36.1176623542034
256.015993118286 36.4594158185562
286.843525409698 36.4958140465648
286.843525409698 36.4958140465648
};
\addplot [color2, forget plot]
table {%
8.00382733345032 26.1664156004219
16.0448417663574 27.3219508865868
32.016988992691 27.9492193901079
64.0773961544037 28.4013485357688
128.039295911789 28.872344134916
165.986980438232 28.98228156006
165.986980438232 28.98228156006
};
\addplot [color2, forget plot]
table {%
8.02896070480347 27.0605117977212
16.0220813751221 27.9171037039622
32.0104048252106 28.5502200721684
64.0719561576843 29.0688429042092
128.040958166122 29.6071820493163
155.838763475418 29.7313884707525
155.838763475418 29.7313884707525
};
\addplot [color2, forget plot]
table {%
8.0210063457489 27.2633647894168
16.0814518928528 28.3555167024447
32.0237545967102 29.1094563253608
64.0067400932312 29.564142478224
128.074329376221 30.0598807981077
156.553154706955 30.1836625529605
156.553154706955 30.1836625529605
};
\addplot [color2, forget plot]
table {%
8.05302786827087 27.4575342003025
16.042062997818 28.529433011808
32.0334355831146 29.3158297822504
64.0607149600983 29.8369505689026
128.006455421448 30.4011323478128
160.364272117615 30.5361457784287
160.364272117615 30.5361457784287
};
\addplot [color2, forget plot]
table {%
8.00056290626526 27.5309847772503
16.0632312297821 28.598871740575
32.0704576969147 29.496787405644
64.000551700592 30.1342594796264
128.042566299438 30.7223571046598
165.250351428986 30.8368975739295
165.250351428986 30.8368975739295
};
\addplot [color2, forget plot]
table {%
8.02212238311768 27.6151543687259
16.0327434539795 28.7652645674358
32.0243892669678 29.5707955754572
64.0231027603149 30.2108527296501
128.024807214737 30.762677414132
177.146847486496 30.9365102853127
177.146847486496 30.9365102853127
};
\addplot [color2, forget plot]
table {%
8.04814767837524 27.29296559467
16.0955483913422 28.5749173486931
32.0147733688354 29.538276107088
64.1098403930664 30.2006946469527
128.027252912521 30.6872526344788
210.114523410797 31.0907662102288
210.114523410797 31.0907662102288
};
\addplot [color2, forget plot]
table {%
8.01872253417969 26.817479716824
16.0701494216919 28.2116937840877
32.0457403659821 29.3189337998601
64.0445663928986 30.1569750200887
128.129016160965 30.6979029312843
256.076171398163 31.153988868157
272.455973863602 31.2246186589718
272.455973863602 31.2246186589718
};
\addplot [semithick, color1, mark=*, mark size=1, mark options={solid}]
table {%
209.925636529922 32.2576160987412
203.399200439453 34.023771217675
202.856100082397 35.1648293771629
206.719816923141 35.6970455781236
213.731005907059 36.1631343913293
212.402442455292 36.4651391639235
231.250281333923 36.4361386595409
286.843525409698 36.4958140465648
};
\addlegendentry{\sceneLego}
\addplot [semithick, color3, mark=*, mark size=1, mark options={solid}]
table {%
165.986980438232 28.98228156006
155.838763475418 29.7313884707525
156.553154706955 30.1836625529605
160.364272117615 30.5361457784287
165.250351428986 30.8368975739295
177.146847486496 30.9365102853127
210.114523410797 31.0907662102288
272.455973863602 31.2246186589718
};
\addlegendentry{\sceneShip}
\draw (axis cs:144.693181753159,28.3095804303865) node[
  scale=0.7,
  anchor=base west,
  text=white!15!black,
  rotate=0.0
]{$\entriesPerLevel = 2^{14}$};
\draw (axis cs:156.917738735676,31.1383206242148) node[
  scale=0.7,
  anchor=base west,
  text=white!15!black,
  rotate=0.0
]{$\entriesPerLevel = 2^{19}$};
\draw (axis cs:252.226865112782,31.4264289978738) node[
  scale=0.7,
  anchor=base west,
  text=white!15!black,
  rotate=0.0
]{$\entriesPerLevel = 2^{21}$};
\end{axis}

\end{tikzpicture}
  \end{center}
  \vspace{-3mm}
  \caption{\label{fig:image_t_sweep}%
  The main curves plot test error over training time for varying hash table size $\entriesPerLevel$ which determines the number of trainable encoding parameters. Increasing $\entriesPerLevel$ improves reconstruction, at the cost of higher memory usage and slower training and inference. A performance cliff is visible at ${\entriesPerLevel > 2^{19}}$ where the cache of our RTX 3090 GPU becomes oversubscribed (particularly visible for SDF and NeRF). The plot also shows model convergence over time leading up to the final state. This highlights how high quality results are already obtained after only a few seconds. Jumps in the convergence (most visible towards the end of SDF training) are caused by learning rate decay. For NeRF and Gigapixel image, training finishes after \num{31000} steps and for SDF after \num{11000} steps. 
  }
\end{figure*}

\begin{figure*}
  \sffamily
  \small
  \vspace{-1mm}
  \begin{center}
  \pgfplotsset{width=6.1cm, height=4cm}
  \begin{tikzpicture}
\tikzstyle{every node}=[font=\footnotesize]

\definecolor{color0}{rgb}{1,0.8,0.266666666666667}
\definecolor{color1}{rgb}{1,0.533333333333333,0.533333333333333}
\definecolor{color2}{rgb}{0.533333333333333,0.8,1}
\definecolor{color3}{rgb}{0.4,0.933333333333333,0.4}

\begin{axis}[
legend cell align={left},
legend style={
  fill opacity=0.8,
  draw opacity=1,
  text opacity=1,
  at={(0.97,0.03)},
  anchor=south east,
  draw=white!80!black,
  nodes={scale=0.75, transform shape}
},
tick align=outside,
tick pos=left,
title={{\small \sffamily Gigapixel image: \sceneTokyo}},
x grid style={white!69.0196078431373!black},
xlabel={Training time (seconds)},
xmin=103.421799862385, xmax=464.261781728268,
xtick style={color=black},
y grid style={white!69.0196078431373!black},
ylabel={PSNR (dB)},
ymin=20.2067203398071, ymax=42.0816579539985,
ytick style={color=black}
]
\addplot [semithick, color0, mark=*, mark size=1, mark options={solid}]
table {%
228.929524421692 21.2010356859067
154.858687400818 29.0460624157928
181.510112047195 36.6448114475036
230.502841949463 39.1480011664734
321.705054044724 40.7173075832271
};
\addlegendentry{F=1}
\addplot [semithick, color1, mark=*, mark size=1, mark options={solid}]
table {%
144.989725351334 21.6963558819769
119.823617219925 32.692822251289
156.119407176971 38.6662794062705
219.022021532059 40.4591431395683
337.516640901566 41.0321198738862
};
\addlegendentry{F=2}
\addplot [semithick, color2, mark=*, mark size=1, mark options={solid}]
table {%
159.452895879745 22.474131419056
134.288972377777 36.0011806809374
181.115212202072 40.1872081125896
253.08021569252 40.8744902370282
397.49270939827 41.0873426078989
};
\addlegendentry{F=4}
\addplot [semithick, color3, mark=*, mark size=1, mark options={solid}]
table {%
154.657912731171 23.3712799933527
142.505985736847 38.1103421888565
202.064184904099 40.330309274246
288.764913082123 40.8040547987758
447.859964370728 40.8710797587731
};
\addlegendentry{F=8}
\draw (axis cs:233.568300902843,21.0058729669335) node[
  scale=0.7,
  anchor=base west,
  text=white!15!black,
  rotate=0.0
]{$\levels = 2$};
\draw (axis cs:159.497463881969,28.8508996968196) node[
  scale=0.7,
  anchor=base west,
  text=white!15!black,
  rotate=0.0
]{$\levels = 4$};
\draw (axis cs:186.148888528347,35.6689978526376) node[
  scale=0.7,
  anchor=base west,
  text=white!15!black,
  rotate=0.0
]{$\levels = 8$};
\draw (axis cs:235.141618430614,38.1721875716074) node[
  scale=0.7,
  anchor=base west,
  text=white!15!black,
  rotate=0.0
]{$\levels = 16$};
\draw (axis cs:319.849543452263,38.9608431124682) node[
  scale=0.7,
  anchor=base west,
  text=white!15!black,
  rotate=0.0
]{$\levels = 32$};
\end{axis}

\end{tikzpicture}
  \hfill
  \begin{tikzpicture}
\tikzstyle{every node}=[font=\footnotesize]

\definecolor{color0}{rgb}{1,0.8,0.266666666666667}
\definecolor{color1}{rgb}{1,0.533333333333333,0.533333333333333}
\definecolor{color2}{rgb}{0.533333333333333,0.8,1}
\definecolor{color3}{rgb}{0.4,0.933333333333333,0.4}

\begin{axis}[
legend cell align={left},
legend style={
  fill opacity=0.8,
  draw opacity=1,
  text opacity=1,
  at={(0.97,0.03)},
  anchor=south east,
  draw=white!80!black,
  nodes={scale=0.75, transform shape}
},
tick align=outside,
tick pos=left,
title={{\small \sffamily Signed Distance Function: \sceneCow}},
x grid style={white!69.0196078431373!black},
xlabel={Training time (seconds)},
xmin=51.1697432518005, xmax=102.985850143433,
xtick style={color=black},
y grid style={white!69.0196078431373!black},
ylabel={MAPE (dB)},
ymin=19.4124704903077, ymax=22.266776617726,
ytick style={color=black}
]
\addplot [semithick, color0, mark=*, mark size=1, mark options={solid}]
table {%
53.5250208377838 19.5422116779176
62.5882759094238 20.6477222263526
69.8922746181488 21.3563778278445
81.3737463951111 22.0737044451526
};
\addlegendentry{F=1}
\addplot [semithick, color1, mark=*, mark size=1, mark options={solid}]
table {%
57.3348355293274 19.8022916036004
58.1764385700226 20.9151951390589
64.0480613708496 21.8495327248969
81.8016700744629 22.1370354301161
};
\addlegendentry{F=2}
\addplot [semithick, color2, mark=*, mark size=1, mark options={solid}]
table {%
57.6387646198273 20.0459844079176
59.7291388511658 21.195460470352
69.8229157924652 21.8141258371745
95.2951185703278 21.9742470581402
};
\addlegendentry{F=4}
\addplot [semithick, color3, mark=*, mark size=1, mark options={solid}]
table {%
58.4805581569672 20.0707805516222
60.9406895637512 21.2177000059208
74.0750935077667 21.6497666358274
100.630572557449 22.0103486885545
};
\addlegendentry{F=8}
\draw (axis cs:58.4805581569672,20.0707805516222) node[
  scale=0.7,
  anchor=base west,
  text=white!15!black,
  rotate=0.0
]{$\levels = 4$};
\draw (axis cs:60.9406895637512,21.0237431922275) node[
  scale=0.7,
  anchor=base west,
  text=white!15!black,
  rotate=0.0
]{$\levels = 8$};
\draw (axis cs:74.0750935077667,21.4558098221342) node[
  scale=0.7,
  anchor=base west,
  text=white!15!black,
  rotate=0.0
]{$\levels = 16$};
\draw (axis cs:96.4155711174011,21.7194134680146) node[
  scale=0.7,
  anchor=base west,
  text=white!15!black,
  rotate=0.0
]{$\levels = 32$};
\end{axis}

\end{tikzpicture}
  \hfill
  \begin{tikzpicture}
\tikzstyle{every node}=[font=\footnotesize]

\definecolor{color0}{rgb}{1,0.8,0.266666666666667}
\definecolor{color1}{rgb}{1,0.533333333333333,0.533333333333333}
\definecolor{color2}{rgb}{0.533333333333333,0.8,1}
\definecolor{color3}{rgb}{0.4,0.933333333333333,0.4}

\begin{axis}[
legend cell align={left},
legend style={
  fill opacity=0.8,
  draw opacity=1,
  text opacity=1,
  at={(0.97,0.03)},
  anchor=south east,
  draw=white!80!black,
  nodes={scale=0.75, transform shape}
},
tick align=outside,
tick pos=left,
title={{\small \sffamily Neural Radiance Field: \sceneLego}},
x grid style={white!69.0196078431373!black},
xlabel={Training time (seconds)},
xmin=161.283349061012, xmax=506.544362044334,
xtick style={color=black},
y grid style={white!69.0196078431373!black},
ylabel={PSNR (dB)},
ymin=32.4409293696302, ymax=36.8024791114829,
ytick style={color=black}
]
\addplot [semithick, color0, mark=*, mark size=1, mark options={solid}]
table {%
176.977031469345 32.6391816306235
219.887970209122 34.9971521778585
247.183665275574 35.9605496985434
309.770037174225 36.3368775803035
};
\addlegendentry{F=1}
\addplot [semithick, color1, mark=*, mark size=1, mark options={solid}]
table {%
199.329049110413 34.3594561463158
200.676287412643 35.8455825108642
224.979252815247 36.3012873233936
320.418728590012 36.5261911390268
};
\addlegendentry{F=2}
\addplot [semithick, color2, mark=*, mark size=1, mark options={solid}]
table {%
192.913296699524 35.572185179762
213.924275636673 36.1229185152962
278.900064468384 36.5581807548899
407.25347161293 36.539436401952
};
\addlegendentry{F=4}
\addplot [semithick, color3, mark=*, mark size=1, mark options={solid}]
table {%
202.953203201294 35.6939234735322
233.376142501831 36.2753909092502
322.851065158844 36.4151717077674
490.850679636002 36.6042268504896
};
\addlegendentry{F=8}
\draw (axis cs:176.977031469345,32.6391816306235) node[
  scale=0.7,
  anchor=base west,
  text=white!15!black,
  rotate=0.0
]{$\levels = 4$};
\draw (axis cs:219.887970209122,34.7013365018841) node[
  scale=0.7,
  anchor=base west,
  text=white!15!black,
  rotate=0.0
]{$\levels = 8$};
\draw (axis cs:247.183665275574,35.664734022569) node[
  scale=0.7,
  anchor=base west,
  text=white!15!black,
  rotate=0.0
]{$\levels = 16$};
\draw (axis cs:309.770037174225,36.0410619043291) node[
  scale=0.7,
  anchor=base west,
  text=white!15!black,
  rotate=0.0
]{$\levels = 32$};
\end{axis}

\end{tikzpicture}
  \end{center}
  \vspace{-3mm}
  \caption{\label{fig:image_ftl_sweep}%
    Test error over training time for fixed values of feature dimensionality $\featuresPerEntry$ as the number of hash table levels $\levels$ is varied.
    To maintain a roughly equal trainable parameter count, the hash table size $\entriesPerLevel$ is set according to ${\featuresPerEntry \cdot \entriesPerLevel \cdot \levels = 2^{24}}$ for SDF and NeRF, whereas gigapixel image uses $2^{28}$.
    Since ${(F=2,\levels = 16)}$ is near the best-case performance and quality (top-left) for all applications, we use this configuration in all results.
    ${F=1}$ is slow on our RTX 3090 GPU since atomic half-precision accumulation is only efficient for 2D vectors but not for scalars.
    For NeRF and Gigapixel image, training finishes after \num{31000} steps whereas SDF completes at \num{11000} steps. 
  }\vspace{-2mm}
\end{figure*}
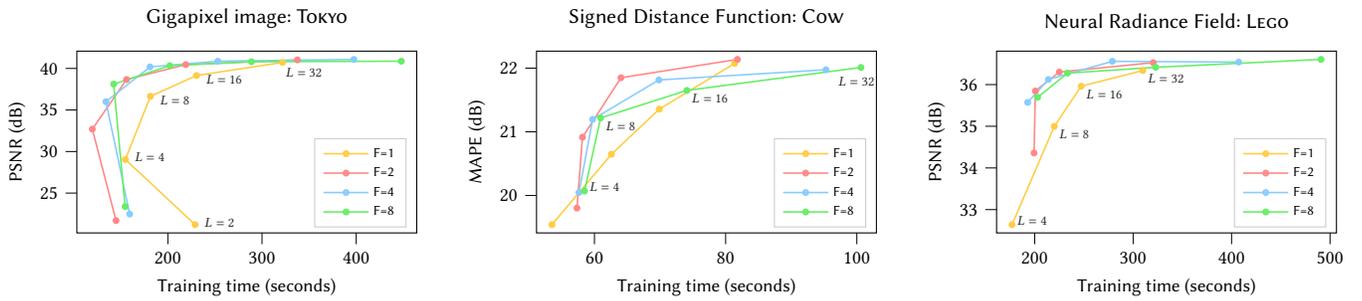

\paragraph{Performance vs.\ quality.}
Choosing the hash table size $\entriesPerLevel$ provides a trade-off between performance, memory and quality.
Higher values of $\entriesPerLevel$ result in higher quality and lower performance.
The memory footprint is linear in $\entriesPerLevel$, whereas quality and performance tend to scale sub-linearly.
We analyze the impact of $\entriesPerLevel$ in \autoref{fig:image_t_sweep}, where we report test error vs.\ training time for a wide range of $\entriesPerLevel$-values for three neural graphics primitives.
We recommend practitioners to use $\entriesPerLevel$ to tweak the encoding to their desired performance characteristics.

The hyperparameters $\levels$ (number of levels) and $\featuresPerEntry$ (number of feature dimensions) also trade off quality and performance, which we analyze for an approximately constant number of trainable encoding parameters $\Params$ in \autoref{fig:image_ftl_sweep}.
In this analysis, we found ${(\featuresPerEntry=2,\levels=16)}$ to be a favorable Pareto optimum in all our applications, so we use these values in all other results and recommend them as the default.

\paragraph{Implicit hash collision resolution.}
It may appear counter-intuitive that this encoding is able to reconstruct scenes faithfully in the presence of hash collisions. 
Key to its success is that the different resolution levels have different strengths that complement each other.
The coarser levels, and thus the encoding as a whole, are injective---that is, they suffer from no collisions at all.
However, they can only represent a low-resolution version of the scene, since they offer features which are linearly interpolated from a widely spaced grid of points.
Conversely, fine levels can capture small features due to their fine grid resolution, but suffer from many collisions---that is, disparate points which hash to the same table entry.
Nearby inputs with equal integer coordinates ${\lfloor \pos_\level\rfloor}$ are not considered a collision; a collision occurs when \emph{different} integer coordinates hash to the same index.
Luckily, such collisions are pseudo-randomly scattered across space, and statistically unlikely to occur \emph{simultaneously} at every level for a given pair of points.

When training samples collide in this way, their gradients average.
Consider that the importance to the final reconstruction of such samples is rarely equal.
For example, a point on a visible surface of a radiance field will contribute strongly to the reconstructed image (having high visibility and high density, both multiplicatively affecting the magnitude of gradients) causing large changes to its table entries, while a point in empty space that happens to refer to the same entry will have a much smaller weight.
As a result, the gradients of the more important samples dominate the collision average and the aliased table entry will naturally be optimized in such a way that it reflects the needs of the higher-weighted point.

\ADD{The multiresolution aspect of the hash encoding covers the full range from a coarse resolution $\minResolution$ that is guaranteed to be collision-free to the finest resolution $\maxResolution$ that the task requires.
Thereby, it guarantees that all scales at which meaningful learning \emph{could} take place are included, regardless of sparsity.
Geometric scaling allows covering these scales with only $\BigO\big(\log{(\maxResolution/\minResolution)}\big)$ many levels, which allows picking a conservatively large value for $\maxResolution$.}

\paragraph{Online adaptivity.}
Note that if the distribution of inputs $\pos$ changes over time during training, for example if they become concentrated in a small region, then finer grid levels will experience fewer collisions and a more accurate function can be learned.
In other words, the multiresolution hash encoding \emph{automatically} adapts to the training data distribution, inheriting the benefits of tree-based encodings~\cite{takikawa2021nglod} without task-specific data structure maintenance that might cause discrete jumps during training.
One of our applications, neural radiance caching in \autoref{Sec:Experiments:nrc}, continually adapts to animated viewpoints and 3D content, greatly benefitting from this feature.

\paragraph{$d$-linear interpolation.}
Interpolating the queried hash table entries ensures that the encoding $\enc(\pos; \Params)$, and by the chain rule its composition with the neural network $\nn(\enc(\pos; \Params); \Phi)$, are continuous.
Without interpolation, grid-aligned discontinuities would be present in the network output, which would result in an undesirable blocky appearance.
One may desire higher-order smoothness, for example when approximating partial differential equations.
A concrete example from computer graphics are signed distance functions, in which case the gradient $\partial\nn(\enc(\pos; \Params); \Phi) / \partial \pos$, i.e.\ the surface normal, would ideally also be continuous.
\ADD{If higher-order smoothness must be guaranteed, we describe a low-cost approach in \autoref{app:smooth-interpolation}, which we however do \emph{not} employ in any of our results due to a small decrease in reconstruction quality.}

\section{Implementation}\label{Sec:Implementation}

To demonstrate the speed of the multiresolution hash encoding, we implemented it in CUDA and integrated it with the fast fully-fused MLPs of the \emph{tiny-cuda-nn} framework~\cite{tiny-cuda-nn}.\footnote{\ADD{We observe speed-ups on the order of ${10\times}$ compared to a na\"{i}ve Python implementation.
We therefore also release PyTorch bindings around our hash encoding and fully fused MLPs to permit their use in existing projects with little overhead.}}
\ADD{We release the source code of the multiresolution hash encoding as an update to \citet{tiny-cuda-nn}
and the source code pertaining to the neural graphics primitives at
{\urlstyle{tt}\url{https://github.com/nvlabs/instant-ngp}}}.

\begin{figure*}
  \small
  \sffamily
  \setlength{\tabcolsep}{1pt}%
  \renewcommand{\arraystretch}{1}%
  \vspace{-2mm}
  \begin{tabular}{cccccc}  
   Hash table size: ${\entriesPerLevel=2^{22}}$  & ${\entriesPerLevel=2^{22}}$  & ${\entriesPerLevel=2^{12}}$ & ${\entriesPerLevel=2^{17}}$ & ${\entriesPerLevel=2^{22}}$ & Reference \\[-0.25mm]
    \makecell{\includegraphics[width=0.2465\linewidth]{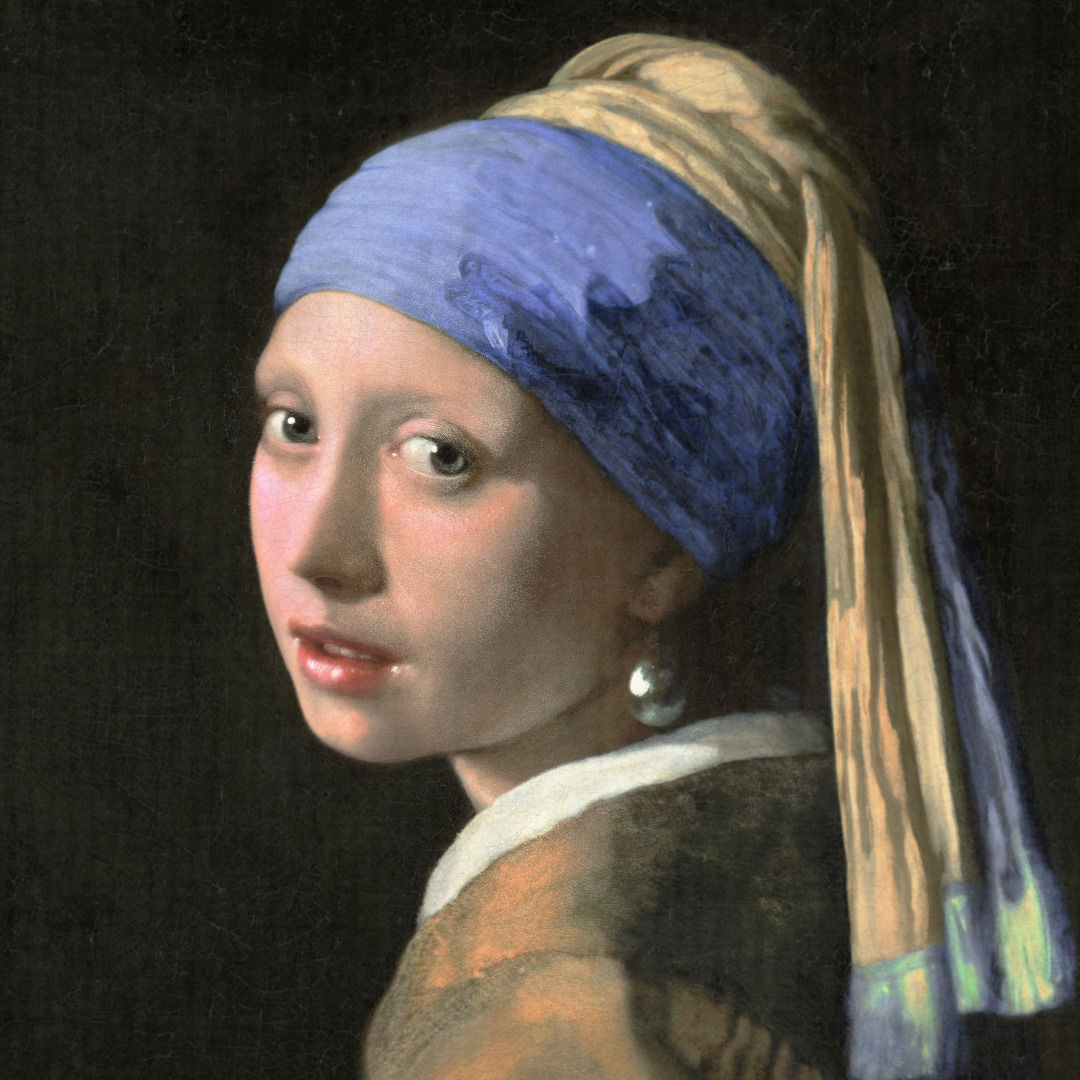}} &
    \makecell{\includegraphics[width=0.2465\linewidth]{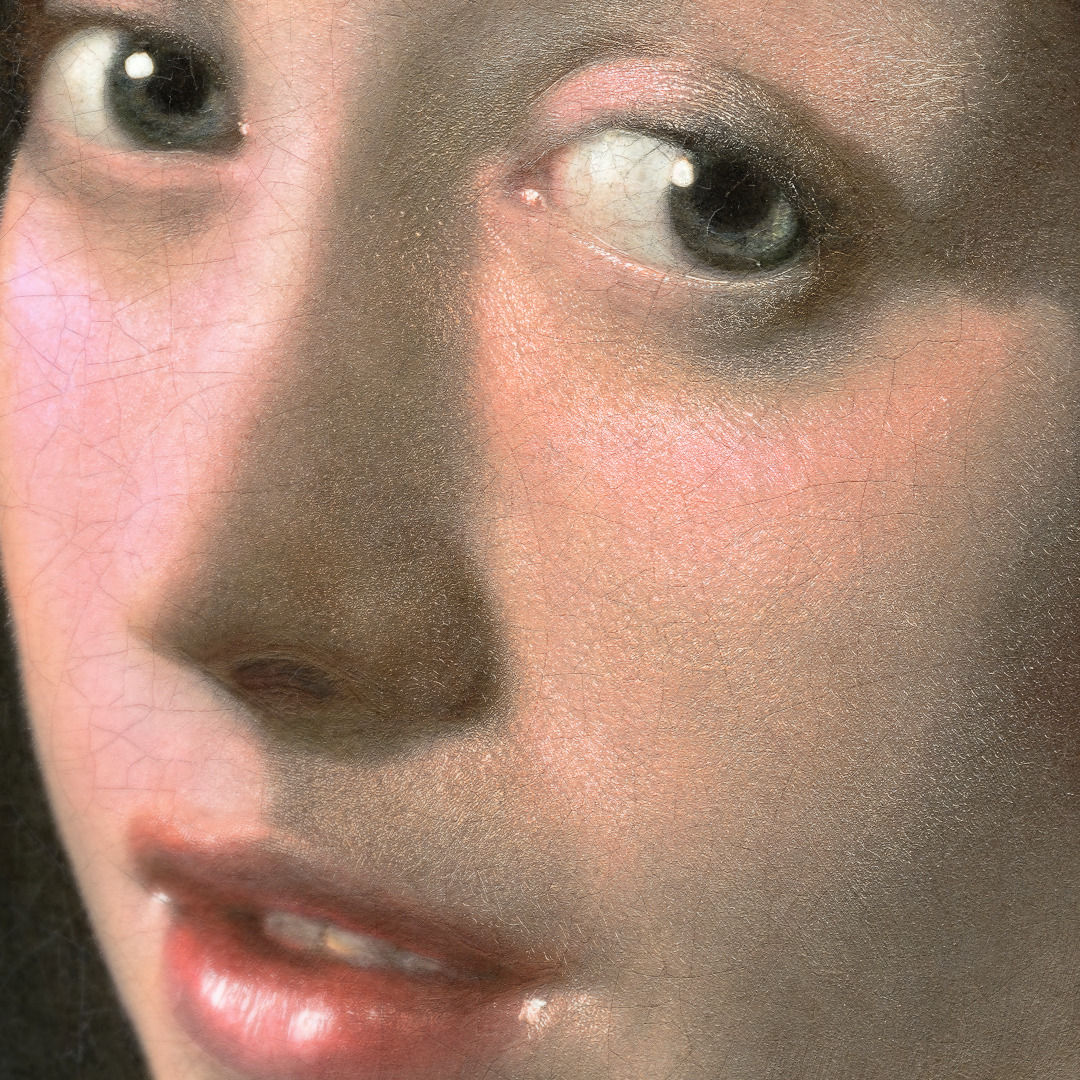}} &
    \makecell{
    \includegraphics[width=0.121\linewidth]{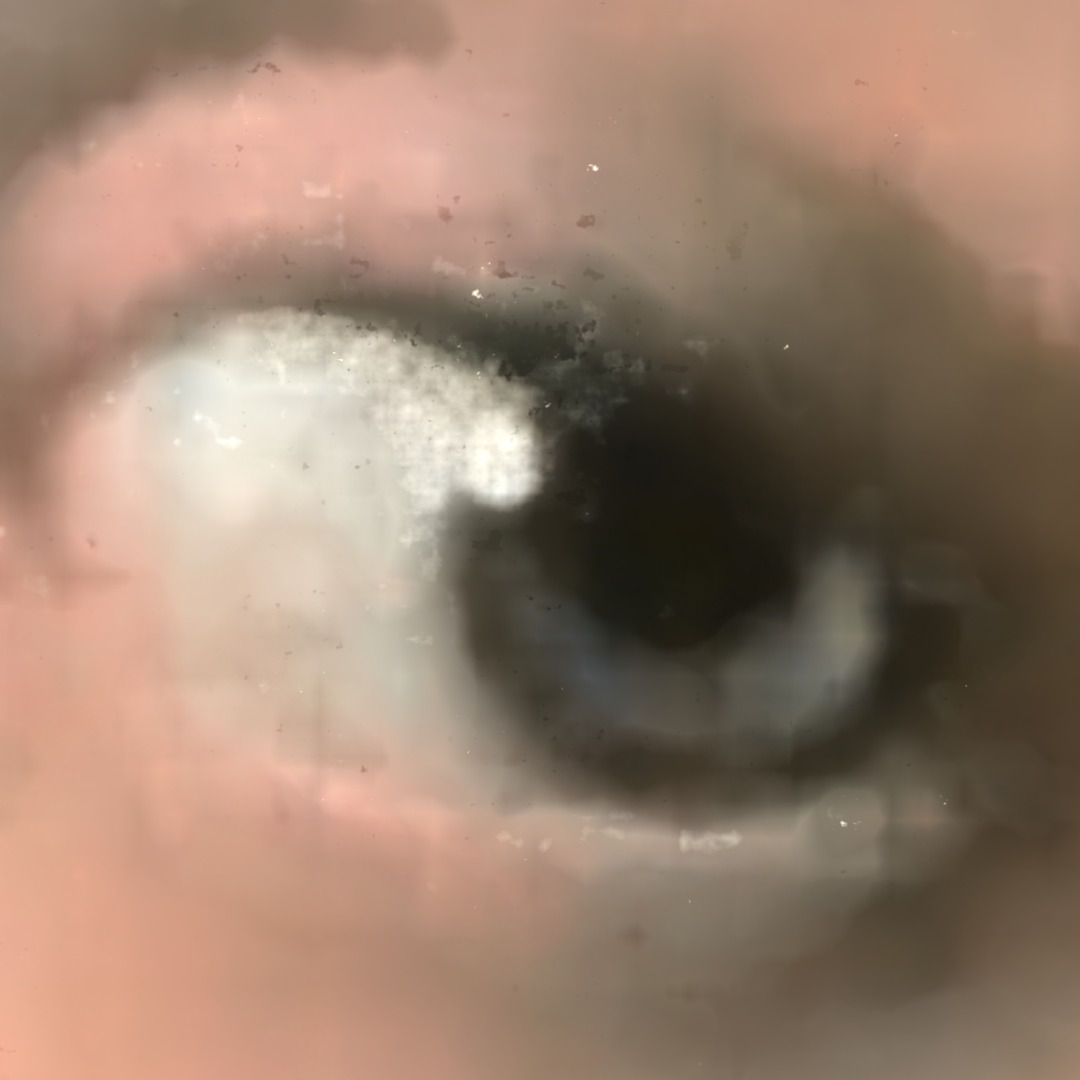} \\[-0.235mm]
    \includegraphics[width=0.121\linewidth]{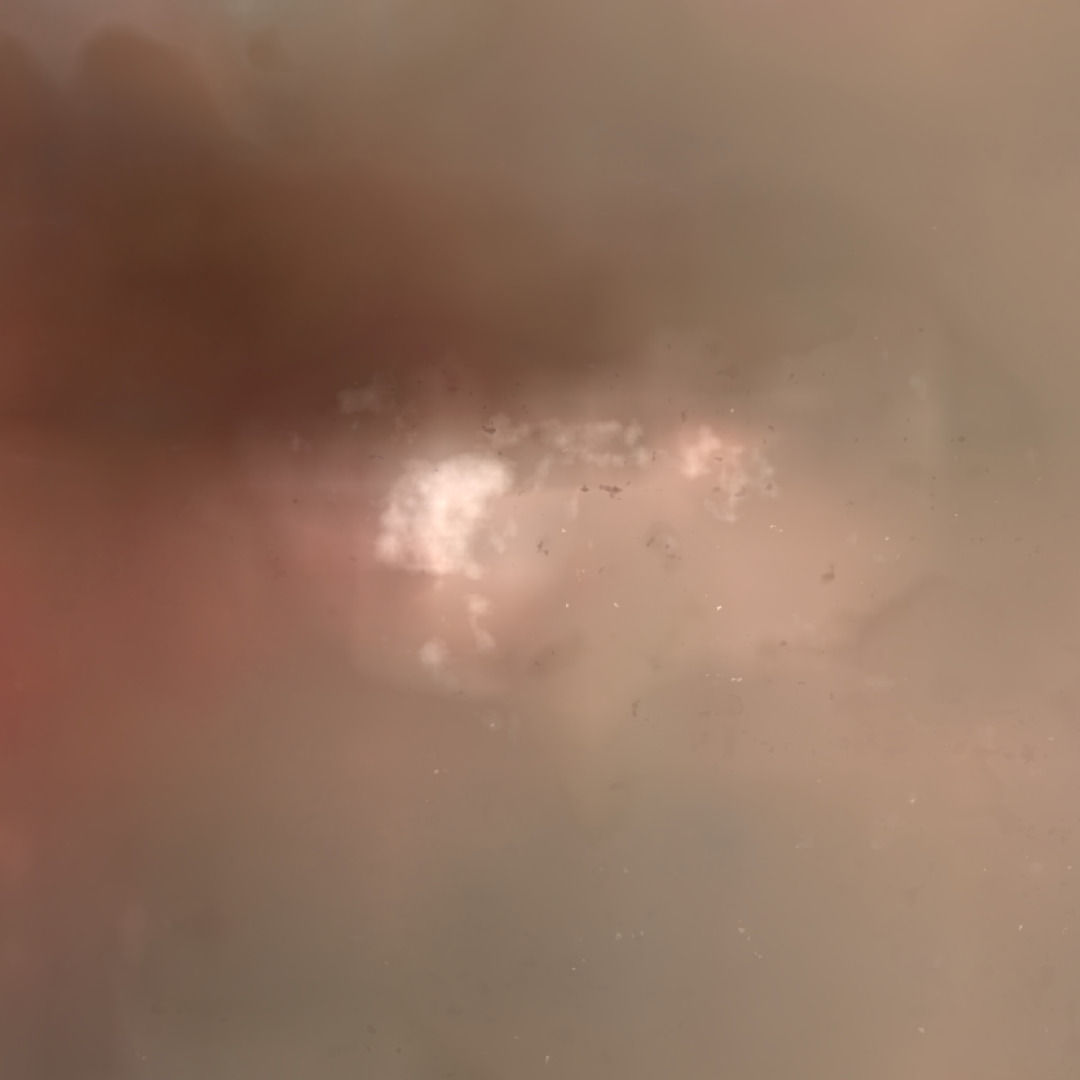} 
    } &
    \makecell{
    \includegraphics[width=0.121\linewidth]{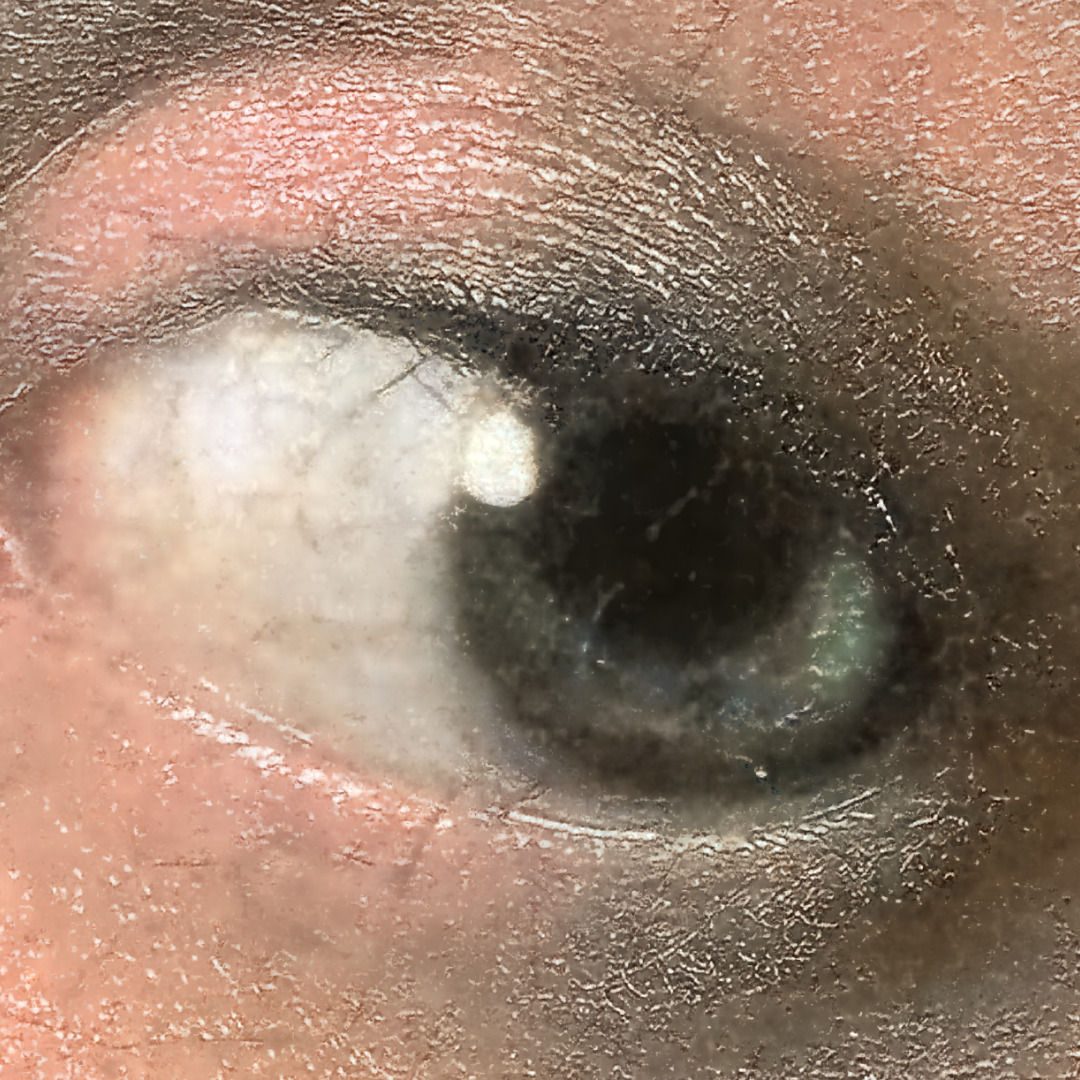} \\[-0.235mm]
    \includegraphics[width=0.121\linewidth]{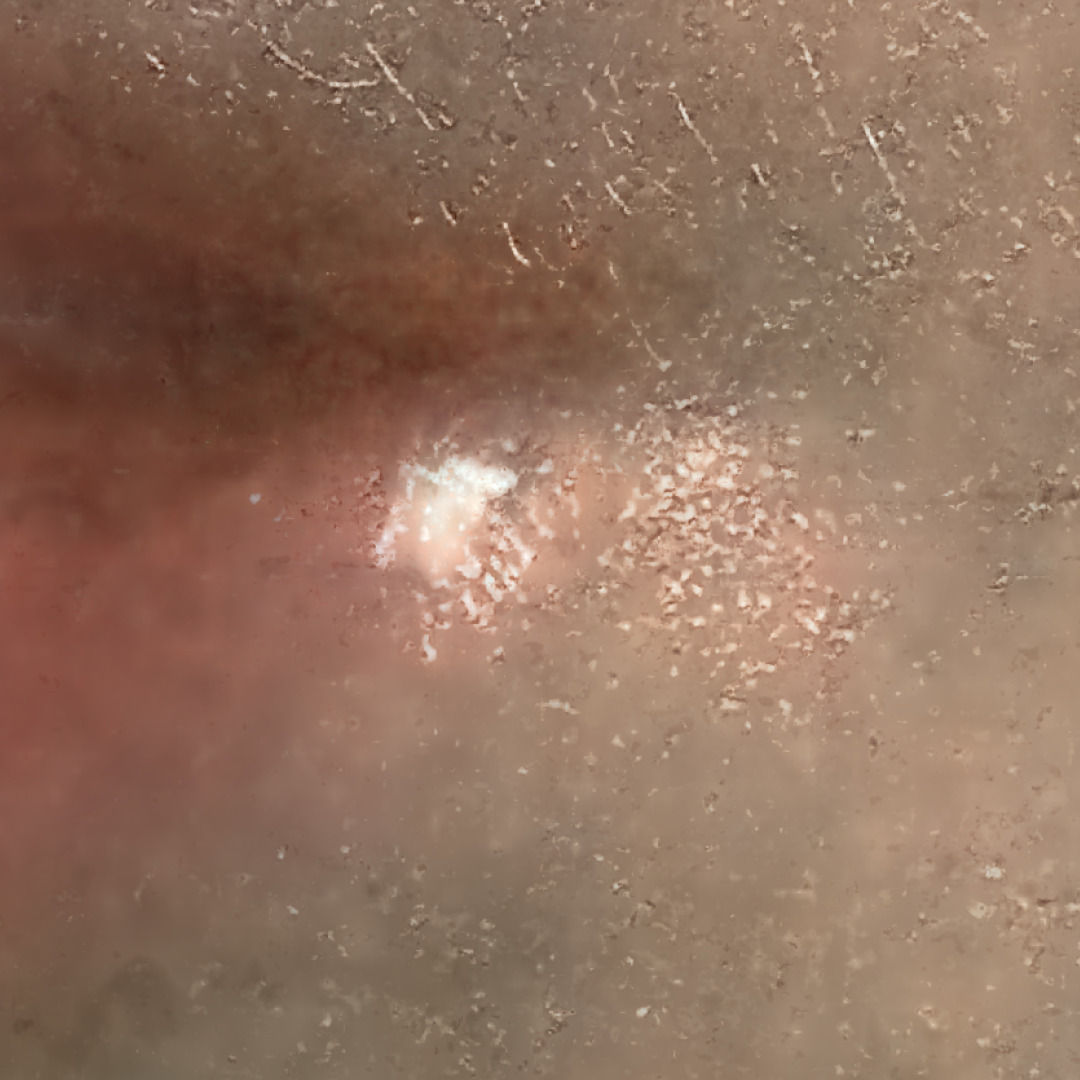} 
    } &
    \makecell{
    \includegraphics[width=0.121\linewidth]{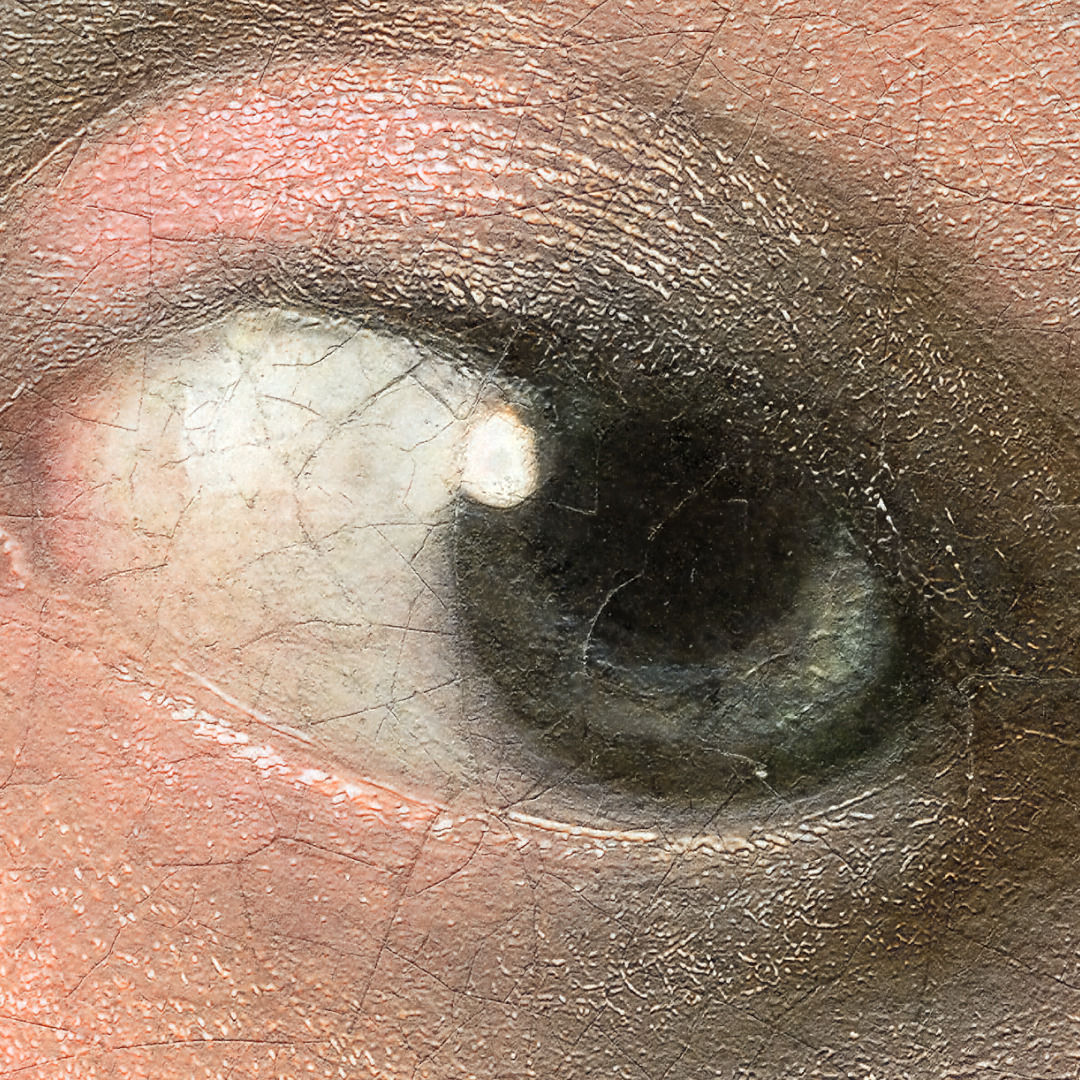} \\[-0.235mm]
    \includegraphics[width=0.121\linewidth]{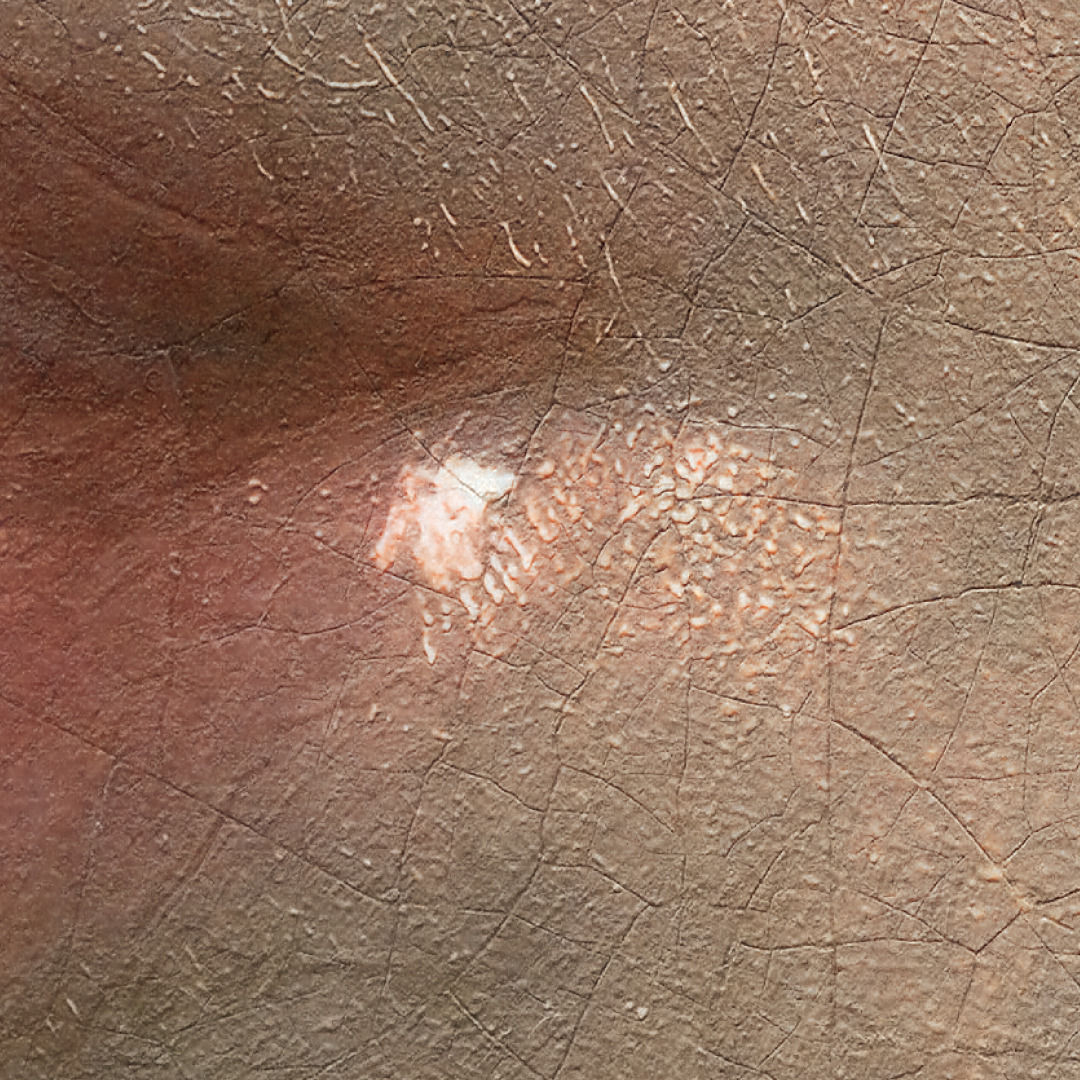} 
    } &
    \makecell{
    \includegraphics[width=0.121\linewidth]{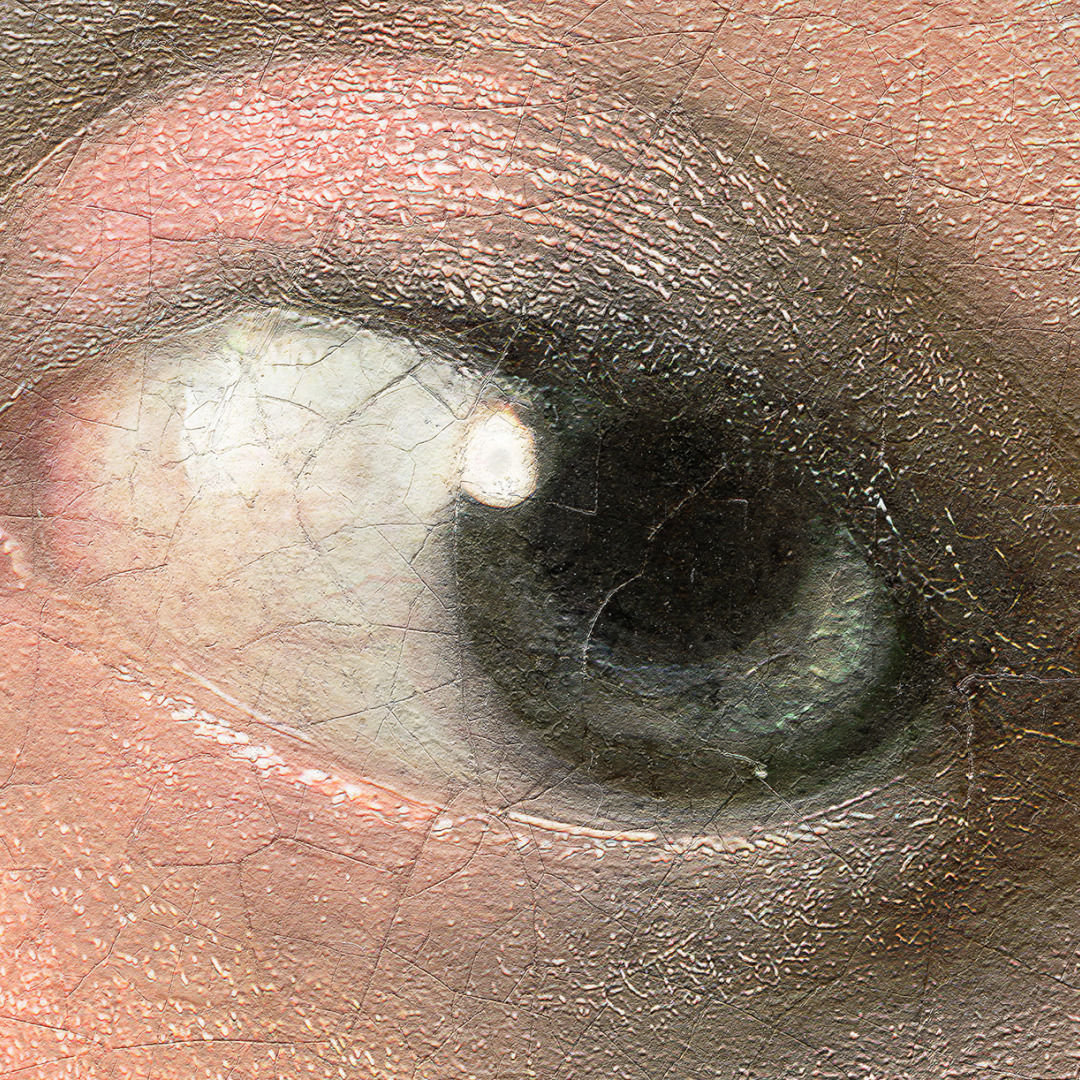} \\[-0.235mm]
    \includegraphics[width=0.121\linewidth]{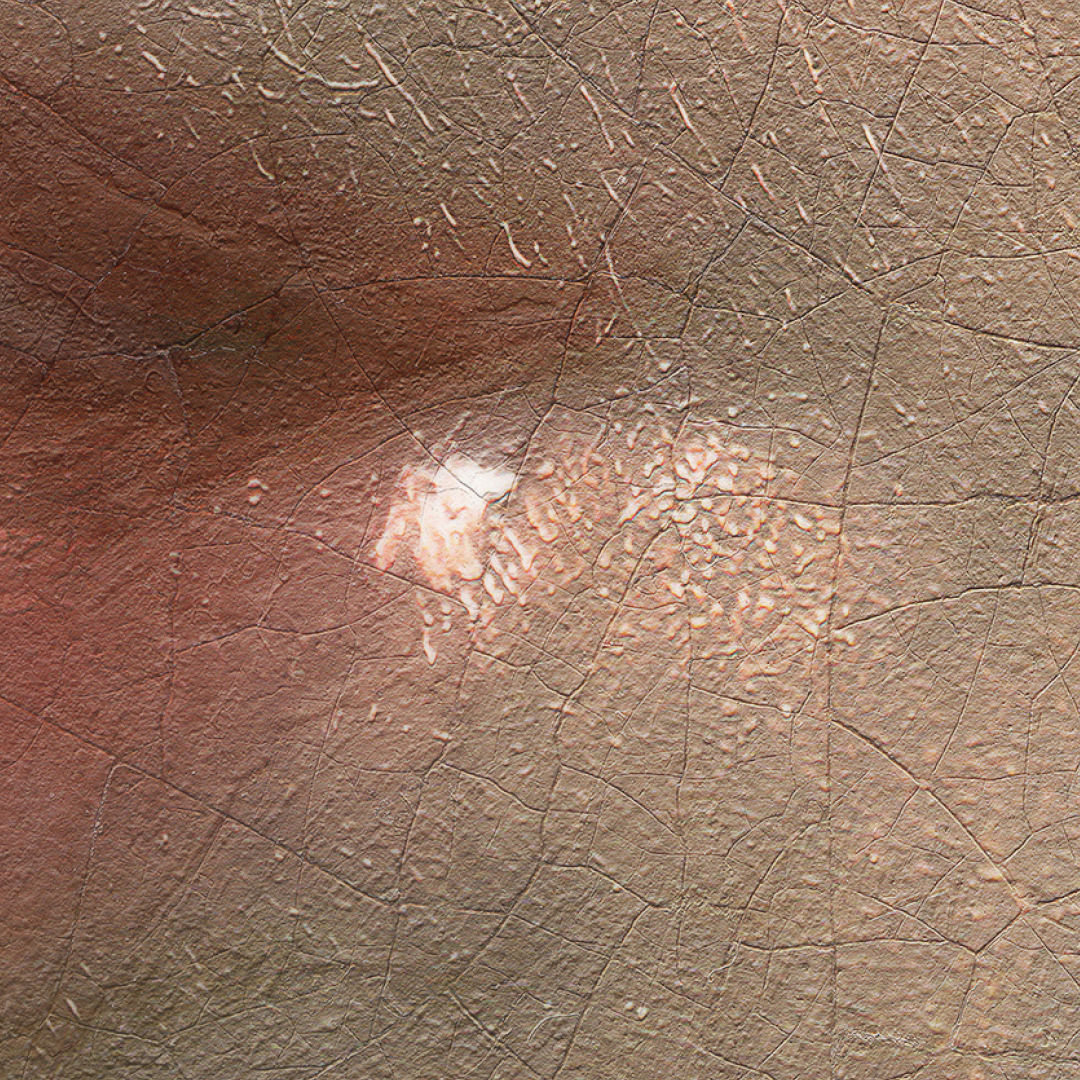} 
    } \\        
  \end{tabular}
  \vspace{-3mm}
  \caption{\label{fig:image_results_pearl}%
    Approximating an RGB image of resolution \num{20000} $\times$ \num{23466} (\SI{469}{\mega\nothing} RGB pixels) with our multiresolution hash encoding. 
    With hash table sizes $\entriesPerLevel$ of $2^{12}$, $2^{17}$, and $2^{22}$ the models shown have \SI{117}{\kilo\nothing}, \SI{2.7}{\mega\nothing}, and \SI{47.5}{\mega\nothing} trainable parameters respectively.
    With only 3.4\% of the degrees of freedom of the input, the last model achieves a reconstruction PSNR of \SI{29.8}{\decibel}.
    ``Girl With a Pearl Earring'' renovation \copyright Koorosh Orooj \href{http://profoundism.com/free_licenses.html}{(CC BY-SA 4.0)}
  }
  \vspace{-2mm}
\end{figure*}

\paragraph{Performance considerations.}
In order to optimize inference and backpropagation performance, we store hash table entries at half precision (2 bytes per entry).
We additionally maintain a master copy of the parameters in full precision for stable mixed-precision parameter updates, following \citet{micikevicius2018mixed}.

To optimally use the GPU's caches, we evaluate the hash tables level by level: when processing a batch of input positions, we schedule the computation to look up the first level of the multiresolution hash encoding for \emph{all} inputs, followed by the second level for all inputs, and so on.
Thus, only a small number of consecutive hash tables have to reside in caches at any given time, depending on how much parallelism is available on the GPU.\@
Importantly, this structure of computation \emph{automatically} makes good use of the available caches and parallelism for a wide range of hash table sizes $\entriesPerLevel$.

On our hardware, the performance of the encoding remains roughly constant as long as the hash table size stays below ${\entriesPerLevel \leq 2^{19}}$.
Beyond this threshold, performance starts to drop significantly; see \autoref{fig:image_t_sweep}.
This is explained by the \SI{6}{\mega\byte} L2 cache of our NVIDIA RTX 3090 GPU, which becomes too small for individual levels when $2 \cdot \entriesPerLevel \cdot \featuresPerEntry > 6 \cdot 2^{20}$, \ADD{with 2 being the size of a half-precision entry.}

The optimal number of feature dimensions $\featuresPerEntry$ per lookup depends on the GPU architecture.
On one hand, a small number favors cache locality in the previously mentioned streaming approach, but on the other hand, a large $\featuresPerEntry$ favors memory coherence by allowing for $\featuresPerEntry$-wide vector load instructions.
${\featuresPerEntry=2}$ gave us the best cost-quality trade-off (see \autoref{fig:image_ftl_sweep}) and we use it in all experiments.

\paragraph{Architecture.}
In all tasks, except for NeRF which we will describe later, we use an MLP with two hidden layers that have a width of \num{64} neurons, rectified linear unit (ReLU) activation functions on their hidden layers, \ADD{and a linear output layer.}
\ADD{The maximum resolution $\maxResolution$ is set to ${2048\,\times}$ scene size for NeRF and signed distance functions, to half of the gigapixel image width, and $2^{19}$ in radiance caching (large value to support close-by objects in expansive scenes).}

\paragraph{Initialization.}
We initialize neural network weights according to Glorot and Bengio~\shortcite{Glorot:2010} to provide a reasonable scaling of activations and their gradients throughout the layers of the neural network.
We initialize the hash table entries using the uniform distribution $\Uniform(-10^{-4}, 10^{-4})$ to provide a small amount of randomness while encouraging initial predictions close to zero.
We also tried a variety of different distributions, including zero-initialization, which all resulted in a very slightly worse initial convergence speed.
The hash table appears to be robust to the initialization scheme.

\paragraph{Training.}
We jointly train the neural network weights and the hash table entries by applying Adam~\cite{KingmaB14}, where we set ${\beta_1 = 0.9, \beta_2 = 0.99, \epsilon=10^{-15}}$, 
The choice of $\beta_1$ and $\beta_2$ makes only a small difference, but the small value of $\epsilon=10^{-15}$ can significantly accelerate the convergence of the hash table entries when their gradients are sparse and weak.
To prevent divergence after long training periods, we apply a weak L2 regularization (factor $10^{-6}$) to the neural network weights, but not to the hash table entries.

\ADD{When fitting gigapixel images or NeRFs, we use the $\Loss^2$ loss.
For signed distance functions, we use the mean absolute percentage error (MAPE), defined as $\frac{|\mathrm{prediction}\,-\,\mathrm{target}|}{|\mathrm{target}|\,+\,0.01}$, and for neural radiance caching we use a luminance-relative $\Loss^2$ loss~\citep{mueller2021realtime}.}

\ADD{We observed fastest convergence with a learning rate of $10^{-4}$ for signed distance functions and $10^{-2}$ otherwise, as well a a batch size of $2^{14}$ for neural radiance caching and $2^{18}$ otherwise.}

Lastly, we skip Adam steps for hash table entries whose gradient is exactly $0$.
This saves ${\sim\!10\%}$ performance when gradients are sparse, which is a common occurrence with ${\entriesPerLevel \gg \mathrm{BatchSize}}$.
Even though this heuristic violates some of the assumptions behind Adam, we observe no degradation in convergence.

\paragraph{Non-spatial input dimensions ${\auxDims \in \R^\numAuxDims}$.}
The multiresolution hash encoding targets spatial coordinates with relatively low dimensionality.
All our experiments operate either in $2$D or $3$D.
However, it is frequently useful to input auxiliary dimensions ${\auxDims \in \R^\numAuxDims}$ to the neural network, such as the view direction and material parameters when learning a light field.
In such cases, the auxiliary dimensions can be encoded with established techniques whose cost does not scale superlinearly with dimensionality; we use the one-blob encoding~\cite{mueller2019nis} in neural radiance caching~\citep{mueller2021realtime} and the spherical harmonics basis in NeRF, similar to concurrent work~\cite{verbin2021refnerf,yu2021plenoxels}.

\section{Experiments}%
\label{Sec:Experiments}

To highlight the versatility and high quality of the encoding, we compare it with previous encodings in four distinct computer graphics primitives that benefit from encoding spatial coordinates.

\begin{figure*}
  \small\sffamily
  \vspace{-3mm}
  
\setlength{\tabcolsep}{1pt}%
\renewcommand{\arraystretch}{1.1}%
\hspace*{1.5mm}\begin{tabular}{c@{\hskip 2.0mm}ccc@{\hskip 2.0mm}ccccc}
	\makecell{Hash (ours)} &
	\makecell{NGLOD} &
	\makecell{Hash (ours)} &
	\makecell{Frequency} &
	&
	\makecell{Frequency} &
	\makecell{Hash (ours)} &
	\makecell{NGLOD} &
	\makecell{Hash (ours)}
	\\[0.5mm]

	\makecell{\includegraphics[width=0.19\linewidth]{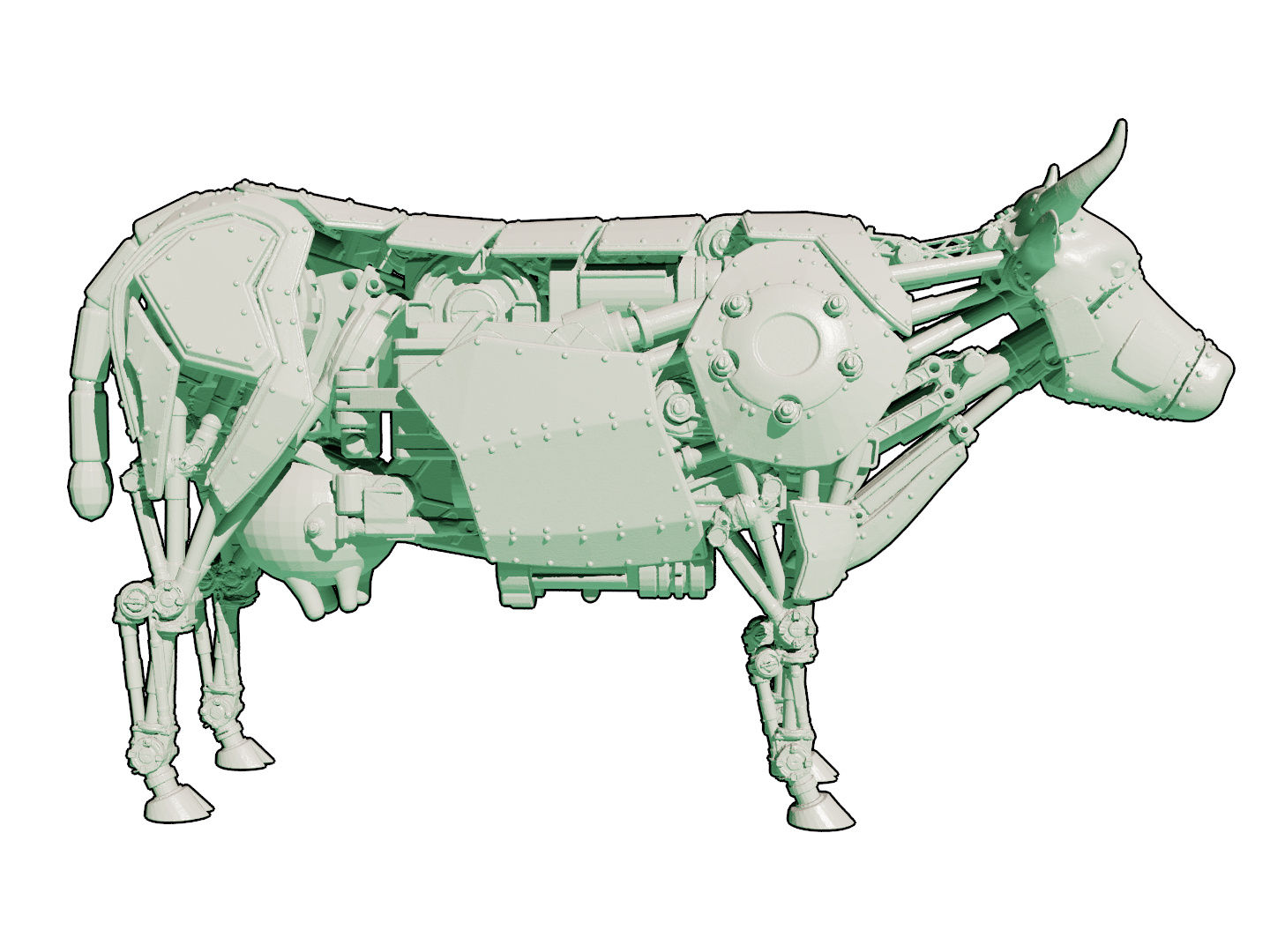}} &
	\makecell{\frame{\includegraphics[width=0.095\linewidth]{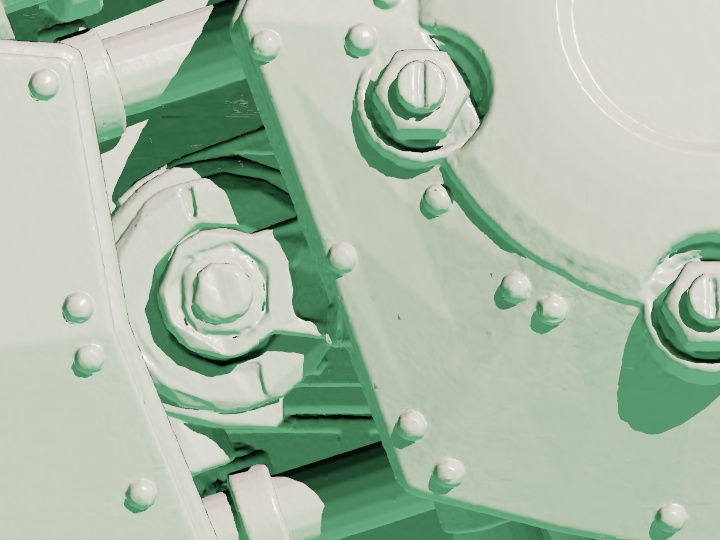}}\\[-0.5mm]\frame{\includegraphics[width=0.095\linewidth]{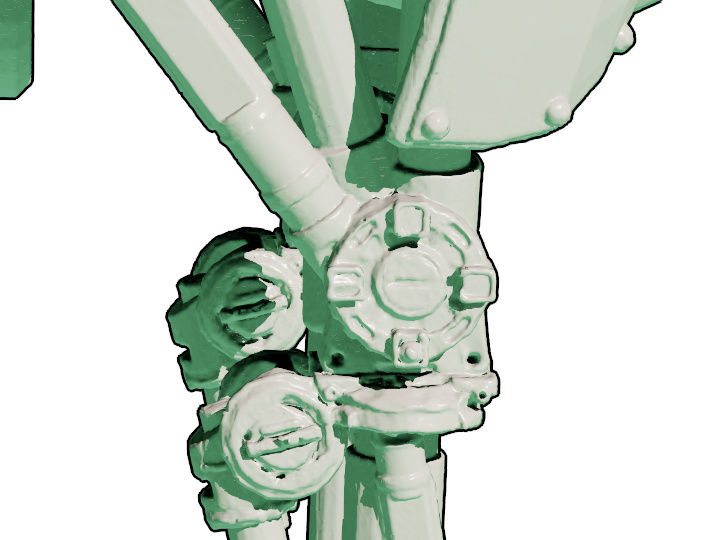}}} &
	\makecell{\frame{\includegraphics[width=0.095\linewidth]{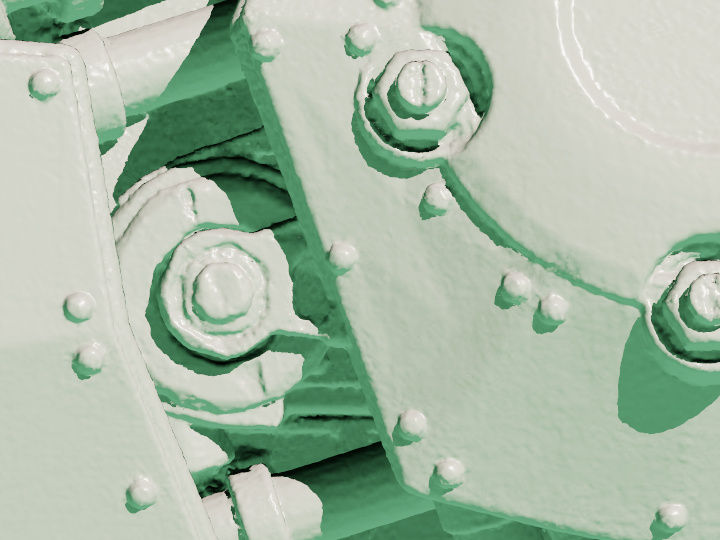}}\\[-0.5mm]\frame{\includegraphics[width=0.095\linewidth]{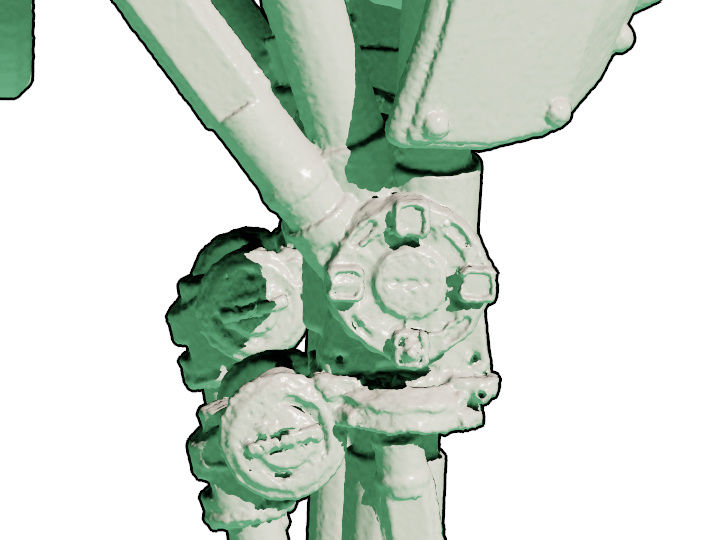}}} &
	\makecell{\frame{\includegraphics[width=0.095\linewidth]{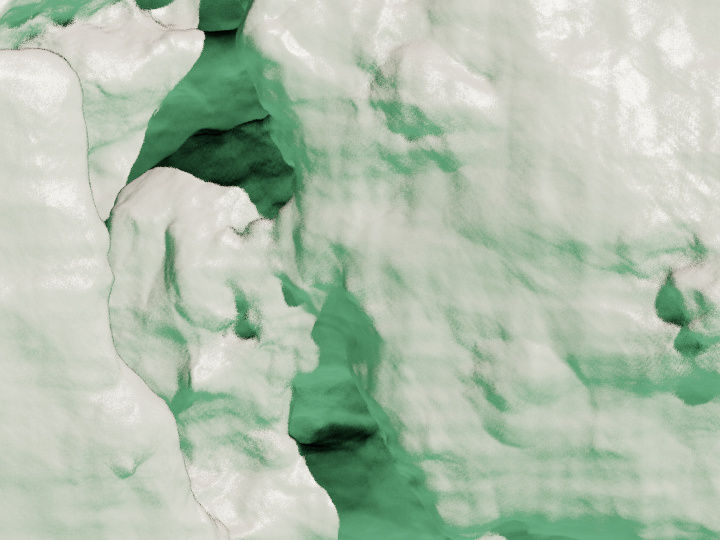}}\\[-0.5mm]\frame{\includegraphics[width=0.095\linewidth]{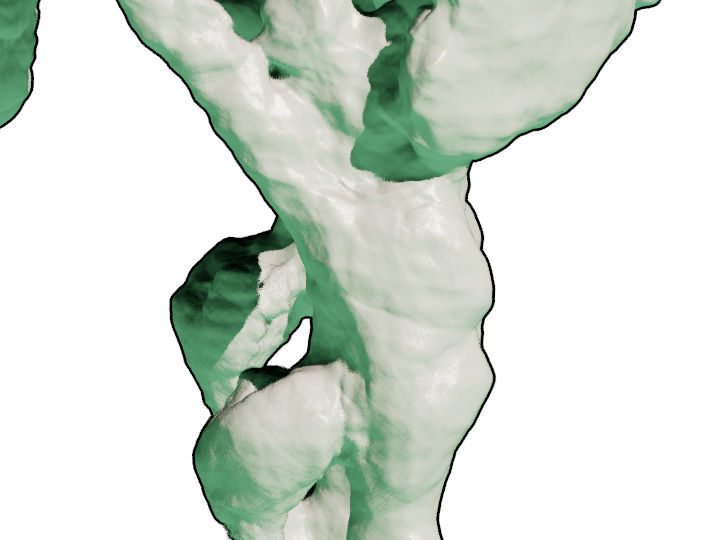}}} &
	&
	\makecell{\frame{\includegraphics[width=0.095\linewidth]{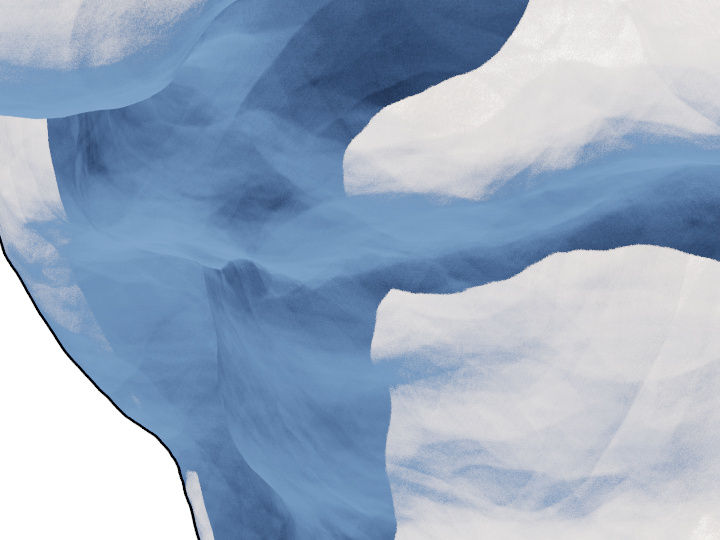}}\\[-0.5mm]\frame{\includegraphics[width=0.095\linewidth]{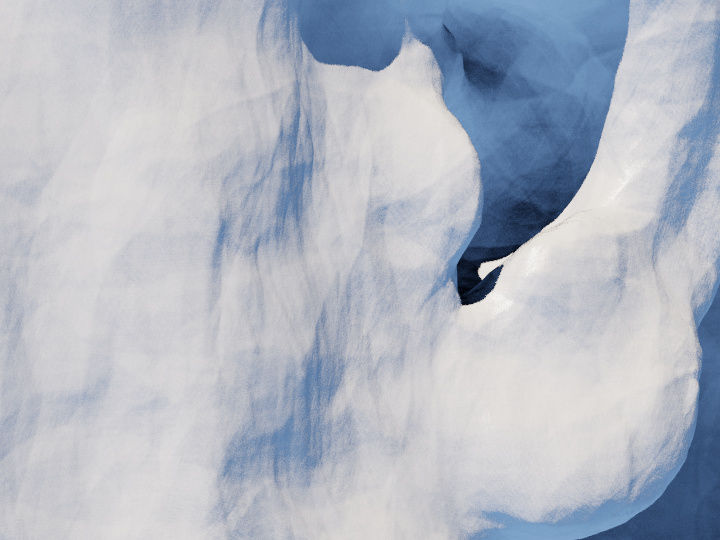}}} &
	\makecell{\frame{\includegraphics[width=0.095\linewidth]{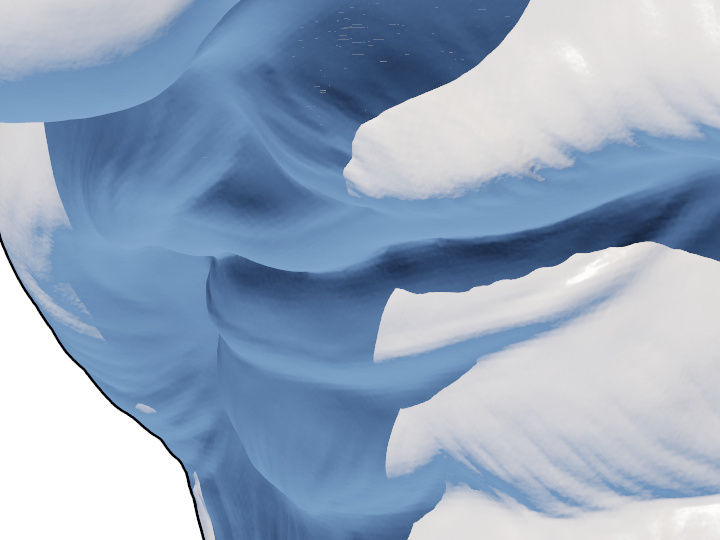}}\\[-0.5mm]\frame{\includegraphics[width=0.095\linewidth]{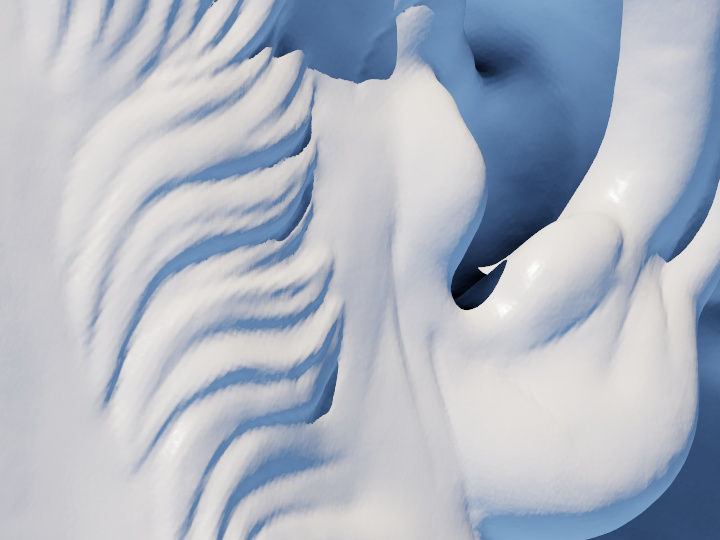}}} &
	\makecell{\frame{\includegraphics[width=0.095\linewidth]{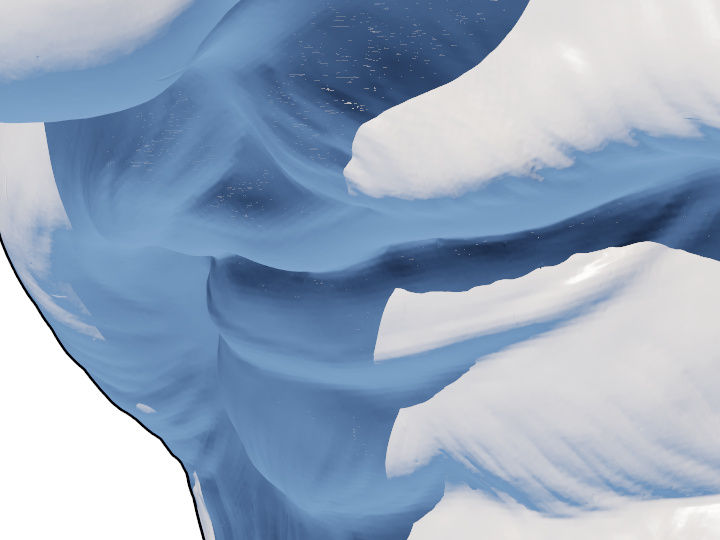}}\\[-0.5mm]\frame{\includegraphics[width=0.095\linewidth]{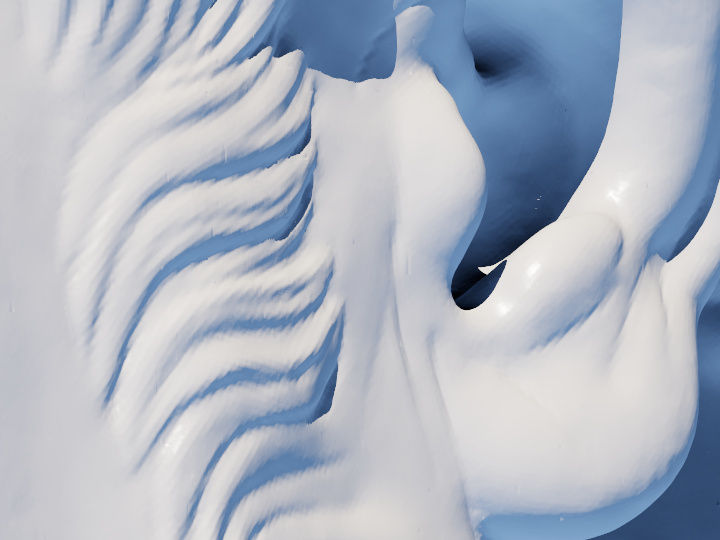}}} &
	\makecell{\hspace*{-1mm}\includegraphics[width=0.19\linewidth]{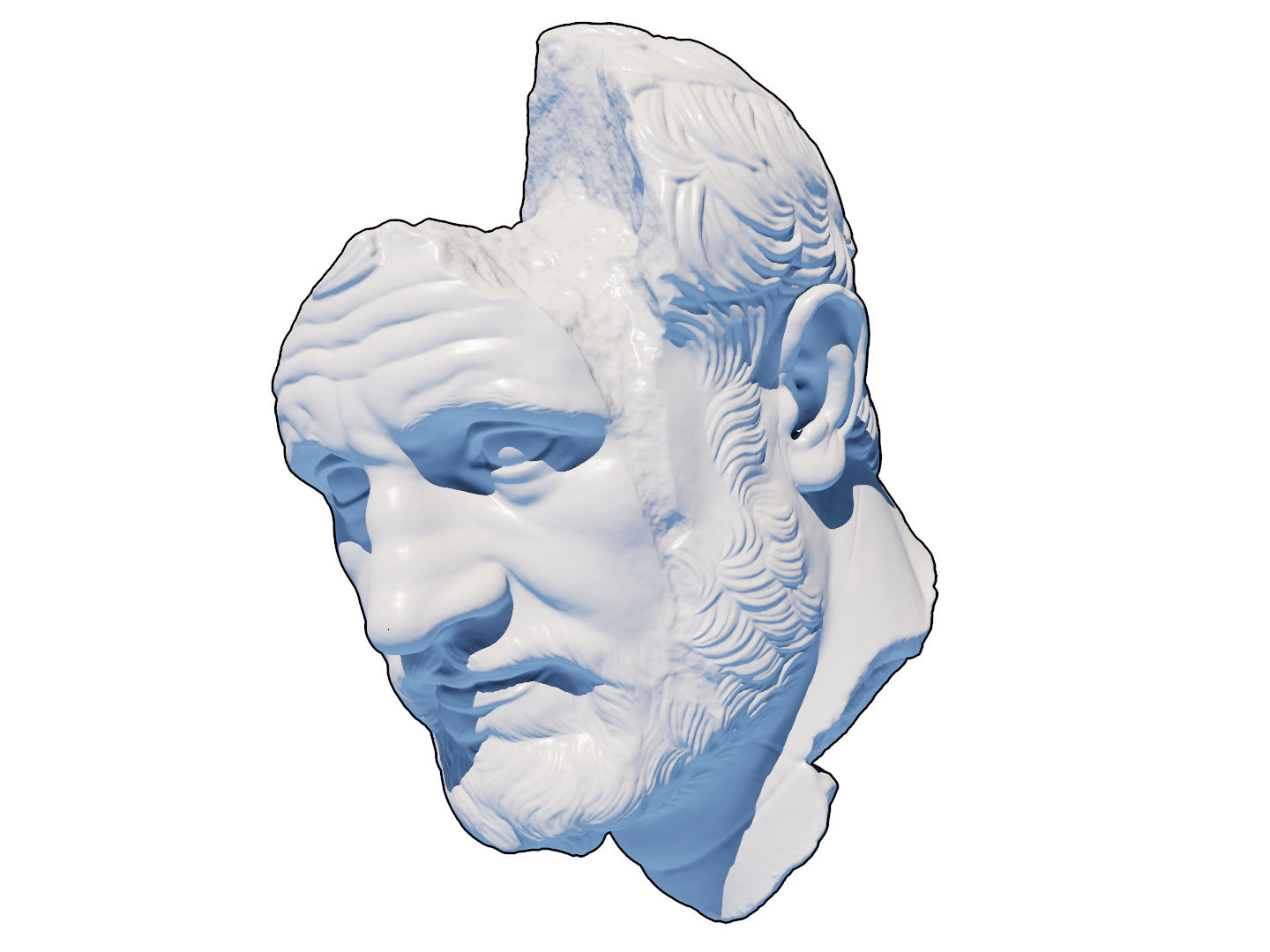}} \\[-1.2mm] &
	\footnotesize \SI{22.3}{\mega\nothing} (params) &
	\footnotesize \SI{12.2}{\mega\nothing}&
	\footnotesize \SI{124.9}{\kilo\nothing} & &
	\footnotesize \SI{124.9}{\kilo\nothing} &
	\footnotesize \SI{12.2}{\mega\nothing} &
	\footnotesize \SI{16.0}{\mega\nothing} &
	\footnotesize \\[-1.2mm] &
	\footnotesize 1:56 (mm:ss) &
	\footnotesize 1:14 &
	\footnotesize 1:32 & &
	\footnotesize 2:10 &
	\footnotesize 1:54 &
	\footnotesize 1:49 &
	\footnotesize \\[-1.2mm] &
	\footnotesize 0.9777 (IoU) &
	\footnotesize 0.9812 &
	\footnotesize 0.8432 & &
	\footnotesize 0.9898 &
	\footnotesize 0.9997 &
	\footnotesize 0.9998 &
	\footnotesize \\[1.0mm]

	\makecell{\vspace*{-4mm}\includegraphics[width=0.19\linewidth]{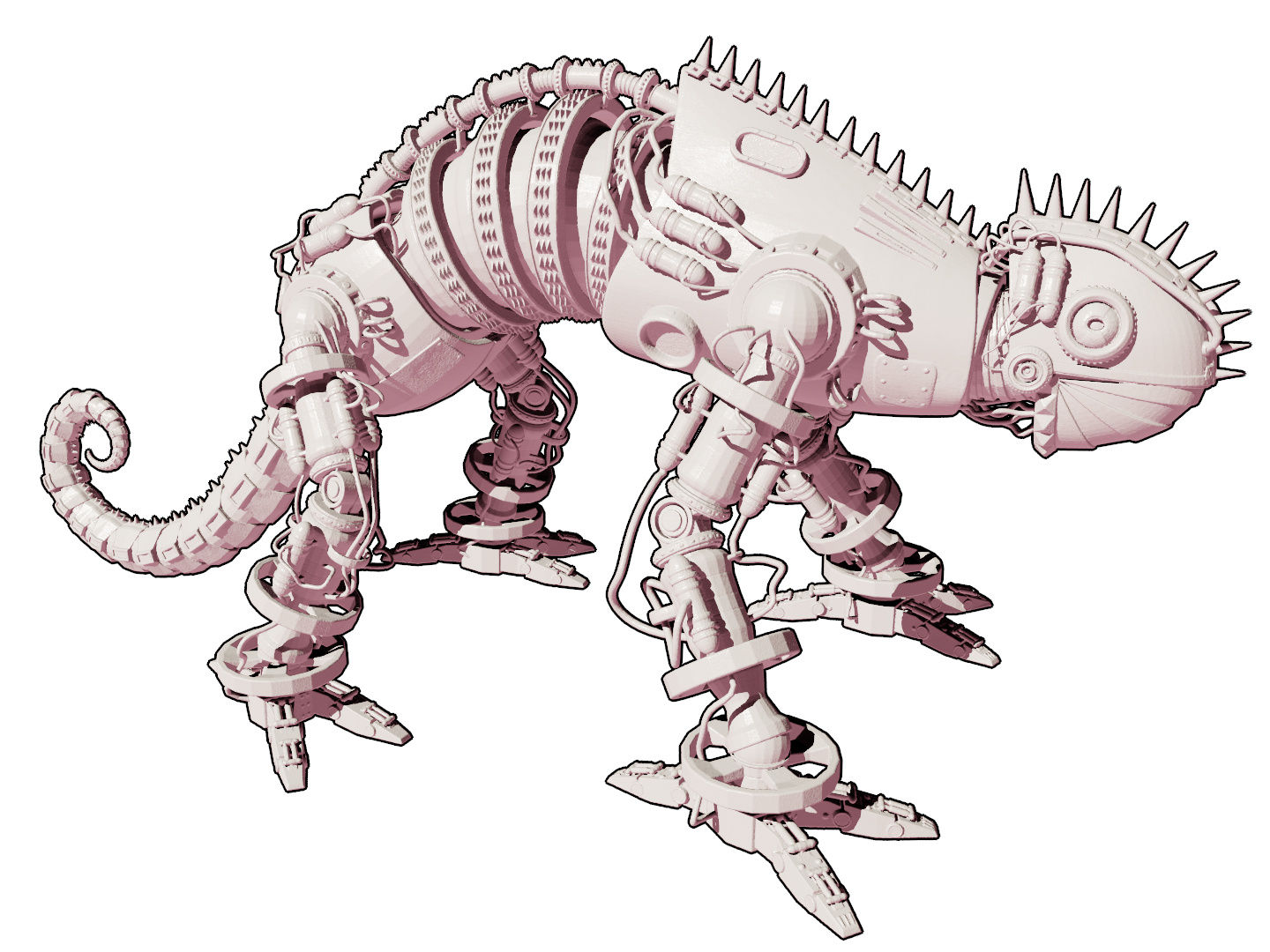}} &
	\makecell{\frame{\includegraphics[width=0.095\linewidth]{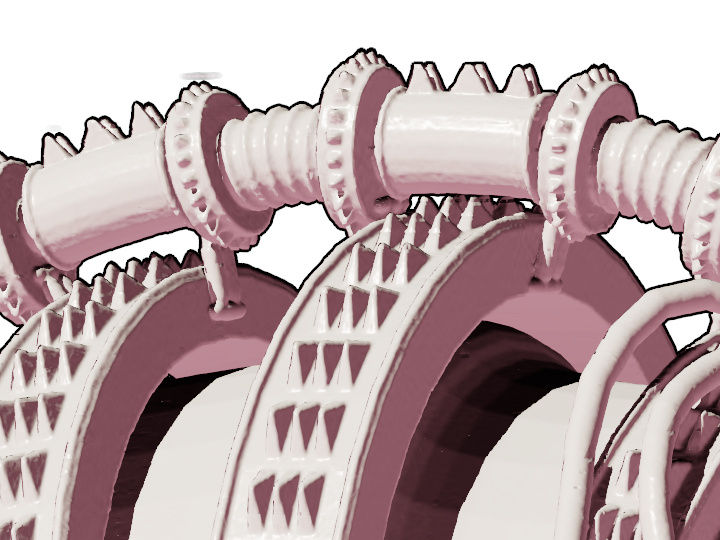}}\\[-0.5mm]\frame{\includegraphics[width=0.095\linewidth]{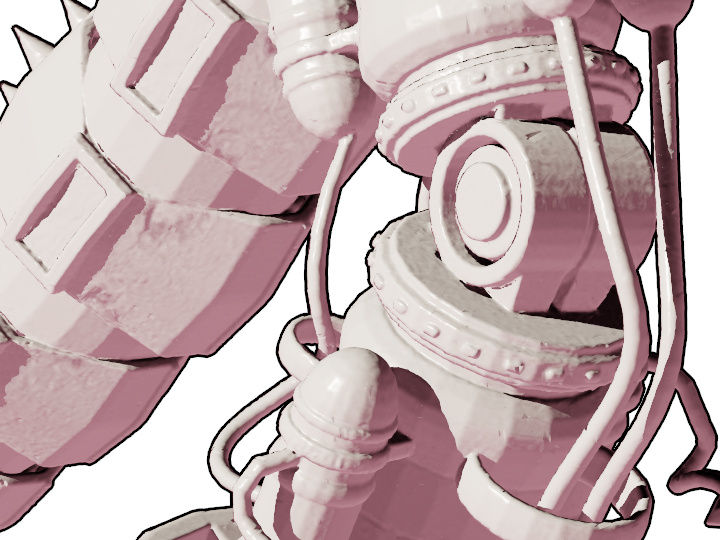}}} &
	\makecell{\frame{\includegraphics[width=0.095\linewidth]{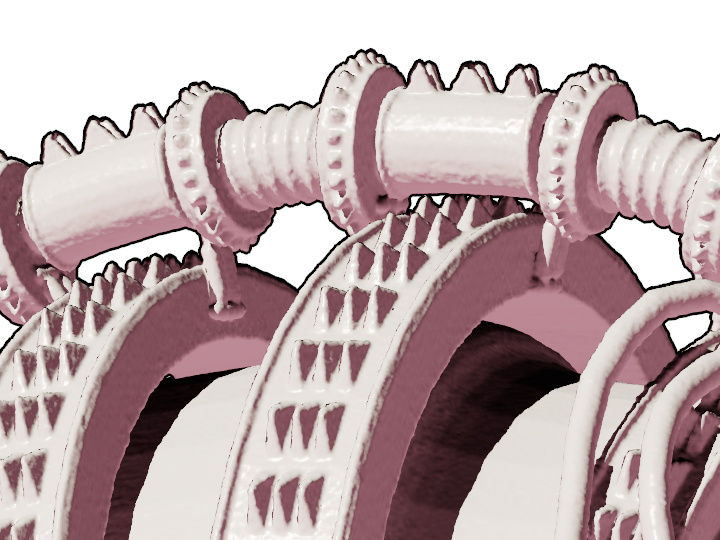}}\\[-0.5mm]\frame{\includegraphics[width=0.095\linewidth]{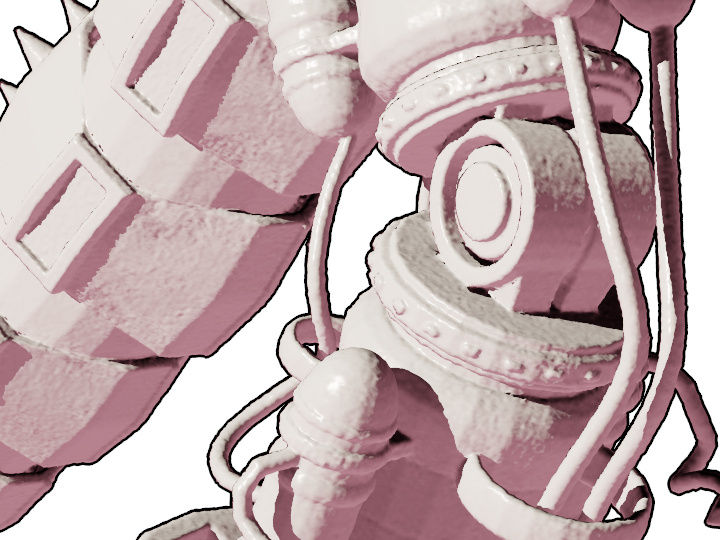}}} &
	\makecell{\frame{\includegraphics[width=0.095\linewidth]{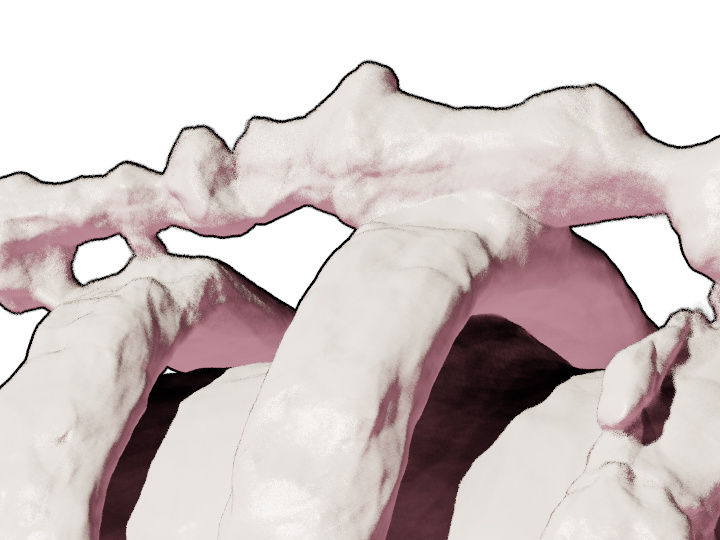}}\\[-0.5mm]\frame{\includegraphics[width=0.095\linewidth]{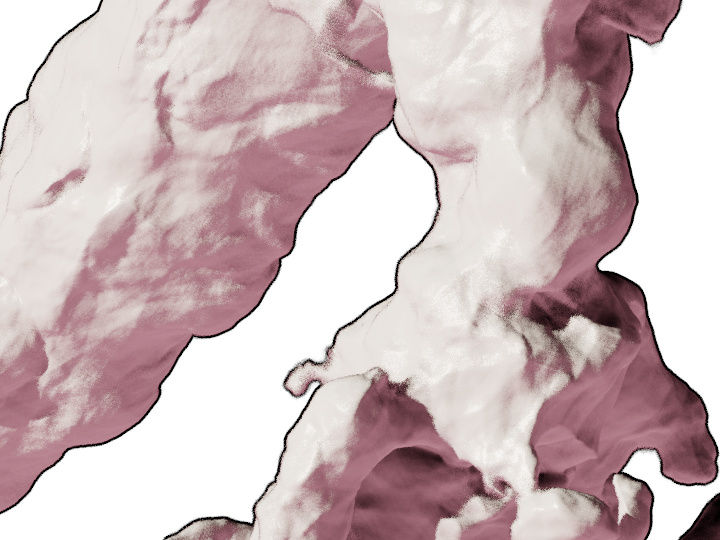}}} &
	&
	\makecell{\frame{\includegraphics[width=0.095\linewidth]{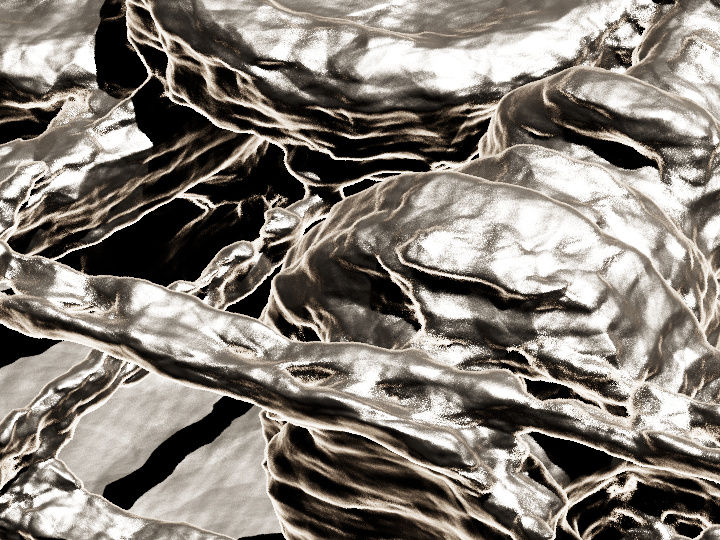}}\\[-0.5mm]\frame{\includegraphics[width=0.095\linewidth]{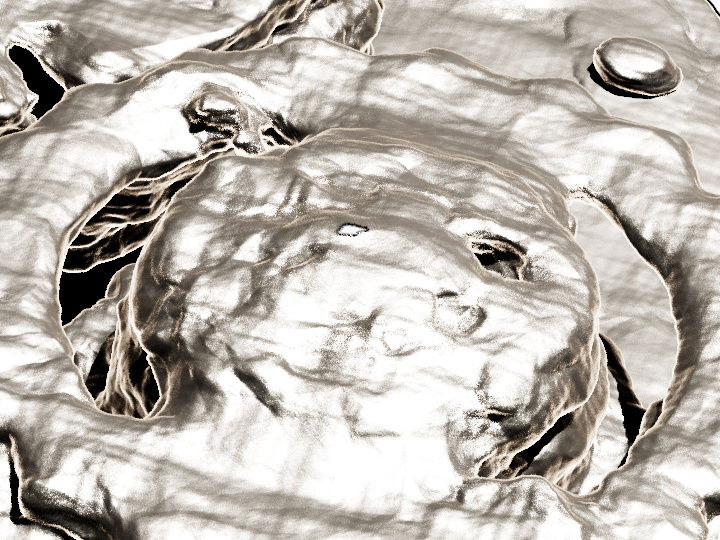}}} &
	\makecell{\frame{\includegraphics[width=0.095\linewidth]{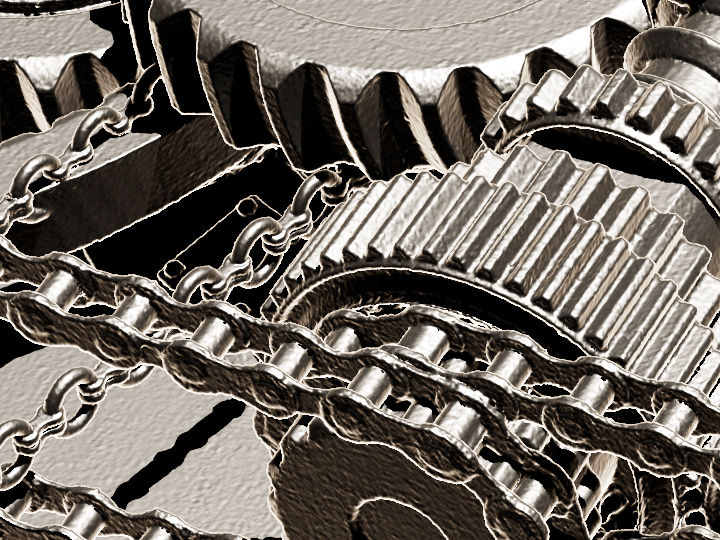}}\\[-0.5mm]\frame{\includegraphics[width=0.095\linewidth]{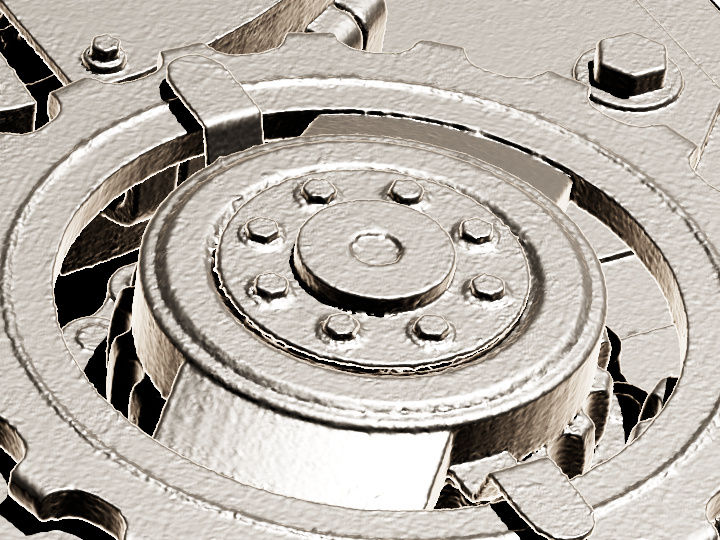}}} &
	\makecell{\frame{\includegraphics[width=0.095\linewidth]{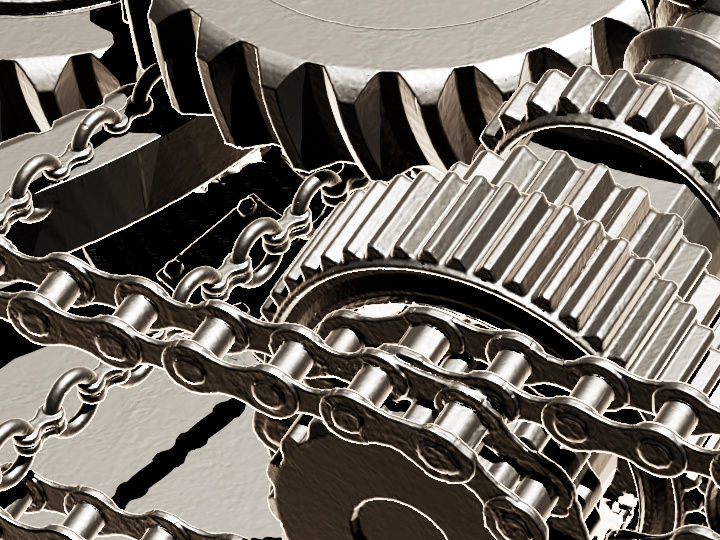}}\\[-0.5mm]\frame{\includegraphics[width=0.095\linewidth]{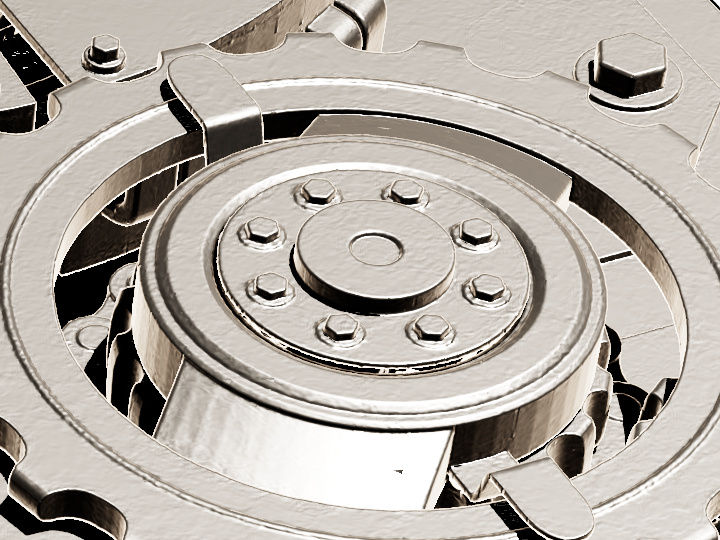}}} &
	\makecell{\includegraphics[width=0.19\linewidth]{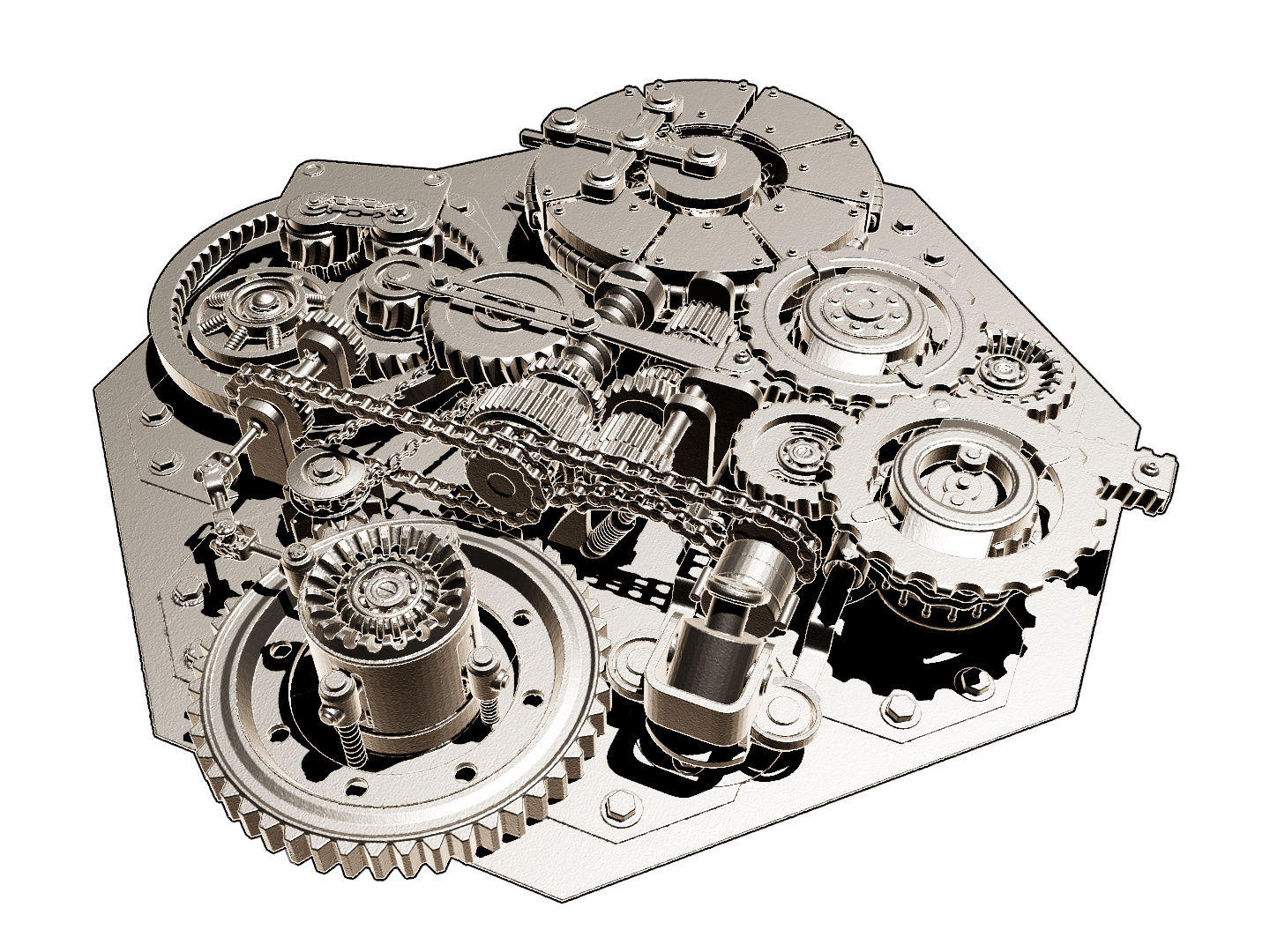}} \\[-1.2mm] &
	\footnotesize \SI{11.1}{\mega\nothing} (params) &
	\footnotesize \SI{12.2}{\mega\nothing} &
	\footnotesize \SI{124.9}{\kilo\nothing} & &
	\footnotesize \SI{124.9}{\kilo\nothing} &
	\footnotesize \SI{12.2}{\mega\nothing} &
	\footnotesize \SI{24.2}{\mega\nothing} &
	\footnotesize \\[-1.2mm] &
	\footnotesize 1:37 (mm:ss) &
	\footnotesize 1:19 &
	\footnotesize 1:35 & &
	\footnotesize 1:21 &
	\footnotesize 1:04 &
	\footnotesize 1:50 &
	\footnotesize \\[-1.2mm] &
	\footnotesize 0.9911 (IoU) &
	\footnotesize 0.9872 &
	\footnotesize 0.8470 & &
	\footnotesize 0.7575 &
	\footnotesize 0.9691 &
	\footnotesize 0.9749 &
	\footnotesize \\
\end{tabular}

  \vspace{-3mm}
  \caption{\label{fig:sdf_results}%
    Neural signed distance functions trained for \num{11000} steps.
    The frequency encoding~\cite{mildenhall2020nerf} struggles to capture the sharp details on these intricate models.
    NGLOD~\cite{takikawa2021nglod} achieves the highest visual quality, at the cost of only training the SDF inside the cells of a close-fitting octree.
    Our hash encoding exhibits similar numeric quality in terms of intersection over union (IoU) and can be evaluated anywhere in the scene.
    However, it also exhibits visually undesirable surface roughness that we attribute to randomly distributed hash collisions. Bearded Man \copyright Oliver Laric \href{https://creativecommons.org/licenses/by-nc-sa/2.0/}{(CC BY-NC-SA 2.0)}
  }
\end{figure*}

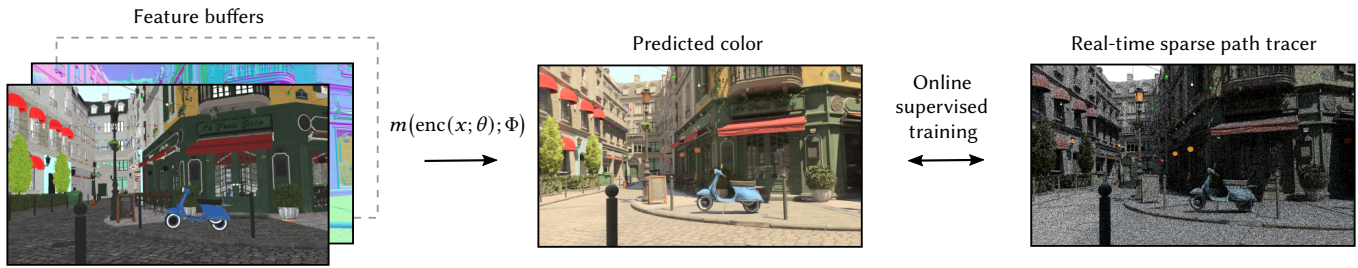
\begin{figure*}
  \small\sffamily%
  \vspace{6mm}
  \begin{overpic}[width=1.0\linewidth]{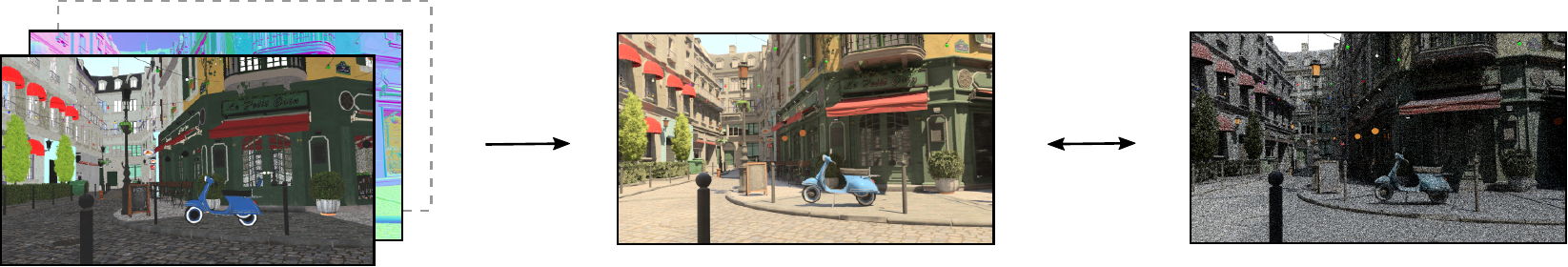}
    \put(9,18) { Feature buffers }
    \put(28.0,10) { $\nn\big(\enc(x; \Params); \Phi\big)$ }
    \put(46,16) { Predicted color }
    \put(66.8,13.1) { Online }
    \put(65.5,11.3) { supervised }
    \put(66.5,9.5) { training }
    \put(78.5,16) { Real-time sparse path tracer }
  \end{overpic}
  \vspace{-6mm}
  \caption{
    Summary of the neural radiance caching application~\cite{mueller2021realtime}.
    The MLP $\nn\big(\enc(x; \Params); \Phi\big)$ is tasked with predicting photorealistic pixel colors from feature buffers \emph{independently for each pixel}.
    The feature buffers contain, among other variables, the world-space position $\pos$, which we propose to encode with our method.
    Neural radiance caching is a challenging application, because it is supervised \emph{online during real-time rendering}.
    The training data are a sparse set of light paths that are continually spawned from the camera view.
    As such, the neural network and encoding do \emph{not} learn a general mapping from features to color, but rather they \emph{continually overfit} to the current shape and lighting. To support animated content, training has a budget of \emph{one} millisecond per frame.
  }\label{fig:nrc-algorithm}
\end{figure*}

\begin{figure*}
  \setlength{\tabcolsep}{1pt}%
  \setlength{\fboxrule}{1.5pt}%
  \setlength{\fboxsep}{0.0pt}%
  \renewcommand{\arraystretch}{0}%
  \small\sffamily%
  \vspace{1mm}
  \begin{tabularx}{\linewidth}{cccccc}%
    \multicolumn{3}{c}{Multiresolution hash encoding (Ours), ${\entriesPerLevel=15}$, 133 FPS} & \multicolumn{3}{c}{Triangle wave encoding \cite{mueller2021realtime}, 147 FPS} \\
    \cmidrule(lr){1-3} \cmidrule(lr){4-6}
    Far view & Medium view & Close-by view & Far view & Medium view & Close-by view \\
    \begin{overpic}[width=0.163\linewidth,trim={400px 0 800px 0},clip]{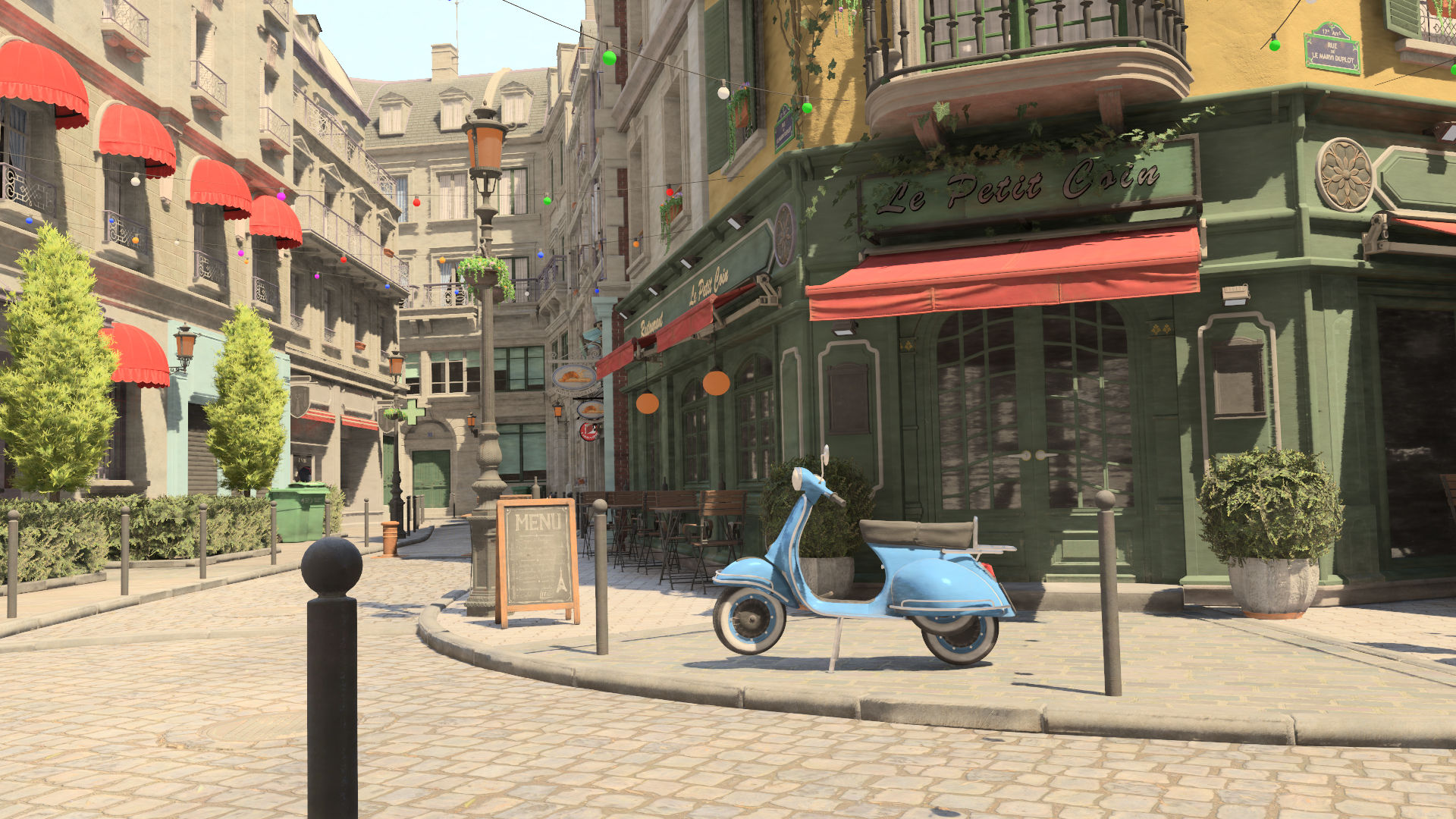}
      \put(14,48.75) { \makebox(0,0){\tikz\draw[red,thick] (0,0) rectangle (0.034\linewidth, 0.016\linewidth);} }
      \put(0.0,2.0) { \fcolorbox{red}{red}{\includegraphics[width=0.11\linewidth,trim={500px 500px 1280px 520px},clip]{nrc/zoom/render-far.jpg}} }
    \end{overpic} &
    \includegraphics[width=0.163\linewidth,trim={600px 0 600px 0},clip]{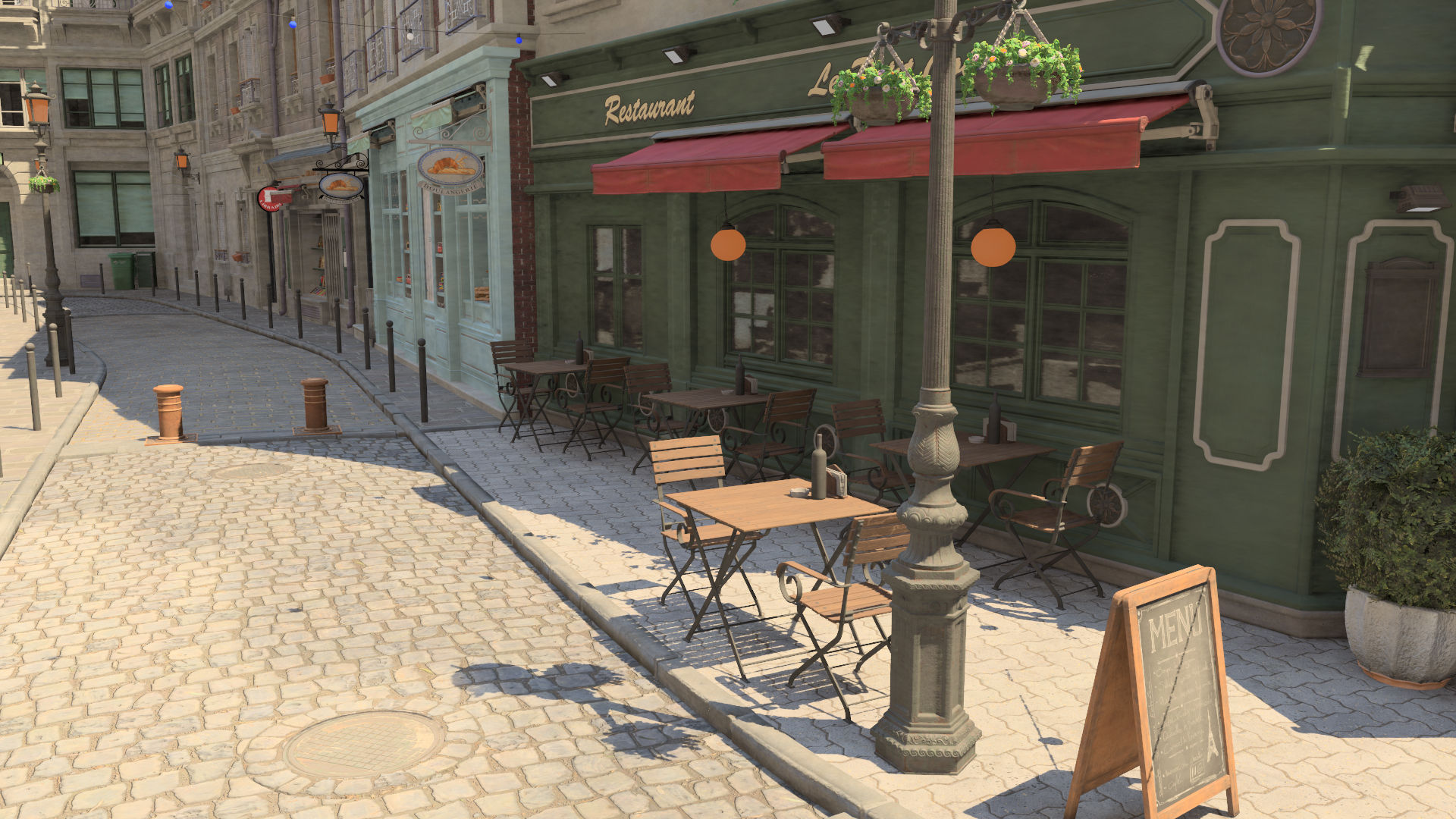} &
    \includegraphics[width=0.163\linewidth,trim={600px 0 600px 0},clip]{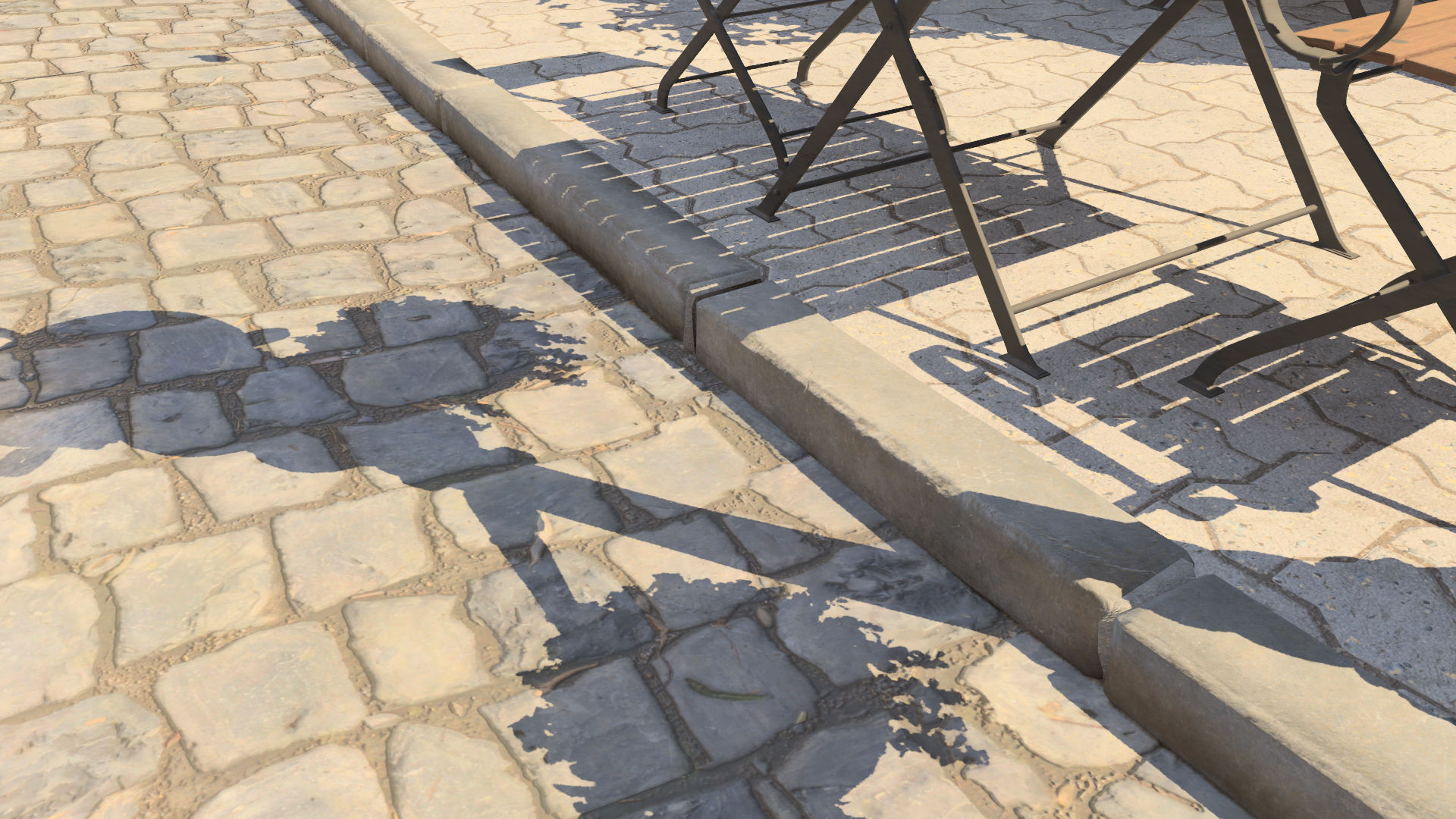} &
    \begin{overpic}[width=0.163\linewidth,trim={400px 0 800px 0},clip]{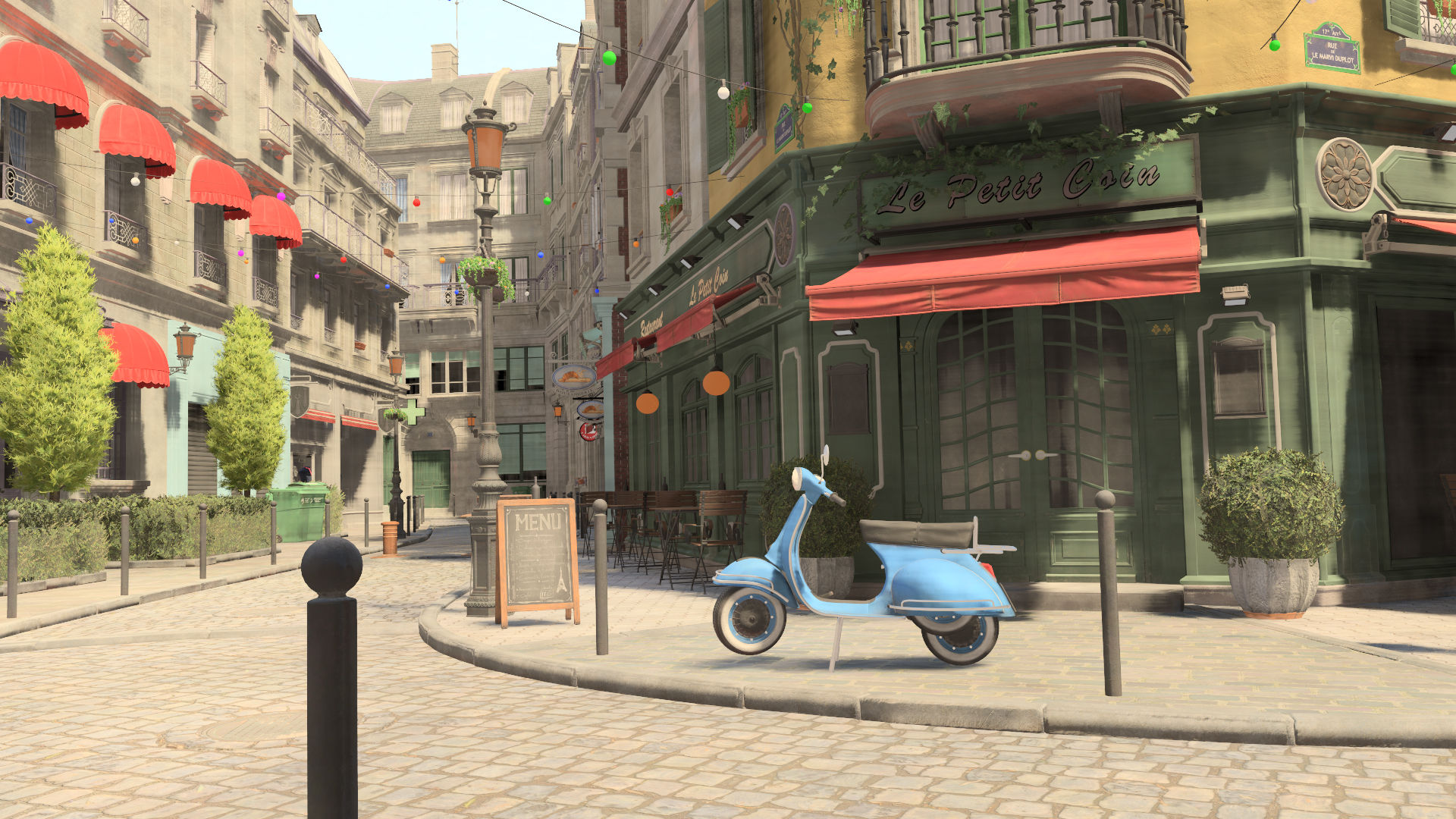}
      \put(14,48.75) { \makebox(0,0){\tikz\draw[red,thick] (0,0) rectangle (0.034\linewidth, 0.016\linewidth);} }
      \put(0.0,2.0) { \fcolorbox{red}{red}{\includegraphics[width=0.11\linewidth,trim={500px 500px 1280px 520px},clip]{nrc/zoom/render-far-freq.jpg}} }
    \end{overpic} &
    \includegraphics[width=0.163\linewidth,trim={600px 0 600px 0},clip]{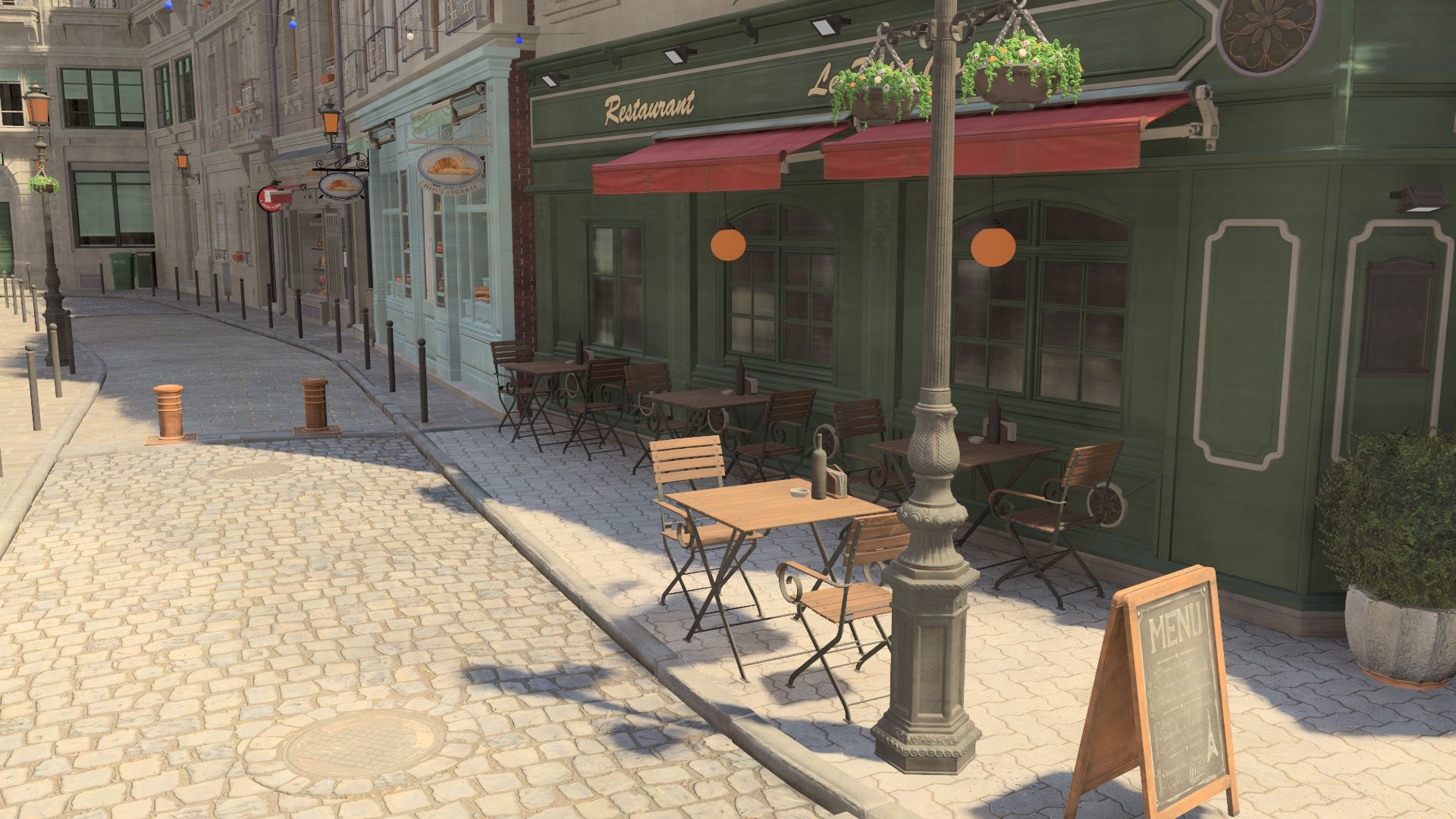} &
    \includegraphics[width=0.163\linewidth,trim={600px 0 600px 0},clip]{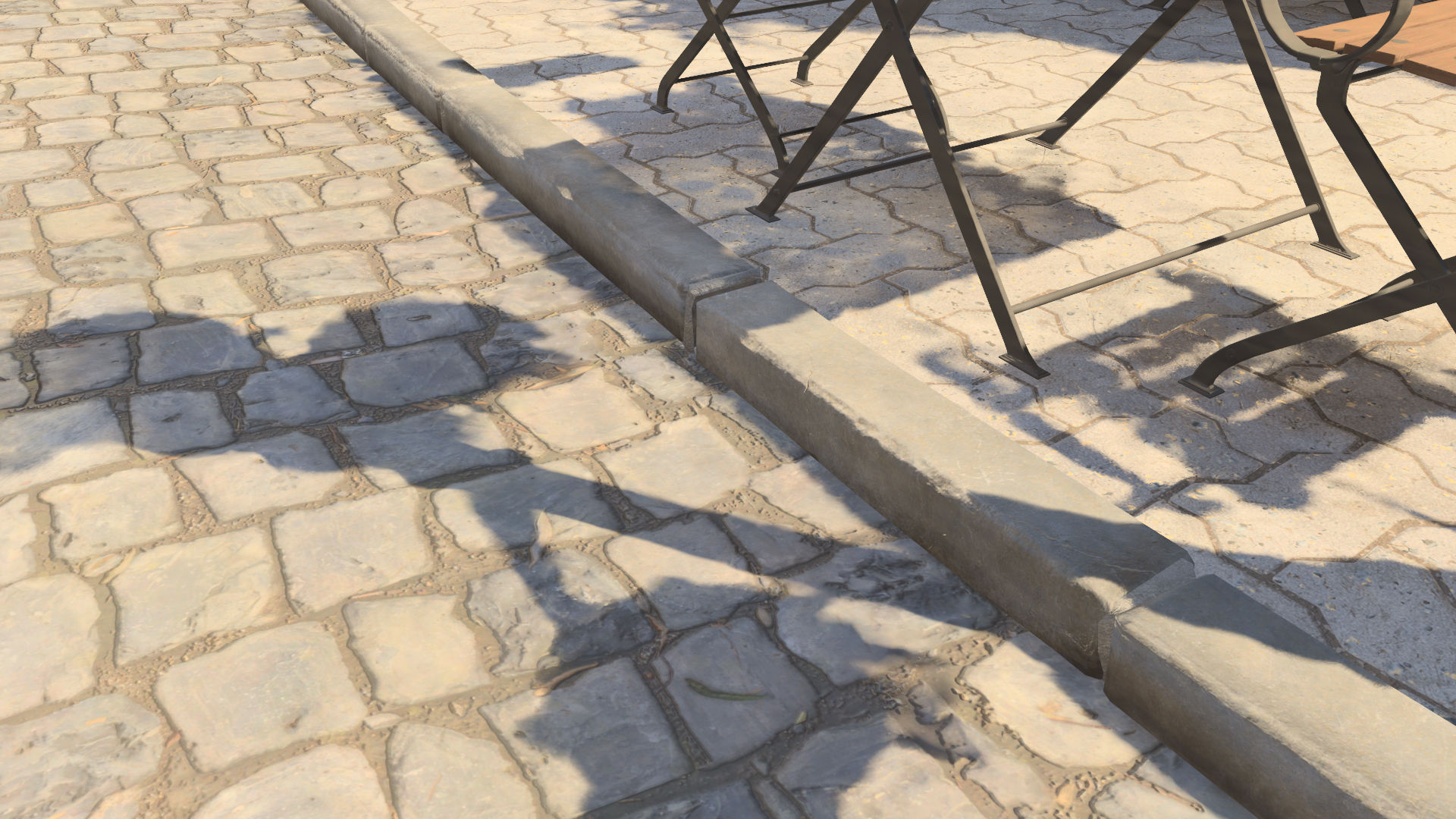}
  \end{tabularx}
  \vspace{-3mm}
  \caption{\label{fig:nrc-comparison}%
    Neural radiance caching~\cite{mueller2021realtime} gains much improved quality from the multiresolution hash encoding with only a mild performance penalty: 133 versus 147 frames per second at a resolution of ${1920\!\times\!1080}$px.
    To demonstrate the online adaptivity of the multiple hash resolutions vs.\ the prior triangle wave encoding, we show screenshots from a smooth camera motion that starts with a far-away view of the scene (left) and zooms onto a close-by view of an intricate shadow (right).
    Throughout the camera motion, which takes just a few seconds, the neural radiance cache continually learns from sparse camera paths, enabling the cache to learn (``overfit'') intricate detail at the scale of the content that the camera is momentarily observing.
  }\vspace{-3mm}
\end{figure*}

\subsection{Gigapixel Image Approximation}%
\label{Sec:Experiments:image_approximation}

Learning the 2D to RGB mapping of image coordinates to colors has become a popular benchmark for testing a model's ability to represent high-frequency detail~\citep{sitzmann2019siren,mueller2019nis,martel2021acorn,tancik2020fourfeat}.
Recent breakthroughs in adaptive coordinate networks (ACORN)~\citep{martel2021acorn} have shown impressive results when fitting very large images---up to a billion pixels---with high fidelity at even the smallest scales.
We target our multiresolution hash encoding at the same task and converge to high-fidelity images in seconds to minutes (\autoref{fig:image_t_sweep}).

For comparison, on the \sceneTokyo{} panorama from \autoref{fig:teaser}, ACORN achieves a PSNR of \SI{38.59}{\decibel} after \SI{36.9}{\hour} of training.
With a similar number of parameters (${\entriesPerLevel = 2^{24}}$), our method achieves the same PSNR after \num{2.5} \emph{minutes} of training, peaking at \SI{41.9}{\decibel} after \SI{4}{\minute}.
\autoref{fig:image_results_pearl} showcases the level of detail contained in our model for a variety of hash table sizes $\entriesPerLevel$ on another image.

It is difficult to directly compare the performance of our encoding to ACORN; a factor of ${\sim\!10}$ stems from our use of fully fused CUDA kernels, provided by the tiny-cuda-nn framework~\citep{tiny-cuda-nn}.
The input encoding allows for the use of a much smaller MLP than with ACORN, which accounts for much of the remaining ${10\times}$--${100\times}$ speedup.
That said, we believe that the biggest value-add of the multiresolution hash encoding is its simplicity.
ACORN relies on an adaptive subdivision of the scene as part of a learning curriculum, none of which is necessary with our encoding.

\subsection{Signed Distance Functions}\label{Sec:Experiments:sdf}

Signed distance functions (SDFs), in which a 3D shape is represented as the zero level-set of a function of position $\pos$, are used in many applications including simulation, path planning, 3D modeling, and video games.
\ADD{DeepSDF~\citep{park2019deepsdf} uses a large MLP to represent one or more SDFs at a time.
In contrast, when just a single SDF needs to be fit, a spatially learned encoding, such as ours can be employed and the MLP shrunk significantly.
This is the application we investigate in this section.
As baseline, we compare with NGLOD~\cite{takikawa2021nglod}, which achieves state-of-the-art results in both quality and speed by prefixing its small MLP with a lookup from an octree of trainable feature vectors.
Lookups along the hierarchy of this octree act similarly to our multiresolution cascade of grids: they are a collision-free analog to our technique, with a fixed growth factor ${\perLevelScale = 2}$.
To allow meaningful comparisons in terms of both performance and quality, we implemented an optimized version of NGLOD in our framework, details of which we describe in \autoref{app:nglod-implementation}.
Details pertaining to real-time training of SDFs are described in \autoref{app:sdf-datagen}.}

In \autoref{fig:sdf_results}, we compare NGLOD with our multiresolution hash encoding at roughly equal parameter count.
We also show a straightforward application of the frequency encoding~\cite{mildenhall2020nerf} to provide a baseline\ADD{, details of which are found in \autoref{app:sdf-freq-details}}.
By using a data structure tailored to the reference shape, NGLOD achieves the highest visual reconstruction quality.
However, even without such a dedicated data structure, our encoding approaches a similar fidelity to NGLOD in terms of the intersection-over-union metric (IoU\footnote{\ADD{IoU is the ratio of volumes of the interiors of the intersection and union of the pair of shapes being compared. IoU is always ${\leq 1}$ with a perfect fit corresponding to ${= 1}$.} We measure IoU by comparing the signs of the SDFs at 128 million points uniformly distributed within the bounding box of the scene.}) with similar performance and memory cost.
\begin{wrapfigure}{r}{0.35\linewidth}%
  \vspace{-5mm}%
  \hspace*{-6mm}\includegraphics[width=1.2\linewidth]{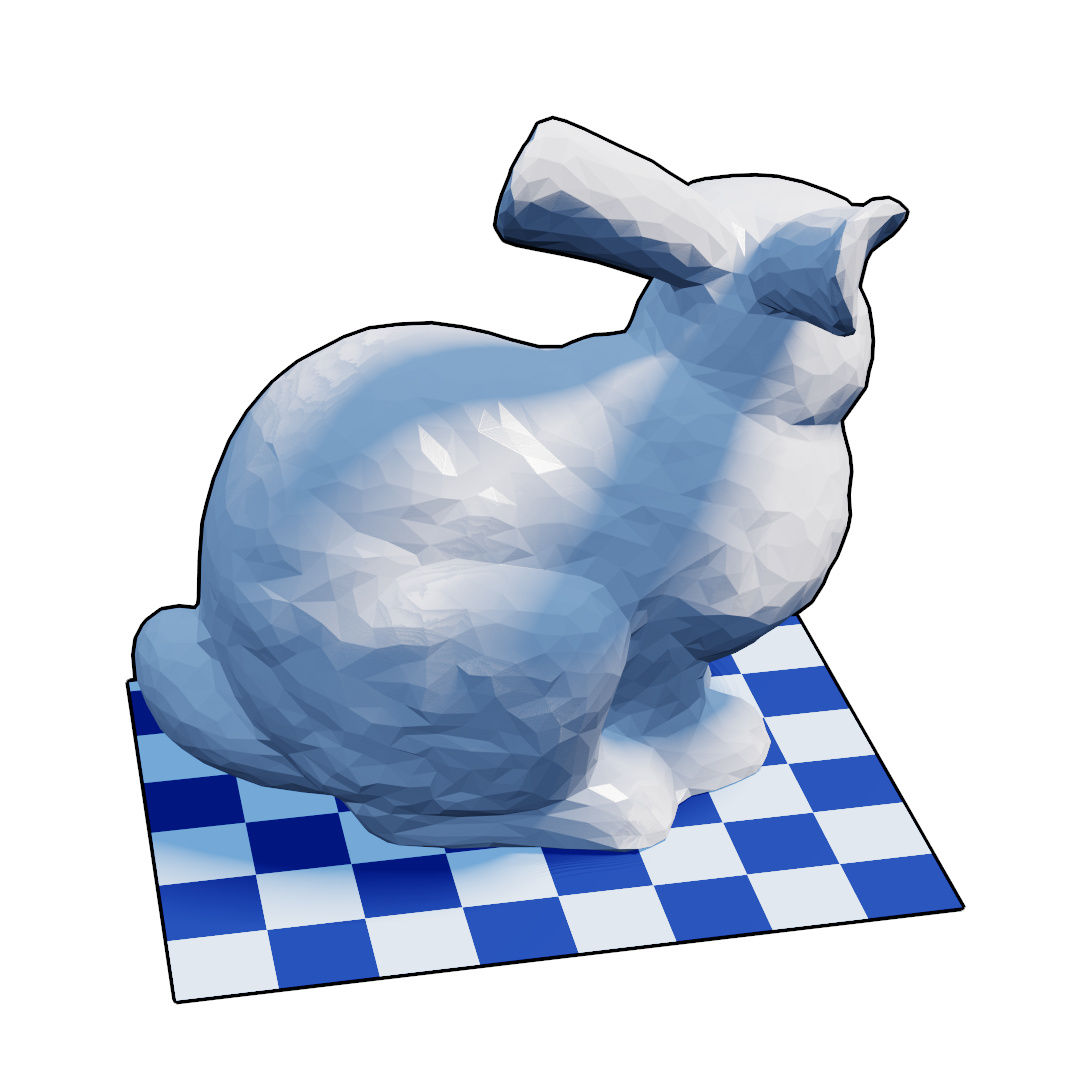}%
  \vspace{-4mm}%
\end{wrapfigure}%
Furthermore, the SDF is defined everywhere within the training volume, as opposed to NGLOD, which is only defined within the octree (i.e.\ close to the surface).
This permits the use of certain SDF rendering techniques such as approximate soft shadows from a small number of off-surface distance samples~\citep{evans2006softshadow}, as shown in the adjacent figure.

\ADD{To emphasize differences between the compared methods, we visualize the SDF using a shading model.
The resulting colors are sensitive to even slight changes in the surface normal, which emphasizes small fluctuations in the prediction more strongly than in other graphics primitives where color is predicted directly.
This sensitivity reveals undesired microstructure in our hash encoding on the scale of the finest grid resolution, which is absent in NGLOD and does not disappear with longer training times.
Since NGLOD is essentially a collision-free analog to our hash encoding, we attribute this artifact to hash collisions.
Upon close inspection, similar microstructure can be seen in other neural graphics primitives, although with significantly lower magnitude.}

\begin{figure}
  \sffamily
  \small
  \pgfplotsset{width=9.0cm, height=4.0cm}
  \vspace{-2.2mm}
  \hspace{-2mm}
  \begin{tikzpicture}
\tikzstyle{every node}=[font=\footnotesize]

\definecolor{color0}{rgb}{1,0.533333333333333,0.533333333333333}
\definecolor{color1}{rgb}{1,0.8,0.266666666666667}
\definecolor{color2}{rgb}{0.533333333333333,0.8,1}

\begin{axis}[
legend cell align={left},
legend style={
  fill opacity=0.8,
  draw opacity=1,
  text opacity=1,
  at={(0.97,0.03)},
  anchor=south east,
  draw=white!80!black,
  nodes={scale=0.75, transform shape}
},
tick align=outside,
tick pos=left,
title={{\small \sffamily Neural Radiance Field: \sceneLego}},
title style={at={(0.5,0.93)}},
x grid style={white!69.0196078431373!black},
xlabel={Training time (seconds)},
xmin=195.985047805309, xmax=777.987818491459,
xtick style={color=black},
y grid style={white!69.0196078431373!black},
ylabel={PSNR (dB)},
ymin=35.0719980023238, ymax=36.5867545285986,
ytick style={color=black}
]
\addplot [semithick, color0, mark=*, mark size=1, mark options={solid}]
table {%
222.772186994553 35.1408505716999
222.439719200134 35.7217079502554
243.241554498672 36.0052576758842
279.554010391235 36.243565324869
412.660806894302 36.26144806088
};
\addlegendentry{$N_\mathrm{layers}=1$}
\addplot [semithick, color1, mark=*, mark size=1, mark options={solid}]
table {%
226.028396368027 35.3159206831678
226.953627824783 35.8484963859825
247.891683340073 36.2100016867744
324.485737800598 36.3951691878594
594.732287883759 36.4161708629578
};
\addlegendentry{$N_\mathrm{layers}=2$}
\addplot [semithick, color2, mark=*, mark size=1, mark options={solid}]
table {%
225.336002111435 35.1770899611643
232.406004190445 35.9588434010099
263.312616586685 36.2269686218852
346.350400924683 36.5179019592225
751.533147096634 36.4782679346175
};
\addlegendentry{$N_\mathrm{layers}=3$}
\draw (axis cs:232.26661798954,35.1408505716999) node[
  scale=0.7,
  anchor=base west,
  text=white!15!black,
  rotate=0.0
]{$N_\mathrm{neurons} = 16$};
\draw (axis cs:231.934150195122,35.6768840506882) node[
  scale=0.7,
  anchor=base west,
  text=white!15!black,
  rotate=0.0
]{$N_\mathrm{neurons} = 32$};
\draw (axis cs:243.241554498672,35.9156098767498) node[
  scale=0.7,
  anchor=base west,
  text=white!15!black,
  rotate=0.0
]{$N_\mathrm{neurons} = 64$};
\draw (axis cs:279.554010391235,36.0866816763838) node[
  scale=0.7,
  anchor=base west,
  text=white!15!black,
  rotate=0.0
]{$N_\mathrm{neurons} = 128$};
\draw (axis cs:412.660806894302,36.1045644123948) node[
  scale=0.7,
  anchor=base west,
  text=white!15!black,
  rotate=0.0
]{$N_\mathrm{neurons} = 256$};
\end{axis}

\end{tikzpicture}
  \vspace{-4.1mm}
  \caption{\label{fig:nerf_network_sweep}%
    The effect of the MLP size on test error vs.\ training time (\num{31000} training steps) on the \sceneLego{} scene.
    Other scenes behave almost identically.
    Each curve represents a different MLP depth, where the color MLP has $N_\mathrm{layers}$ hidden layers and the density MLP has \num{1} hidden layer; we do not observe an improvement with deeper density MLPs.
    The curves sweep the number of neurons the hidden layers of the density and color MLPs from \num{16} to \num{256}.
    Informed by this analysis, we choose ${N_\mathrm{layers} = 2}$ and ${N_\mathrm{neurons} = 64}$.
  }\vspace{-2mm}
\end{figure}
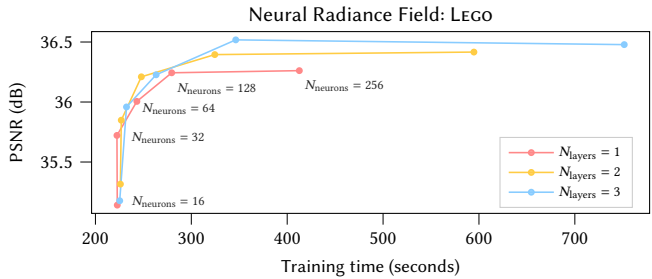

\subsection{Neural Radiance Caching}\label{Sec:Experiments:nrc}

In neural radiance caching~\cite{mueller2021realtime}, the task of the MLP is to predict photorealistic pixel colors from feature buffers; see \autoref{fig:nrc-algorithm}.
The MLP is run independently for each pixel (i.e.\ the model is not convolutional), so the feature buffers can be treated as per-pixel feature vectors that contain the 3D coordinate $\pos$ as well as additional features.
We can therefore directly apply our multiresolution hash encoding to $\pos$ while treating all additional features as auxiliary encoded dimensions $\auxDims$ to be concatenated with the encoded position, using the same encoding as \citet{mueller2021realtime}.
We integrated our work into M\"uller et al.'s implementation of neural radiance caching and therefore refer to their paper for implementation details.

For photorealistic rendering, the neural radiance cache is typically queried only for \emph{indirect} path contributions, which masks its reconstruction error behind the first reflection.
In contrast, we would like to \emph{emphasize} the neural radiance cache's error, and thus the improvement that can be obtained by using our multiresolution hash encoding, so we directly visualize the neural radiance cache at the first path vertex.

\autoref{fig:nrc-comparison} shows that---compared to the triangle wave encoding of \citet{mueller2021realtime}---our encoding results in sharper reconstruction while incurring only a mild performance overhead of \SI{0.7}{\milli\second} that reduces the frame rate from \num{147} to \num{133} FPS at a resolution of $1920\times1080$px.
Notably, the neural radiance cache is trained online---during rendering---from a path tracer that runs in the background, which means that the \SI{0.7}{\milli\second} overhead includes \emph{both} training and runtime costs of our encoding.

\subsection{Neural Radiance and Density Fields (NeRF)}\label{Sec:Experiments:nerf}

In the NeRF setting, a volumetric shape is represented in terms of a spatial (3D) density function and a spatiodirectional (5D) emission function, which we represent by a similar neural network architecture as \citet{mildenhall2020nerf}.
We train the model in the same ways as Mildenhall et al.:\ by backpropagating through a differentiable ray marcher driven by 2D RGB images from known camera poses.

\paragraph{Model Architecture.}
Unlike the other three applications, our NeRF model consists of two concatenated MLPs: a density MLP followed by a color MLP~\citep{mildenhall2020nerf}.
The density MLP maps the hash encoded position ${\encOut = \enc(\pos; \Params)}$ to $16$ output values, the first of which we treat as log-space density.
The color MLP adds view-dependent color variation.
Its input is the concatenation of
\begin{itemize}[leftmargin=*]
  \item the $16$ output values of the density MLP, and
  \item the view direction projected onto the first $16$ coefficients of the spherical harmonics basis (i.e.\ up to degree $4$). This is a natural frequency encoding over unit vectors.
\end{itemize}
Its output is an RGB color triplet, for which we use either a sigmoid activation when the training data has low dynamic-range (sRGB) or an exponential activation when it has high dynamic range (linear HDR).
We prefer HDR training data due to the closer resemblance to physical light transport.
This brings numerous advantages as has also been noted in concurrent work~\cite{mildenhall2021nerf}.

Informed by the analysis in \autoref{fig:nerf_network_sweep}, our results were generated with a 1-hidden-layer density MLP and a 2-hidden-layer color MLP, both \num{64} neurons wide.

\begin{figure}
  \vspace{-2mm}
  {\input{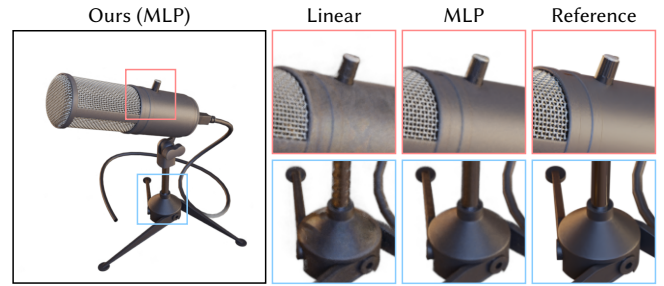}
  \vspace{-2mm}
  \caption{\label{fig:mlp_vs_linear}
    Feeding the result of our encoding through a linear transformation (no neural network) versus an MLP when learning a NeRF.\@
    The models were trained for \trainingTimeMLPorNot\si{\minute}.
    The MLP allows for resolving specular details and reduces the amount of background noise caused by hash collisions.
    Due to the small size and efficient implementation of the MLP, it is only 15\% more expensive---well worth the significantly improved quality.}
  }\vspace{-2mm}
\end{figure}

\begin{table*}
  \vspace{-0mm}
  \caption{\label{tab:nerf}
    Peak signal to noise ratio (PSNR) of \ADD{our NeRF implementation with multiresolution hash encoding (``Ours: Hash'')} vs.\ NeRF~\citep{mildenhall2020nerf}, mip-NeRF~\citep{barron2021mipnerf}, and NSVF~\citep{liu2020neural}, which require $\sim$hours to train \ADD{(values taken from the respective papers).}
    To demonstrate the comparatively rapid training of our method, we list its results after training for \SI{1}{\second} to \SI{5}{\minute}.
    For each scene, we mark the methods with least error using gold \tikzcircle[gold,fill=gold]{2pt}, silver \tikzcircle[silver,fill=silver]{2pt}, and bronze \tikzcircle[bronze,fill=bronze]{2pt} medals.
    \ADD{To analyze the degree to which our speedup originates from our optimized implementation vs.\ from our hash encoding, we also report PSNR for a nearly identical version of our implementation, in which the hash encoding has been replaced by the frequency encoding and the MLP correspondingly enlarged to match \citet{mildenhall2020nerf} (``Ours: Frequency''; details in \autoref{app:sdf-freq-details}).}
    It approaches NeRF's quality after training for just ${\sim\!\SI{5}{\minute}}$, yet is outperformed by our full method after training for $\SI{5}{\second}$--$\SI{15}{\second}$, \ADD{amounting to a $20$--${60\times}$ improvement that can be attributed to the hash encoding}.
  }\vspace{-2mm}
  \small \sffamily
\rowcolors{2}{gray!8}{white}
\setlength{\tabcolsep}{9.9pt}
\begin{tabularx}{\linewidth}{llllllllll}
\toprule
\rowcolor{white}
& \sceneMic& \sceneFicus& \sceneChair& \sceneHotdog& \sceneMaterials& \sceneDrums& \sceneShip& \sceneLego& avg.\\
\midrule
\rowcolor{white}

Ours: Hash (\SI{1}{\second})& $26.09$& $21.30$& $21.55$& $21.63$& $22.07$& $17.76$& $20.38$& $18.83$& 	21.202\\
Ours: Hash (\SI{5}{\second})& $32.60$& $30.35$& $30.77$& $33.42$& $26.60$& $23.84$& $26.38$& $30.13$& 	29.261\\
Ours: Hash (\SI{15}{\second})& $34.76$& $32.26$& $32.95$& $35.56$& $28.25$& $25.23$& $28.56$& $33.68$& 	31.407\\
Ours: Hash (\SI{1}{\minute})& $35.92$ \tikzcircle[bronze,fill=bronze]{2pt}& $33.05$ \tikzcircle[bronze,fill=bronze]{2pt}& $34.34$ \tikzcircle[bronze,fill=bronze]{2pt}& $36.78$& $29.33$& $25.82$ \tikzcircle[silver,fill=silver]{2pt}& $30.20$ \tikzcircle[bronze,fill=bronze]{2pt}& $35.63$ \tikzcircle[bronze,fill=bronze]{2pt}& 	32.635 \tikzcircle[bronze,fill=bronze]{2pt}\\
Ours: Hash (\SI{5}{\minute})& $36.22$ \tikzcircle[silver,fill=silver]{2pt}& $33.51$ \tikzcircle[gold,fill=gold]{2pt}& $35.00$ \tikzcircle[silver,fill=silver]{2pt}& $37.40$ \tikzcircle[silver,fill=silver]{2pt}& $29.78$ \tikzcircle[bronze,fill=bronze]{2pt}& $26.02$ \tikzcircle[gold,fill=gold]{2pt}& $31.10$ \tikzcircle[gold,fill=gold]{2pt}& $36.39$ \tikzcircle[gold,fill=gold]{2pt}& 	33.176 \tikzcircle[gold,fill=gold]{2pt}\\
\midrule
mip-NeRF ($\sim$hours)& $36.51$ \tikzcircle[gold,fill=gold]{2pt}& $33.29$ \tikzcircle[silver,fill=silver]{2pt}& $35.14$ \tikzcircle[gold,fill=gold]{2pt}& $37.48$ \tikzcircle[gold,fill=gold]{2pt}& $30.71$ \tikzcircle[silver,fill=silver]{2pt}& $25.48$ \tikzcircle[bronze,fill=bronze]{2pt}& $30.41$ \tikzcircle[silver,fill=silver]{2pt}& $35.70$ \tikzcircle[silver,fill=silver]{2pt}& 	33.090 \tikzcircle[silver,fill=silver]{2pt}\\
NSVF ($\sim$hours)& $34.27$& $31.23$& $33.19$& $37.14$ \tikzcircle[bronze,fill=bronze]{2pt}& $32.68$ \tikzcircle[gold,fill=gold]{2pt}& $25.18$& $27.93$& $32.29$& 	31.739\\
NeRF ($\sim$hours)& $32.91$& $30.13$& $33.00$& $36.18$& $29.62$& $25.01$& $28.65$& $32.54$& 	31.005\\
\midrule

Ours: Frequency (\SI{5}{\minute})& $31.89$& $28.74$& $31.02$& $34.86$& $28.93$& $24.18$& $28.06$& $32.77$& 	30.056\\
Ours: Frequency (\SI{1}{\minute})& $26.62$& $24.72$& $28.51$& $32.61$& $26.36$& $21.33$& $24.32$& $28.88$& 	26.669\\
\bottomrule
\end{tabularx}

  \vspace{-2mm}
\end{table*}

\paragraph{Accelerated ray marching.}
When marching along rays for both training and rendering, we would like to place samples such that they contribute somewhat uniformly to the image, minimizing wasted computation.
Thus, we concentrate samples near surfaces by maintaining an occupancy grid that coarsely marks empty vs.\ non-empty space.
In large scenes, we additionally cascade the occupancy grid and distribute samples exponentially rather than uniformly along the ray.
\autoref{app:nerf-occupancy} describes these procedures in detail.

\begin{figure}
  \setlength{\tabcolsep}{1pt}%
  \vspace{1mm}
  \hspace*{-0.3mm}\begin{tabular}{cc}
    \includegraphics[width=0.495\linewidth]{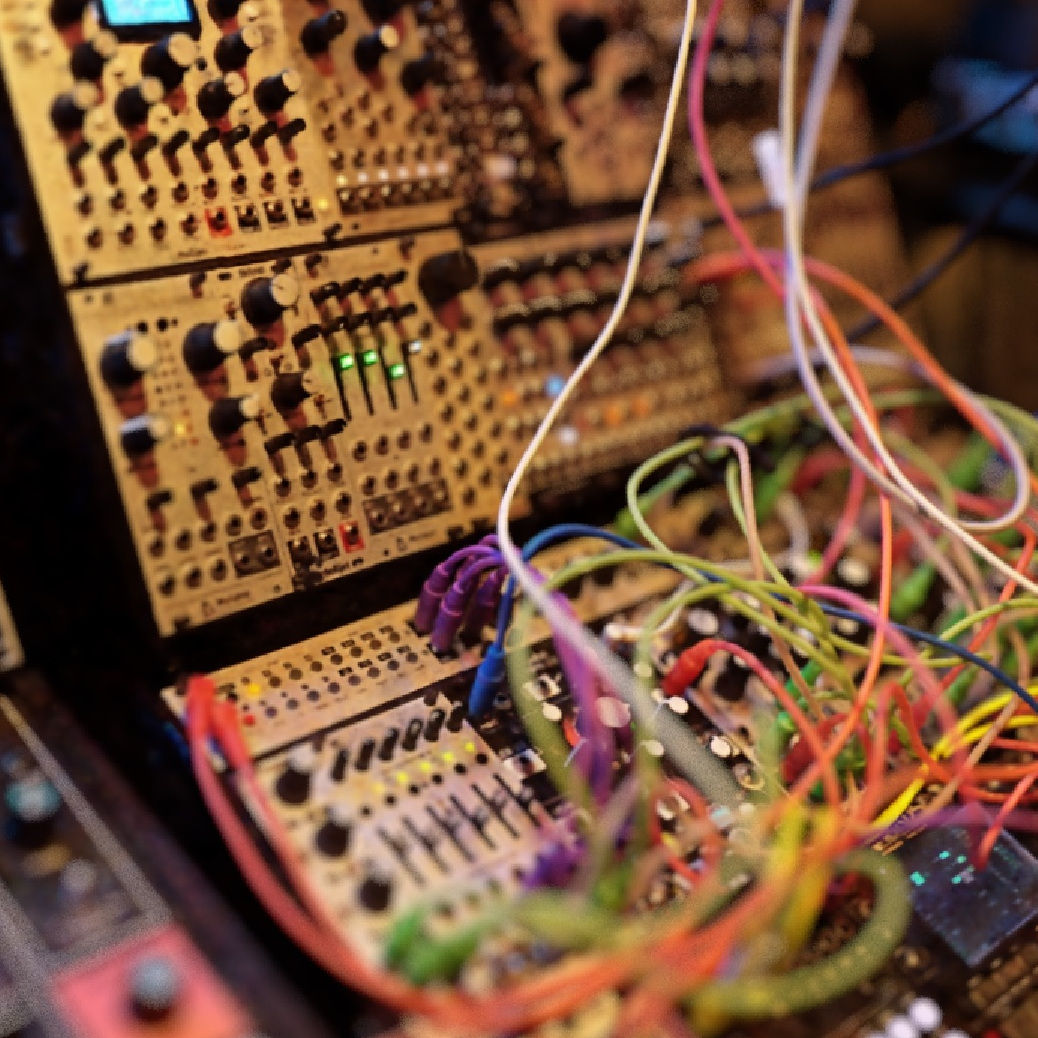} &
    \includegraphics[width=0.495\linewidth]{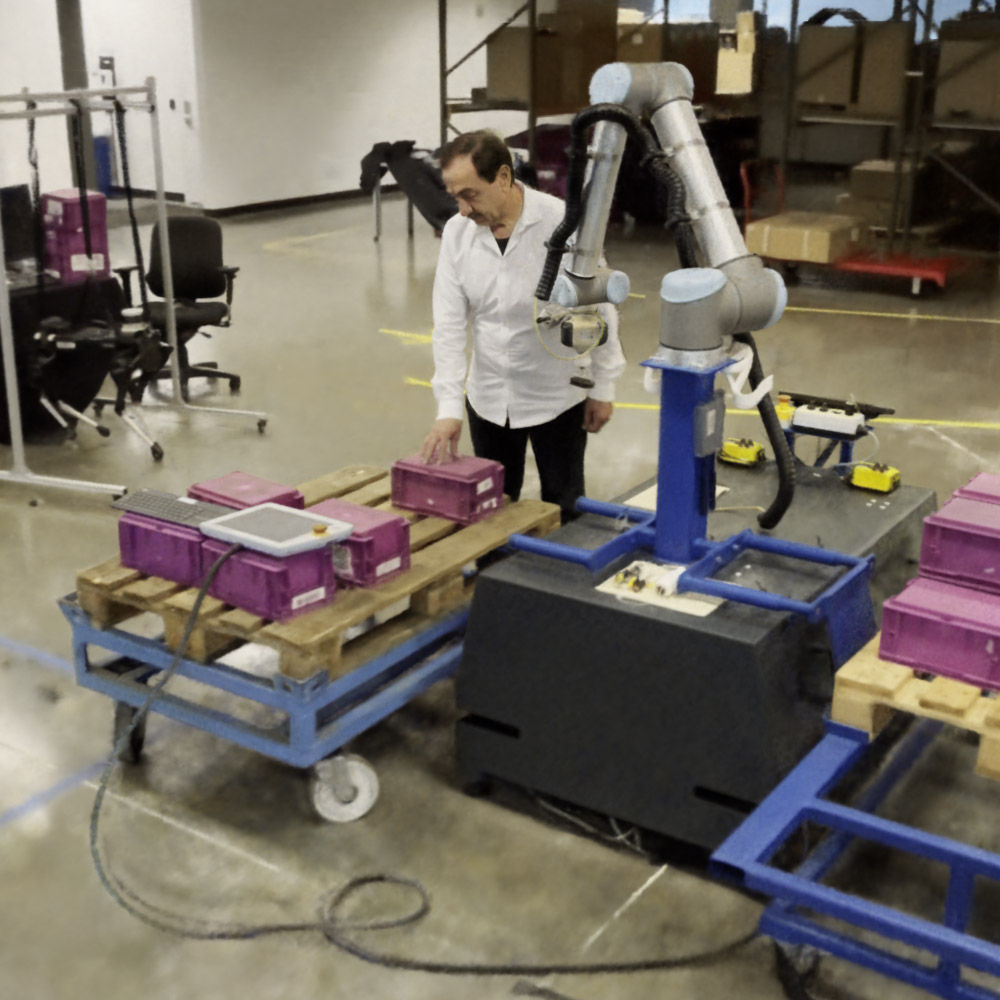}
  \end{tabular}
  \vspace{-3mm}
  \caption{\label{fig:modular-and-360}
    NeRF reconstruction of a modular synthesizer and large natural 360 scene. The left image took 5 seconds to accumulate 128 samples at 1080p on a single RTX 3090 GPU, allowing for brute force defocus effects. The right image was taken from an interactive session running at 10 frames per second on the same GPU.\@
  }
  \vspace{-2mm}
\end{figure}

At HD resolutions, synthetic and even real-world scenes can be trained in seconds and rendered at \num{60} FPS, without the need of caching of the MLP outputs \citep{garbin2021fastnerf,yu2021plenoctrees,Wizadwongsa2021NeX}.
This high performance makes it tractable to add effects such as anti-aliasing, motion blur and depth of field by brute-force tracing of multiple rays per pixel, as shown in \autoref{fig:modular-and-360}.

\paragraph{Comparison with direct voxel lookups.}
\autoref{fig:mlp_vs_linear} shows an ablation where we replace the entire neural network with a single linear matrix multiplication, in the spirit of (although not identical to) concurrent direct voxel-based NeRF~\cite{yu2021plenoxels,sun2021direct}.
While the linear layer is capable of reproducing view-dependent effects, the quality is significantly compromised as compared to the MLP, which is better able to capture specular effects and to resolve hash collisions across the interpolated multiresolution hash tables (which manifest as high-frequency artifacts).
Fortunately, the MLP is only 15\% more expensive than the linear layer, thanks to its small size and efficient implementation.

\paragraph{Comparison with high-quality offline NeRF}
In \autoref{tab:nerf}, we compare the peak signal to noise ratio (PSNR) \ADD{our NeRF implementation with multiresolution hash encoding (``Ours: Hash'')} with that of NeRF~\citep{mildenhall2020nerf}, mip-NeRF~\citep{barron2021mipnerf}, and NSVF~\citep{liu2020neural}, which all require on the order of hours to train.
In contrast, we list results of our method after training for \SI{1}{\second} to \SI{5}{\minute}.
Our PSNR is competitive with NeRF and NSVF after just \SI{15}{\second} of training, and competitive with mip-NeRF after \SI{1}{\minute} to \SI{5}{\minute} of training.

On one hand, our method performs best on scenes with high geometric detail, such as \sceneFicus{}, \sceneDrums{}, \sceneShip{} and \sceneLego{}, achieving the best PSNR of all methods.
On the other hand, mip-NeRF and NSVF outperform our method on scenes with complex, view-dependent reflections, such as \sceneMaterials{}; we attribute this to the much smaller MLP that we necessarily employ to obtain our speedup of several orders of magnitude over these competing implementations.

\ADD{Next, we analyze the degree to which our speedup originates from our efficient implementation versus from our encoding.
To this end, we additionally report PSNR for a nearly identical version of our implementation: we replace the hash encoding by the frequency encoding and enlarge the MLP to approximately match the architecture of \citet{mildenhall2020nerf} (``Ours: Frequency''); see \autoref{app:sdf-freq-details} for details.
This version of our algorithm approaches NeRF's quality after training for just ${\sim\!\SI{5}{\minute}}$, yet is outperformed by our full method after training for a much shorter duration ($\SI{5}{\second}$--$\SI{15}{\second}$), amounting to a $20$--${60\times}$ improvement caused by the hash encoding and smaller MLP.\@}

\ADD{For ``Ours: Hash'', the cost of each training step is roughly constant at $\sim$\SI{6}{\milli\second} per step.
This amounts to \SI{50}{\kilo\nothing} steps after \SI{5}{\minute} at which point the model is well converged.
We decay the learning rate after \SI{20}{\kilo\nothing} steps by a factor of \num{0.33}, which we repeat every further \SI{10}{\kilo\nothing} steps.}
\ADD{In contrast, the larger MLP used in ``Ours: Frequency'' requires $\sim$\SI{30}{\milli\second} per training step, meaning that the PSNR listed after \SI{5}{\minute} corresponds to about \SI{10}{\kilo\nothing} steps.
It could thus keep improving slightly if trained for extended periods of time, as in the offline NeRF variants that are often trained for several \SI{100}{\kilo\nothing} steps.}

\begin{figure*}
  \small
  \vspace{0mm}
  \setlength{\tabcolsep}{1pt}%
  \hspace*{-5mm}\begin{tabular}{ccc}
    \includegraphics[width=0.33\linewidth,trim={70px 270px 70px 230px},clip]{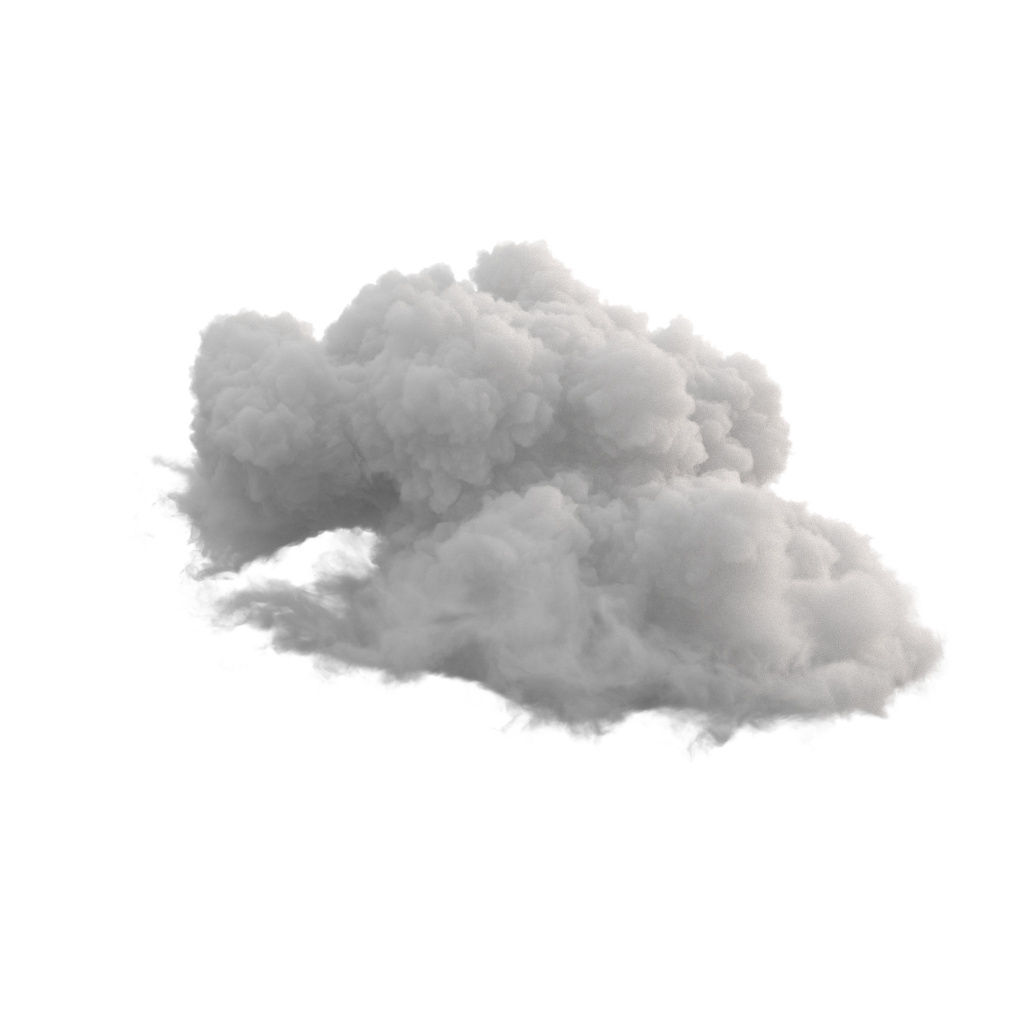} &
    \includegraphics[width=0.33\linewidth,trim={70px 270px 70px 230px},clip]{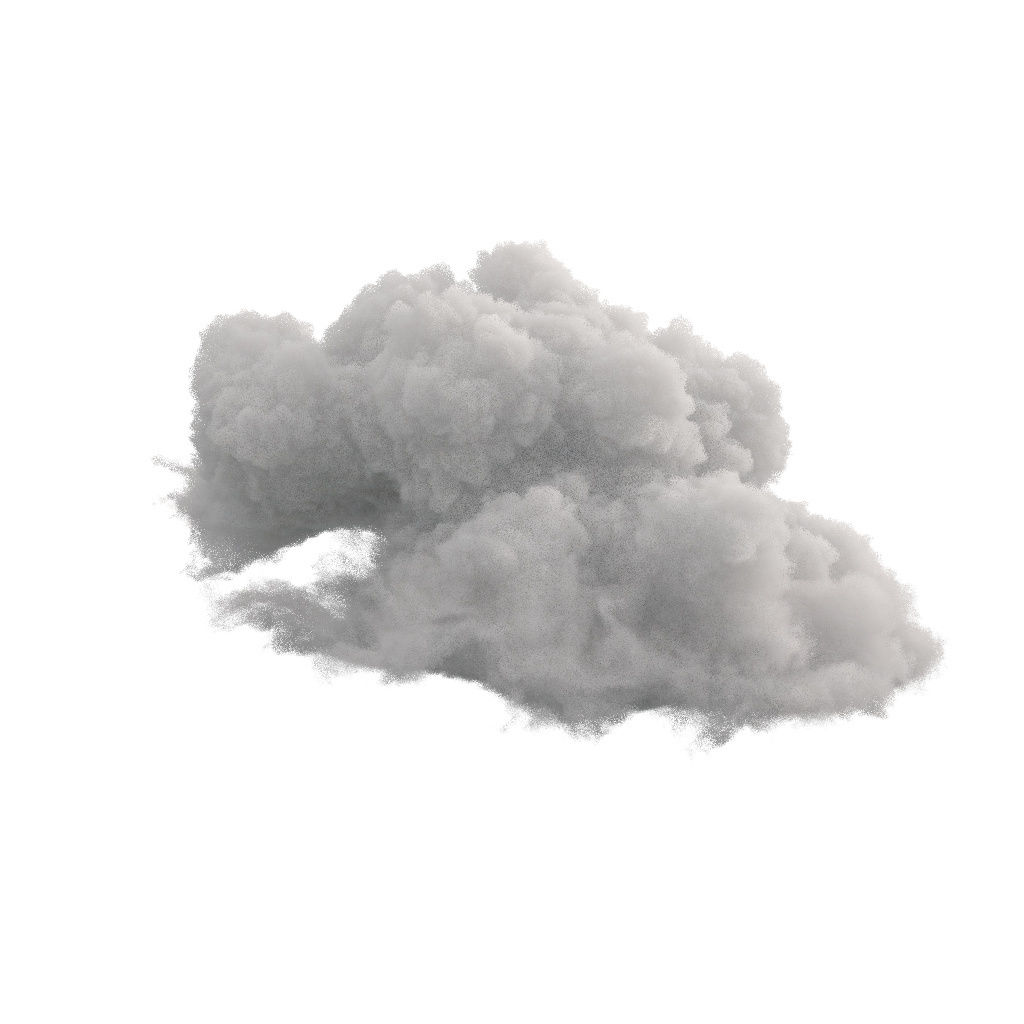} &
    \includegraphics[width=0.33\linewidth,trim={70px 270px 70px 230px},clip]{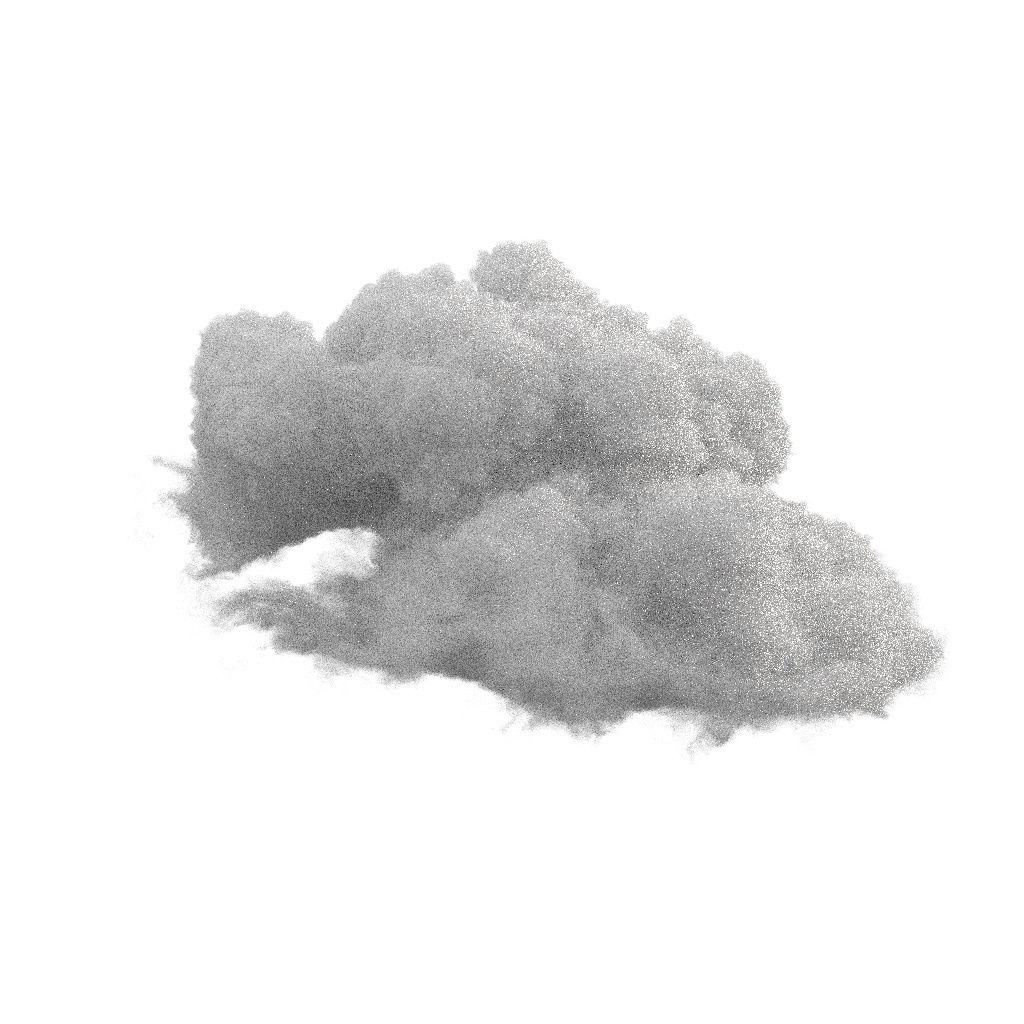} \\[0mm]
    \textbf{(a)} Offline rendered reference &
    \textbf{(b)} Hash (ours), trained for \SI{10}{\second} &
    \textbf{(c)} Path tracer \\
    & \ADD{Rendered in \SI{32}{\milli\second} (\num{2} samples per pixel)} & \ADD{Rendered in \SI{32}{\milli\second} (\num{16} samples per pixel)}
  \end{tabular}
  \vspace{-1mm}
  \caption{\label{fig:cloud}
    Preliminary results of training a NeRF cloud model \textbf{(b)} from real-time path tracing data.
    Within \SI{32}{\milli\second}, a ${1024\!\times\!1024}$ image of our model convincingly approximates the offline rendered ground truth \textbf{(a)}.
    Our model exhibits less noise than a GPU path tracer that ran for an equal amount of time \textbf{(c)}. The cloud data is \copyright Walt Disney Animation Studios \href{http://creativecommons.org/licenses/by-sa/3.0/}{(CC BY-SA 3.0)}
  }
  \vspace{-0mm}
\end{figure*}

\ADD{While we isolated the performance and convergence impact of our hash encoding and its small MLP, we believe an additional study is required to quantify the impact of advanced ray marching schemes (such as ours, coarse-fine~\citep{mildenhall2020nerf}, or DONeRF~\citep{neff2021donerf}) independently from the encoding and network architecture.
We report additional information in \autoref{app:nerf-batch-size} to aid in such an analysis.}

\section{Discussion and Future Work}\label{Sec:Discussion}

\paragraph{Concatenation vs.\ reduction.}
At the end of the encoding, we \emph{concatenate} rather than reduce (for example, by summing) the $\featuresPerEntry$-di\-men\-sional feature vectors obtained from each resolution.
We prefer concatenation for two reasons.
First, it allows for independent, fully parallel processing of each resolution.
Second, a reduction of the dimensionality of the encoded result $\encOut$ from $\levels\featuresPerEntry$ to $\featuresPerEntry$ may be too small to encode useful information.
While $\featuresPerEntry$ could be increased proportionally, it would make the encoding much more expensive.

However, we recognize that there may be applications in which reduction is favorable, such as when the neural network is significantly more expensive than the encoding, in which case the added computational cost of increasing $\featuresPerEntry$ could be insignificant.
We thus argue for concatenation \emph{by default} and not as a hard-and-fast rule.
In our applications, concatenation, coupled with ${\featuresPerEntry=2}$ always yielded by far the best results.

\paragraph{\ADD{Choice of hash function.}}
\ADD{A good hash function is efficient to compute, leads to coherent look-ups, and uniformly covers the feature vector array regardless of the structure of query points.
We chose our hash function for its good mixture of these properties and also experimented with three others:}
\ADD{\begin{enumerate}[leftmargin=*]
  \item The PCG32~\citep{oneill:pcg2014} RNG, which has superior statistical properties. Unfortunately, it did not yield a higher-quality reconstruction, making its higher cost not worthwhile.
  \item Ordering the least significant bits of $\Z^d$ by a space-filling curve and only hashing the higher bits. This leads to better look-up coherence at the cost of worse reconstruction quality. However, the speed-up is only marginally better than setting ${\primeNumber_1 := 1}$ as done in our hash, and is thus not worth the reduced quality.
  \item Even better coherence can be achieved by treating the hash function as a tiling of space into dense grids. Like (2), the speed-up is small in practice with significant detriment to quality.
\end{enumerate}}

Alternatively to hand-crafted hash functions, it is conceivable to \emph{optimize} the hash function in future work, turning the method into a dictionary-learning approach.
Two possible avenues are \textbf{(1)}~developing a continuous formulation of indexing that is amenable to analytic differentiation or \textbf{(2)}~applying an evolutionary optimization algorithm that can efficiently explore the discrete function space.

\paragraph{Microstructure due to hash collisions.}
The salient artifact of our encoding is a small amount of ``grainy'' microstructure, most visible on the learned signed distance functions (\autoref{fig:teaser} and \autoref{fig:sdf_results}).
The graininess is a result of hash collisions that the MLP is unable to fully compensate for.
We believe that the key to achieving state-of-the-art quality on SDFs with our encoding will be to find a way to overcome this microstructure, for example by filtering hash table lookups or by imposing an additional smoothness prior on the loss.

\paragraph{Generative setting} Parametric input encodings, when used in a generative setting, typically arrange their features in a dense grid which can then be populated by a separate generator network, typically a CNN such as StyleGAN \cite{chan2021efficient, devries2021unconstrained, peng2020convolutional}.  Our hash encoding adds an additional layer of complexity, as the features are not arranged in a regular pattern through the input domain; that is, the features are not bijective with a regular grid of points. We leave it to future work to determine how best to overcome this difficulty.

\paragraph{Other applications.}
We are interested in applying the multiresolution hash encoding to other low-dimensional tasks that require accurate, high-frequency fits.
The frequency encoding originated from the attention mechanism of transformer networks~\citep{vaswani2017attention}.
We hope that parametric encodings such as ours can lead to a meaningful improvement in general, attention-based tasks.

Heterogenous volumetric density fields, such as cloud and smoke stored in a VDB \cite{vdb,museth2021} data structure, often include empty space on the outside, a solid core on the inside, and sparse detail on the volumetric surface.
This makes them a good fit for our encoding.
In the code released alongside this paper, we have included a preliminary implementation that fits a radiance and density field directly from the noisy output of a volumetric path tracer.
The initial results are promising, as shown in \autoref{fig:cloud}, and we intend to pursue this direction further in future work.

\section{Conclusion}

Many graphics problems rely on task specific data structures to exploit the sparsity or smoothness of the problem at hand.
Our multi-resolution hash encoding provides a practical learning-based alternative that automatically focuses on relevant detail, independent of the task.
Its low overhead allows it to be used even in time-constrained settings like online training and inference.
In the context of neural network input encodings, it is a drop-in replacement, for example speeding up NeRF by several orders of magnitude and matching the performance of concurrent non-neural 3D reconstruction techniques.

Slow computational processes in any setting, from lightmap baking to the training of neural networks, can lead to frustrating workflows due to long iteration times \cite{endertonworkflow}.
We have demonstrated that single-GPU training times measured in seconds are within reach for many graphics applications, allowing neural approaches to be applied where previously they may have been discounted.

\begin{acks}
We are grateful to
Andrew Tao,
Andrew Webb,
Anjul Patney,
David Luebke,
Fabrice Rousselle,
Jacob Munkberg,
James Lucas,
Jonathan Granskog,
Jonathan Tremblay,
Koki Nagano,
Marco Salvi,
Nikolaus Binder, and
Towaki Takikawa
for profound discussions, proofreading, feedback, and early testing.
We also thank Arman Toorians and Saurabh Jain for the factory robot dataset in \autoref{fig:modular-and-360} (right).
\end{acks}

\bibliographystyle{ACM-Reference-Format}
\bibliography{egbib}

\appendix

\section{Smooth Interpolation}\label{app:smooth-interpolation}

One may desire smoother interpolation than the $d$-linear interpolation that our multiresolution hash encoding uses by default.

In this case, the obvious solution would be using a $d$-quadratic or $d$-cubic interpolation, both of which are however very expensive due to requiring the lookup of $3^d$ and $4^d$ instead of $2^d$ vertices, respectively.
As a low-cost alternative, we recommend applying the smoothstep function,
\begin{align}
  \smoothstep(x) = x^2(3 - 2x) \,,
\end{align}
to the $d$-linear interpolation weights.
Crucially, the derivative of the smoothstep,
\begin{align}
  \smoothstep'(x) &= 6x(1-x) \,,
\end{align}
vanishes at $0$ and at $1$, causing the discontinuity in the derivatives of the encoding to vanish by the chain rule.
The encoding thus becomes $C^1$-smooth.

However, by this trick, we have merely traded discontinuities for zero-points in the individual levels which are not necessarily more desirable.
So, we offset each level by half of its voxel size $1/(2\resolution_\level)$, which prevents the zero derivatives from aligning across all levels.
The encoding is thus able to learn smooth, non-zero derivatives for all spatial locations $\pos$.

For higher-order smoothness, higher-order smoothstep functions $S_n$ can be used at small additional cost.
In practice, the computational cost of the $1$st order smoothstep function $\smoothstep$ is hidden by memory bottlenecks, making it essentially free.
However, the reconstruction quality tends to decrease as higher-order interpolation is used.
This is why we do not use it by default.
Future research is needed to explain the loss of quality.

\section{Implementation Details of NGLOD}\label{app:nglod-implementation}

\ADD{We designed our implementation of NGLOD~\citep{takikawa2021nglod} such that it closely resembles that of our hash encoding, only differing in the underlying data structure; i.e.\ using the vertices of an octree around ground-truth triangle mesh to store collision-free feature vectors, rather than relying on hash tables.
This results in a notable difference to the original NGLOD:\ the looked-up feature vectors are concatenated rather than summed, which in our implementation serendipitously resulted in higher reconstruction quality compared to the summation of an equal number of trainable parameters.

The octree implies a fixed growth factor ${\perLevelScale = 2}$, which leads to a smaller number of levels than our hash encoding.
We obtained the most favorable performance vs.\ quality trade-off at a roughly equal number of trainable parameters as our method, through the following configuration:
\begin{enumerate}[leftmargin=*]
  \item the number of feature dimensions per entry is ${\featuresPerEntry = 8}$,
  \item the number of levels is ${\levels = 10}$, and
  \item look-ups start at level ${\level = 4}$.
\end{enumerate}
The last point is important for two reasons: first, it matches the coarsest resolution of our hash tables ${2^4 = 16 = \minResolution}$, and second, it prevents a performance bottleneck that would arise when all threads of the GPU atomically accumulate gradients in few, coarse entries.
We experimentally verified that this does not lead to reduced quality, compared to looking up the entire hierarchy.}

\section{Real-time SDF Training Data Generation}\label{app:sdf-datagen}

In order to not bottleneck our SDF training, we must be able to generate a large number of ground truth signed distances to high-resolution meshes very quickly ($\sim$millions per second).

\subsection{Efficient Sampling of 3D Training Positions}\label{app:sdf-training-samples}
Similar to prior work~\citep{takikawa2021nglod}, we distribute some (${1/8}$th) of our training positions uniformly in the unit cube, some (${4/8}$ths) uniformly on the surface of the mesh, and the remainder (${3/8}$ths) \emph{perturbed} from the surface of the mesh.

The uniform samples in the unit cube are trivial to generate using any pseudorandom number generator; we use a GPU implementation of PCG32~\citep{oneill:pcg2014}.

To generate the uniform samples \emph{on the surface of the mesh}, we compute the area of each triangle in a preprocessing step, normalize the areas to represent a probability distribution, and store the corresponding cumulative distribution function (CDF) in an array.
Then, for each sample, we select a triangle proportional to its area by the inversion method---a binary search of a uniform random number over the CDF array---and sample a uniformly random position \emph{on that triangle} by standard sample warping~\citep{PBRT3e}.

Lastly, for those surface samples that must be perturbed, we add a random 3D vector, each dimension independently drawn from a logistic distribution (similar shape to a Gaussian, but cheaper to compute) with standard deviation $r/1024$, where $r$ is the bounding radius of the mesh.

\paragraph{Octree sampling for NGLOD}
When training our implementation of \citet{takikawa2021nglod}, we must be careful to rarely generate training positions outside of octree leaf nodes.
To this end, we replace the uniform unit cube sampling routine with one that creates uniform 3D positions in the leaf nodes of the octree by first rejection sampling a uniformly random leaf node from the array of all nodes and then generating a uniform random position within the node's voxel.
Fortunately, the standard deviation $r/1024$ of our logistic perturbation is small enough to almost never leave the octree, so we do not need to modify the surface sampling routine.

\subsection{Efficient Signed Distances to the Triangle Mesh}
For each sampled 3D position $\pos$, we must compute the signed distance to the triangle mesh.
To this end, we first construct a triangle bounding volume hierarchy (BVH) with which we perform efficient \emph{unsigned} distance queries; $\BigO\big(\log{N_\mathrm{triangles}}\big)$ on average.

Next, we \emph{sign} these distances by tracing $32$ ``stab rays''~\citep{nooruddin2003simplification}, which we distribute uniformly over the sphere using a Fibonacci lattice that is pseudorandomly and independently offset for every training position.
If any of these rays reaches infinity, the corresponding position $\pos$ is deemed ``outside'' of the object and the distance is marked positive.
Otherwise, it is marked negative.\footnote{If the mesh is watertight, it is cheaper to sign the distance based on the normal(s) of the closest triangle(s) from the previous step. We also implemented this procedure, but disable it by default due to its incompatibility with typical meshes in the wild.}

For maximum efficiency, we use NVIDIA ray tracing hardware through the OptiX \num{7} framework, which is over an order of magnitude faster than using the previously mentioned triangle BVH for ray-shape intersections on our RTX 3090 GPU.\@

\section{Baseline MLPs with Frequency Encoding}\label{app:sdf-freq-details}

\ADD{In our signed distance function (SDF), neural radiance caching (NRC), and neural radiance and density fields (NeRF) experiments, we use an MLP prefixed by a frequency encoding as baseline.
The respective architectures are equal to those in the main text, except that the MLPs are larger and that the hash encoding is replaced by sine and cosine waves (SDF and NeRF) or triangle waves (NRC).
The following table lists the number of hidden layers, neurons per hidden layer, frequency cascades (each scaled by a factor of \num{2} as per \citet{vaswani2017attention}), and adjusted learning rates.}
\begin{table}[h]
  \small
  \vspace{-2mm}
  \newcolumntype{R}{>{\raggedleft\arraybackslash}X}
  \begin{tabularx}{\linewidth}{lrrrr}%
    \toprule%
    Primitive & Hidden layers & Neurons & Frequencies & Learning rate \\
    \midrule%
    SDF & 8 & 128 & 10 & $3 \cdot 10^{-4}$ \\
    NRC & 3 & 64 & 10 & $10^{-2}$ \\
    NeRF & 7 / 1 & 256 / 256 & 16 / 4 & $10^{-3}$ \\
    \bottomrule%
  \end{tabularx}
  \vspace{-3mm}
\end{table}
\ADD{\newline For NeRF, the first listed number corresponds to the density MLP and the second number to the color MLP.\@}
\ADD{For SDFs, we make two additional changes:\ \textbf{(1)} we optimize against the relative $\Loss^2$ loss~\citep{Lehtinen:2018} instead of the MAPE described in the main text, and \textbf{(2)} we perturb training samples with a standard deviation of ${r / 128}$ as opposed to the value of ${r/1024}$ from Appendix~\ref{app:sdf-training-samples}.
Both changes smooth the loss landscape, resulting in a better reconstruction with the above configuration.}

\ADD{Notably, even though the above configurations have fewer parameters \emph{and} are slower than our configurations with hash encoding, they represent favorable performance vs.\ quality trade-offs.
An equal parameter count comparison would make pure MLPs too expensive due to their scaling with $\BigO(n^2)$ as opposed to the sub-linear scaling of trainable encodings.
On the other hand, an equal throughput comparison would require prohibitively small MLPs, thus underselling the reconstruction quality that pure MLPs are capable of.}

\ADD{We also experimented with Fourier features~\citep{tancik2020fourfeat} but did not obtain better results compared to  the axis-aligned frequency encodings mentioned previously.}

\section{Accelerated NeRF Ray Marching}\label{app:nerf-occupancy}

The performance of ray marching algorithms such as NeRF strongly depends on the marching scheme.
We utilize three techniques with imperceivable error to optimize our implementation:
\begin{enumerate}[leftmargin=*]
  \item exponential stepping for large scenes,
  \item skipping of empty space and occluded regions, and
  \item compaction of samples into dense buffers for efficient execution.
\end{enumerate}

\subsection{Ray Marching Step Size and Stopping}

In synthetic NeRF scenes, which we bound to the unit cube $[0,1]^3$, we use a fixed ray marching step size equal to ${\Delta t := \sqrt{3}/1024}$; $\sqrt{3}$ represents the diagonal of the unit cube.

In all other scenes, based on the intercept theorem\footnote{The appearance of objects stays the same as long as their size and distance from the observer remain proportional.}, we set the step size \emph{proportional} to the distance $t$ along the ray ${\Delta t := t / 256}$, clamped to the interval $\big[\sqrt{3}/1024, s \cdot \sqrt{3}/1024\big]$, where $s$ is size of the largest axis of the scene's bounding box.
This choice of step size exhibits exponential growth in $t$, which means that the computation cost grows only logarithmically in scene diameter, with no perceivable loss of quality.

Lastly, we stop ray marching and set the remaining contribution to zero as soon as the transmittance of the ray drops below a threshold; in our case ${\epsilon = 10^{-4}}$.

\paragraph{Related work.}
\ADD{\citet{mildenhall2019llff} already identified a non-linear step size as benefitial:\ they recommend sampling uniformly in the disparity-space of the average camera frame, which is more aggressive than our exponential stepping, requiring on one hand only a constant number of steps, but on the other hand can lead to a loss of fidelity compared to exponential stepping~\citep{neff2021donerf}.}

\ADD{In addition to non-linear stepping, some prior methods propose to warp the 3D domain of the scene towards the origin, thereby improving the numerical properties of their input encodings~\citep{neff2021donerf,mildenhall2020nerf,barron2021mipnerf360}.
This causes rays to curve, which leads to a worse reconstruction in our implementation.
In contrast, we \emph{linearly} map input coordinates into the unit cube before feeding them to our hash encoding, relying on its exponential multiresolution growth to reach a proportionally scaled maximum resolution $\maxResolution$ with a constant number of levels (variable $\perLevelScale$ as in \autoref{Eqn:PerLevelScale}) or logarithmically many levels $\levels$ (constant $\perLevelScale$).}

\subsection{Occupancy Grids}
To skip ray marching steps in empty space, we maintain a cascade of $K$ multiscale occupancy grids, where ${K=1}$ for all synthetic NeRF scenes (single grid) and ${K\in[1,5]}$ for larger real-world scenes (up to $5$ grids, depending on scene size).
Each grid has a resolution of $128^3$, spanning a geometrically growing domain $[-2^{k-1} + 0.5,2^{k-1} + 0.5]^3$ that is centered around ${(0.5, 0.5, 0.5)}$.

Each grid cell stores occupancy as a single bit.
The cells are laid out in Morton (z-curve) order to facilitate memory-coherent traversal by a digital differential analyzer (DDA).
During ray marching, whenever a sample is to be placed according to the step size from the previous section, the sample is skipped if its grid cell's bit is low.

Which one of the $K$ grids is queried is determined by both the sample position $\pos$ and the step size $\Delta t$: among the grids covering $\pos$, the finest one with cell side-length larger than $\Delta t$ is queried.

\paragraph{Updating the occupancy grids.}
To continually update the occupancy grids while training, we maintain a second set of grids that have the same layout, except that they store full-precision floating point density values rather than single bits.

We update the grids after every $16$ training iterations by performing the following steps. We
\begin{enumerate}[leftmargin=*]
  \item decay the density value in each grid cell by a factor of $0.95$,
  \item randomly sample $M$ candidate cells, and set their value to the maximum of their current value and the density component of the NeRF model at a random location within the cell, and
  \item update the occupancy bits by thresholding each cell's density with ${t = 0.01 \cdot 1024/\sqrt{3}}$, which corresponds to thresholding the opacity of a minimal ray marching step by ${1 - \exp(-0.01) \approx 0.01}$.
\end{enumerate}
The sampling strategy of the $M$ candidate cells depends on the training progress since the occupancy grid does not store reliable information in early iterations.
During the first $256$ training steps, we sample ${M = K \cdot 128^3}$ cells uniformly without repetition.
For subsequent training steps we set ${M = K \cdot 128^3 / 2}$ which we partition into two sets. The first ${M/2}$ cells are sampled uniformly among all cells. Rejection sampling is used for the remaining samples to restrict selection to cells that are currently occupied. 

\paragraph{Related work.}
\ADD{The idea of constraining the MLP evaluation to occupied cells has already been exploited in prior work on trainable, cell-based encodings~\citep{liu2020neural,yu2021plenoctrees,yu2021plenoxels,sun2021direct}.
In contrast to these papers, our occupancy grid is independent from the learned encoding, allowing us to represent it more compactly as a bitfield (and thereby at a resolution that is decoupled from that of the encoding) and to utilize it when comparing against other methods that do not have a trained spatial encoding, e.g.\ ``Ours: Frequency'' in \autoref{tab:nerf}.}

\ADD{Empty space can also be skipped by importance sampling the depth distribution, such as by resampling the result of a coarse prediction~\citep{mildenhall2020nerf} or through neural importance sampling~\citep{mueller2019nis} as done in DONeRF~\citep{neff2021donerf}.}

\subsection{Number of Rays Versus Batch Size}\label{app:nerf-batch-size}

\ADD{The batch size has a significant effect on the quality and speed of NeRF convergence.
We found that training from a larger number of rays, i.e.\ incorporating more viewpoint variation into the batch, converged to lower error in fewer steps.}
\ADD{In our implementation where the number of samples per ray is variable due to occupancy, we therefore include as many rays as possible in batches of fixed size rather than building variable-size batches from a fixed ray count.}

\ADD{In \autoref{tab:batch-size}, we list ranges of the resulting number of rays per batch and corresponding samples per ray.}
\begin{table}
  \small
  \vspace{0.7mm}
  \caption{\label{tab:batch-size}\ADD{Batch size, number of rays per batch, and number of samples per ray for our full method (``Ours: Hash''), our implementation of frequency encoding NeRF (``Ours: Freq.'') and mip-NeRF.
  Since the values corresponding to our method vary by scene, we report minimum and maximum values over the synthetic scenes from \autoref{tab:nerf}.}}
  \vspace{-3mm}
  \newcolumntype{R}{>{\raggedleft\arraybackslash}X}
  \begin{tabularx}{\linewidth}{lrrr}%
    \toprule%
    Method & Batch size & $=\,\,\,\,\,\,\,$ Samples per ray & $\times\,\,\,\,\,\,\,\,$ Rays per batch \\
    \midrule%
    Ours: Hash & \SI{256}{\kibi\nothing} & \num{3.1} to \num{25.7} & \SI{10}{\kibi\nothing} to \SI{85}{\kibi\nothing} \\
    Ours: Freq.\ & \SI{256}{\kibi\nothing} & \num{2.5} to \num{9} & \SI{29}{\kibi\nothing} to \SI{105}{\kibi\nothing} \\
    mip-NeRF & \SI{1}{\mebi\nothing} & \num{128} coarse + \num{128} fine & \SI{4}{\kibi\nothing} \\
    \bottomrule%
  \end{tabularx}
\end{table}
\ADD{We use a batch size of \SI{256}{\kibi\nothing}, which resulted in the fastest wall-clock convergence in our experiments.
This is ${4\times}$ smaller than the batch size chosen in mip-NeRF, likely due to the larger number of samples each of their rays requires.
However, due to the myriad other differences across implementations, a more detailed study must be carried out to draw a definitive conclusion.}

\ADD{Lastly, we note that the occupancy grid in our frequency-encoding baseline (``Ours: Freq.''; \autoref{app:sdf-freq-details}) produces even \emph{fewer} samples than when used alongside our hash encoding.
This can be explained by the slightly more detailed reconstruction of the hash encoding:\ when the extra detail is finer than the occupancy grid resolution, its surrounding empty space can not be effectively culled away and must be traversed by extra steps.}

\end{document}